%% file: arxiv_version.tex
\documentclass{article}

    \PassOptionsToPackage{numbers, compress}{natbib}
\usepackage{graphicx} 
\usepackage{comment}
\usepackage{wrapfig}
\usepackage{subcaption}
\usepackage{enumitem}
\usepackage[subtle]{savetrees}

\usepackage{iclr2026_conference,times}

\input{math_commands.tex}

\usepackage[utf8]{inputenc} % allow utf-8 input
\usepackage[T1]{fontenc}    % use 8-bit T1 fonts
\usepackage{hyperref}       % hyperlinks
\usepackage{url}            % simple URL typesetting
\usepackage{booktabs}       % professional-quality tables
\usepackage{amsfonts}       % blackboard math symbols
\usepackage{nicefrac}       % compact symbols for 1/2, etc.
\usepackage{microtype}      % microtypography
\usepackage[dvipsnames]{xcolor}         % colors
\usepackage{amsmath}
\usepackage{amsthm}
\usepackage{amssymb}

\usepackage{titlesec}

\titlespacing{\paragraph}{%
  0pt}{%              left margin
  0.3\baselineskip}{% space before (vertical)
  1em}%               space after (horizontal)

\newtheorem{definition}{Definition}
\newtheorem{theorem}{Theorem}

\def\enc{{A}}
\def\dec{{B}}
\def\encvec{{a}}

\def\ci{\textrm{CI}}

\usepackage{xcolor}
\newcommand{\jennifer}[1]{\textcolor{green!60!black}{[JL: #1]}}

\newcommand{\naomi}[1]{\textcolor{magenta}{[NPS: #1]}}
\newcommand{\igl}[1]{\textcolor{Periwinkle}{[igl: #1]}}

\title{Adversarial Concept Search: Predicting Compositional Errors From Feature Geometry}

\author{%
Jennifer Meng Lu\\
MIT\thanks{Work done at Brown University.}\\
\texttt{mlu108@mit.edu}
\And
Ruochen Zhang\\
Brown University\\
\texttt{ruochen\_zhang@brown.edu}
\And
Isabelle Lee\\
University of Southern California\\
\texttt{lee.isabelle.g@gmail.com}
\AND
David Alvarez-Melis\\
Harvard University\\
\texttt{dam@seas.harvard.edu}
\And
Ellie Pavlick\\
Brown University\\
\texttt{ellie\_pavlick@brown.edu}
\And
Naomi Saphra\\
Boston University\\
\texttt{nsaphra@bu.edu}
}

\iclrfinalcopy

\begin{document}

\maketitle

\begin{abstract}
% Because models are expected to handle a wide range of individual concepts, it is computationally infeasible to exhaustively test all possible scenarios. 
% \igl{are we maybe overclaiming a little here} \naomi{what is the overclaim exactly}
Humans cannot always intuit what scenarios are most challenging to LLMs. To capture challenging edge cases, developers either design problems to be difficult for humans or curate extensive benchmarks, but both approaches are flawed. What if we could instead focus our data collection efforts by anticipating which scenarios will challenge a model? In this paper, we use an LLM's representational geometry to predict which concept combinations it will fail on. We attribute this compositional failure to interference between salient linear features. In tasks that require systematic composition---toy programmatic settings, multihop reasoning, and multilingual factual recall---we find that when a pair of concepts is encoded near-orthogonally, the model reliably composes them. When their linear encodings are close, producing interference, the model fails to compose them. Our method reliably anticipates failure modes across different compositional tasks, \textit{without} evaluating specific compositional inputs. These results lay the groundwork to use representational geometry to identify high-risk examples, construct targeted stress tests, and provide a scalable foundation for active learning in real-world deployment.%\looseness=-1

\end{abstract}

\section{Introduction}
\label{sec:intro}
As large language models (LLMs) improve across a wide range of tasks and domains, it is increasingly difficult to identify remaining challenges. Since humans cannot reliably predict which concept combinations will be challenging for LLMs,  this gap in understanding makes dataset curation inefficient and limits our ability to anticipate LLM failure modes. Developers either design problems to challenge humans---but not LLMs----or they curate broad benchmarks, hoping to cover informative edge cases. What if we could instead anticipate which scenarios an individual model will fail on, helping to focus our data collection efforts?

We call this objective \textit{Adversarial Concept Search} (ACS): the task of identifying meaningful conceptual scenarios that are likely to induce model failure, \textit{without} evaluating the model on specific inputs that instantiate those scenarios. In this paper, we efficiently identify adversarial 
scenarios by predicting failures of \textit{compositional generalization}, specifically systematicity~\citep{fodor1988connectionism,hupkes2020compositionality}. A system generalizes compositionally when it successfully recombines known atomic concepts in novel configurations that were not observed during training. For example, if the training distribution includes ``black dog'' and ``white cat'', a compositional model will reliably process ``black cat''. This capacity is crucial for robust performance in out-of-distribution (OOD) settings.% such as SCAN~\citep{lake2018generalization}.
 By mapping these compositional capabilities, we could systematically construct custom challenge sets for efficient, targeted stress testing. 

As a practical demonstration, we successfully leverage a model's internal representational geometry to predict its failures in untested scenarios. These predictions are made directly from an abstract concept combination, without processing the target inputs themselves. This distinguishes our approach from standard error-prediction methods used in settings such as active learning~\citep{settles2009active}, which measure inference on existing inputs. When curating datasets for semi-supervised settings like language modeling, such methods are of limited practical use. In these settings, we lack diverse existing inputs and are constrained by the expense of generating and processing every possible input. To accommodate this limitation, our approach predicts errors from only a \textit{description} of the concepts involved and their atomic representations. 

ACS allows a developer to prioritize generating or collecting coherent, challenging inputs. For example, it may be prohibitively expensive to professionally translate all English corpora into Russian for a multilingual LLM, but by identifying which concepts will produce Russian errors, we can prioritize collecting Russian data in problem areas like \textit{notable deaths in Russian}. Across multiple tasks, we will predict these compositional errors from the geometry of their atomic concepts.
\begin{figure}[t]
    \centering
    \includegraphics[width=.85\linewidth]{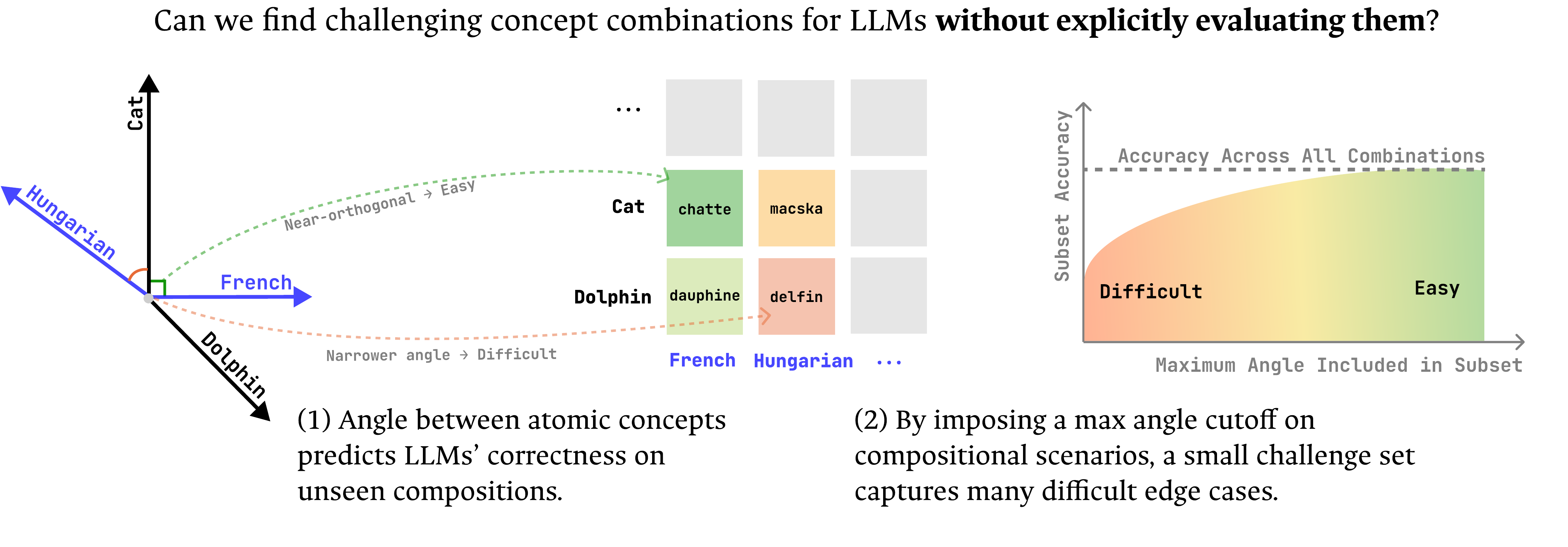}
    \caption{Angles between atomic concept representations identify the most challenging compositions, enabling failure prediction across the combinatorial space without accessing the composed inputs or evaluating them directly.}
    \label{fig:title}
     \vspace{-3mm}
\end{figure}

These predictions rely on our hypothesis that models make mistakes when features stored in \textit{superposition} cause interference. LLMs rely on superposition \citep{arora2018linear, elhage2022toy} to encode many features in limited dimensions by sharing linear directions. It has been shown that, even without orthogonal feature encodings, this compression can in principle be perfectly \textit{lossless}---that is, invertible---as long as feature activations are sparse \citep{candes2006robust}. In practice, however, we posit that models learn a \emph{lossy} coding of their input features: when multiple non-orthogonal feature encodings are simultaneously active, they can interfere with one another and damage model performance.

This interference is especially salient in compositional settings, where multiple task-relevant features must be jointly represented. We hypothesize that LLM failures can arise from geometric interference between atomic concept representations, and that these failures can be predicted from the angles between their encoded feature directions.
% We demonstrate that compositional interference destroys access to input features in models with imperfect linear decoding. 
Our conceptual model can use representational geometry to proactively identify failure cases, paving the way for dynamic stress testing and scalable active learning. Our contributions are as follows:

\begin{enumerate}[noitemsep,topsep=2pt,parsep=2pt,partopsep=0pt,leftmargin=16pt]

\item \textbf{Description and analysis of compositional interference in lossy superposition.} We attribute compositional errors to interference between  non-orthogonal atomic concept representations. Because only a few features activate at a time, prior theory argues that an ideal decoder can recover the active features even when they are encoded in superposition.
% , with similar linear representations.
We posit that when decoding is lossy, recovery error is governed by interference between the features being composed.

\item \textbf{Proof-of-concept in a controlled setting (SCAN).}
We validate this hypothesis in SCAN \citep{lake2018generalization}, a synthetic compositional generalization benchmark. In this controlled setting, we measure pairwise interactions between concept representations and show that smaller angles correlate with compositional errors. This result holds across various data conditions and model sizes, confirming the predicted relationship between interference and failure.% Across random seeds, this geometry offers signal about generalization that is not included in the model's training distribution.

\item \textbf{Predicting compositional failure in real LLMs.} We  predict successful compositional generalization in realistic LLM tasks from geometric interference between the constituent concepts. In multihop question answering (QA), the angle between component single-hop representations predicts whether the LLM will successfully compose them. In multilingual factual recall, the angle between fact representations and language subspaces predicts retrieval accuracy.  Across both settings, we find that greater separation between atomic representations corresponds to more reliable composition, demonstrating that representation geometry can predict compositional failures without evaluating the composed task itself. This establishes a scalable foundation for identifying challenging input scenarios and guiding active learning in real-world deployment.

\end{enumerate}

\section{Compositionality and Lossy Superposition}
\label{sec:theory}

%\naomi{we need to settle on defining and consistently using certain vocabulary terms, many of which are subtly different in ways that could be averted by rephrasing: correlation vs angle vs coherence vs interference, concept vs feature vs encoding vs representation, probe vs recovery vs reconstruction vs decoding, concept interference vs compositional interference etc}

We will first explain why LLMs may fail to compose non-orthogonal feature representations and then explain how to leverage this phenomenon for error prediction. Compositionality has long been associated with orthogonal atomic feature representations \citep{smolensky1990tensor}, but modern theories of feature superposition permit lossless reconstruction from non-orthogonal encodings, even with exponentially more input features than representation dimensions~\citep{candes2006robust,LRH_garg,cheung2019superpositionmodels}. In practice, however, we contend that superposition is \textit{lossy} and, therefore, that compositional errors are predictable from the geometry of the active features being combined. We then describe how features can be extracted from language models and used to empirically measure angular distance between them across different settings.

\subsection{Background: Lossy Superposition}
\label{sec:theory_intuition}

% \naomi{ideally we want to reduce this section further but use it to introduce any notation that's going to be a used in the methods}

Intuitively, one can store $d$ features in $d$ linear dimensions.
So why is lossless compression of more than $d$ features possible under current theories of superposition? 
The key assumption is \textbf{$k$-sparsity}: only $k \ll d$ features can be simultaneously active in any given input. This realistic sparsity assumption permits sufficiency theorems from compressed sensing \citep{candes2006robust}, guaranteeing that the encoded representation can be inverted to recover the original features exactly. This sparsity allows us to recover an \textit{exponential} number of features $m$ from a \textit{linear} number of dimensions $d$. 
Specifically, if we know the \textbf{linear encoder} $A \in \mathbb{R}^{d \times m}$, we can exactly recover any given \textbf{feature vector} $z \in [-1,1]^m$ with $k$-sparse \textbf{support} $\textrm{supp}(z) \subset [m]$ from its encoded representation $Az$. Given $Az$, we can reconstruct the feature vector $\hat{z}$ with zero \textbf{recovery error},
\begin{align}
    \|z - \hat{z}\|_2 = 0.
\end{align}
Crucially, \textit{this guarantee does not degrade when feature representations are non-orthogonal}. Even if we prohibit nonlinear decoding by the next LLM layer, concepts encoded with very similar representations can still be recovered using a biorthogonal decoder dictionary \citep{LRH_garg}. 

Although this ideal decoding exists under mild assumptions,\footnote{For classic compressed sensing, the assumption is Restricted Isometry Property \citep{candes2005decodinglinearprogramming}.} it is unlikely to hold during real LLM inference; we can assume some noise in the representation. However, \textit{lossless recovery is only guaranteed  in noiseless settings.} In compressed sensing theory, recovery becomes sensitive to noise when the feature encodings have high similarity \citep{Ben_Haim_2010}. Specifically, robust decoding requires an encoder matrix $A$ with low \textit{global coherence}, defined as the maximum similarity between all columns,
$$
\rho = \max_{\substack{i,j \in [m], j\neq i}} \left|\cos{(a_i, a_j)}\right|.
$$
Worst-case recovery error bounds do not depend on low coherence with an \textit{ideal} decoder, but do when decoding under \textit{noise}. Robust decoding motivates why correlated features are known to be encoded orthogonally, while anti-correlated features exhibit negative interference \citep{elhage2022toy}. 

Global coherence provides bounds on worst-case errors, but we are concerned about errors on specific feature combinations. 
In a specific scenario, not all interference is equal. We are most concerned with features \textit{salient to that scenario}---specifically, the sparse support  $\textrm{supp}(z)$ and any features that are relevant in the context of the sparse support.\footnote{One simple way to limit the salient feature set is by leveraging structured sparsity and assuming features are organized into active blocks of related features. This approach is suggested by \citet{adcock2021oracle} and inspires our concept-salient subspace construction by SVD in Section \ref{sec:real_LLM_section}.}
%Regardless of how the LLM identifies the \textbf{salient support} $\mathcal{S}(C)$ for an active concept set $C$,\footnote{For inputs instantiating concept set $C$ defined in~\ref{sec:ci_theory}, let $\mathcal{Z}(C)$ denote the induced latent activation vectors; for any $z \in \mathcal{Z}(C)$, $\operatorname{supp}(z) \subseteq \mathcal{S}(C)$.}
Regardless of how the LLM identifies the \textbf{salient support} $\mathcal{S}$, if it restricts decoding to only salient features, then the dimension required for robust recovery is controlled with high probability by the structured bound of \citet{adcock2021oracle}. In effect, this bound depends on each example's interaction with the salient support, which across examples of a shared scenario is bounded by the salient support's \textbf{local cumulative coherence},
\begin{equation}
\alpha(\mathcal{S})
=
\max_{i\in \mathcal{S}}
\sum_{\substack{j\in \mathcal{S}, j\neq i}}
\left|\cos{(a_i, a_j)}\right|.
    \label{eq:cum_coherence}
\end{equation}
Further theoretical details on the relevant bound, as well as a related bound for robust \textit{linear} compressed sensing, are provided in Appendix \ref{sec:theory_apx}.

Why might feature recovery be less robust when the salient support has high cumulative coherence? Intuitively, robust recovery is hampered by destructive interference from the most damaging feature: the salient feature that has the smallest angle with other features salient to the support. This is captured by the local cumulative coherence term, which we operationalize below as \textit{Compositional Interference} (CI). We will leverage this interference metric to rank concept combinations by likelihood of LLM compositional failure. Our intuition is simple: when an LLM cannot robustly recover a set of features in superposition  due to interference, it is more likely to make mistakes during processing.%\igl{ (possible appendix level discussion cause it's prob a lot) i know we sorta tacitly assume a perfect encoder, but do we ever mention that this recovery error could also be encoder error?}

\subsection{Measuring compositional interference}
\label{sec:ci_theory}
The theory above predicts that compositional failures should be more likely when the features active or salient in a compositional input have high cumulative coherence. How can we compute this value \textit{without access to an example} of the compositional scenario? In real LLMs, the ground-truth concepts learned by the model are not directly accessible. While methods such as Sparse Autoencoders (SAEs)~\citep{cunningham2023sparse} are proposed to disentangle the representation space into discrete features, they introduce additional assumptions and implementation challenges. For a proof of concept, we will prefer a simple proxy for compositional interference: using models' residual representations directly.

\paragraph{Atomic concepts and salient features.}
We distinguish between atomic \textit{concepts} expressed in the input and latent \textit{features} in the model representation. Let \(\mathcal{C}\) denote the set of atomic concepts, and let \(C \subseteq \mathcal{C}\) denote a set of active atomic concepts which are instantiated by a specific input \(x \in \mathcal{X}(C)\). These concepts correspond to a salient feature support \(\mathcal{S}(C)\), i.e., the indices of latent features that are active or relevant for these concepts.\footnote{In particular, each input $x$ has a corresponding feature vector $z(x)$; for any \(x \in \mathcal{X}(C)\), \(\operatorname{supp}(z(x)) \subseteq \mathcal{S}(C)\).} For example, the concept \texttt{Spanish} may correspond not to a single direction, but to a set of feature directions that are highly active in Spanish text and encode different aspects of the language. 
\begin{figure}[t]
    \centering

    \begin{subfigure}[t]{0.45\textwidth}
        \centering
        \includegraphics[width=\linewidth]{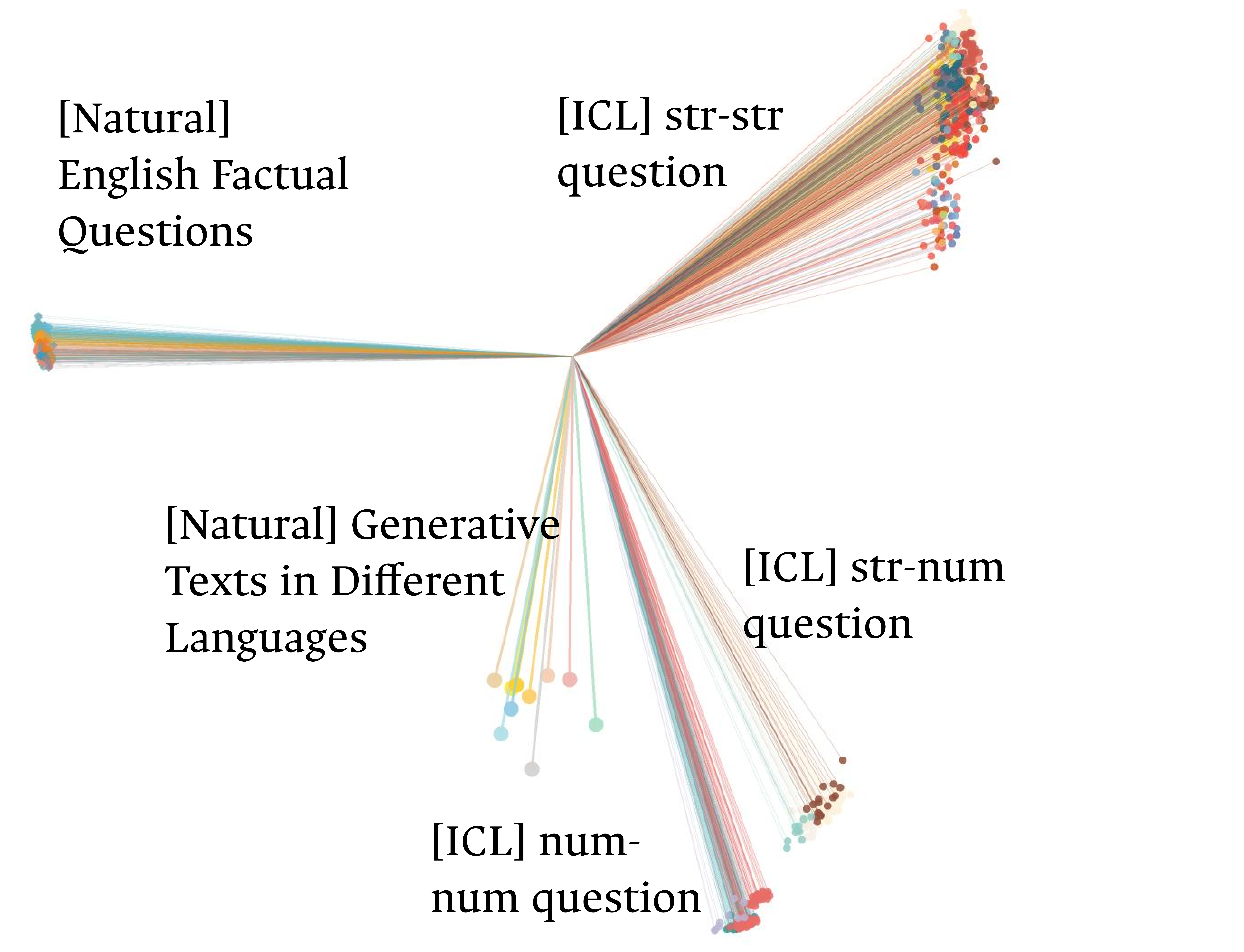}
        \caption{Large-scale structure in representation space.}
        \label{fig:multiscale_vis}
    \end{subfigure}
    \hspace{0.02\textwidth}
    \begin{subfigure}[t]{0.33\textwidth}
        \centering
        \includegraphics[width=\linewidth]{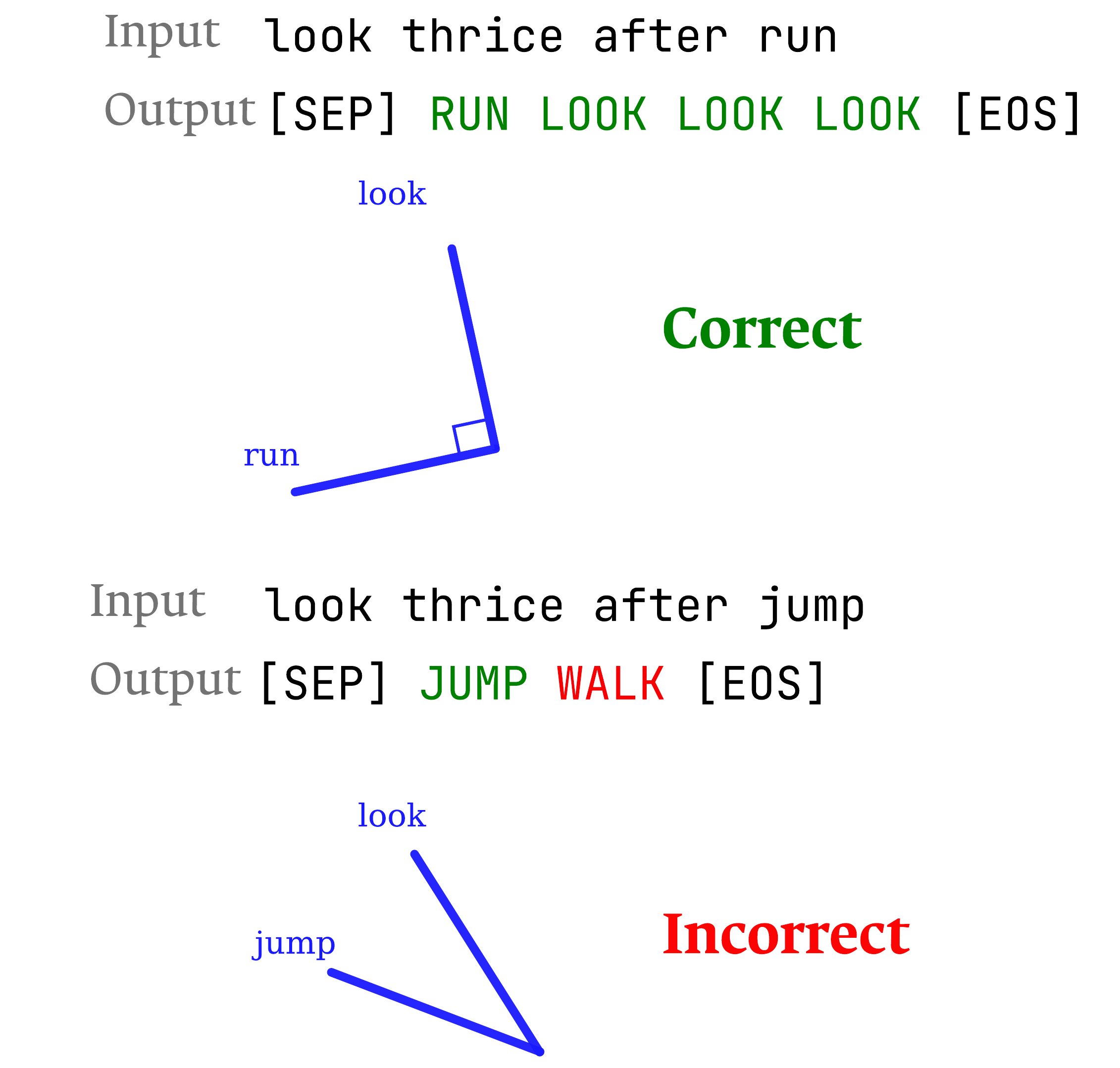}
        \caption{SCAN example.}
        \label{fig:scan_example}
    \end{subfigure}

    \caption{\textbf{After controlling large-scale representation structure, compositional generalization can be predicted from feature interference.} 
    (a) Representations cluster by global structures that are irrelevant to the concepts of interest. Colors indicate examples from the same topic subset (Layer 9).
    (b) Hypothesized illustration of compositional failure for Section~\ref{sec:scan}. Two commands share the same structure but differ in a single concept (\texttt{run} vs.\ \texttt{jump}). The relevant atomic concepts at the top are near-orthogonal, enabling correct composition.
    }
    \label{fig:figure2_combined}
     \vspace{-3mm}
\end{figure}
\paragraph{Estimating salient feature encodings from examples.}
In real LLMs, neither the salient support \(\mathcal{S}(C)\) nor the salient feature encodings \(A_{\mathcal{S}(C)} = \{a_i : i \in \mathcal{S}(C)\}\) are directly observed. We therefore estimate \(A_{\mathcal{S}(C)}\) using residual-stream representations from examples that instantiate each concept in \(C\). Preferably, for an atomic concept \(c \in C\), a vector representation would simply use a residual-stream representation from a single input \(x \in \mathcal{X}(c)\) that isolates  concept $c$. In practice, the best proxy depends on how the concept is instantiated in each empirical setting. Accordingly, we approximate concept representations using a single activation vector, an average across contexts, or a set of high-variance directions across examples of the concept; in the last case, multiple feature encodings in \(A_{\mathcal{S}(C)}\) would jointly represent the atomic concept. For example, in multilingual fact recall, the input \texttt{la capital de Espa\~na es} instantiates both a factual concept, \texttt{capital of Spain}, and a language concept, \texttt{Spanish}. We estimate their representations separately: the fact is represented  by activations from the English version of the input (\texttt{the capital of Spain is}) while the Spanish concept is represented by high-variance directions across Spanish texts.

\paragraph{Accounting for multiscale structure.}
Raw representations in LLMs are often dominated by large-scale structure irrelevant to the atomic concepts of interest. In our LLM experiments (Section~\ref{sec:real_LLM_section}), residuals cluster by prompt type, task family, and other contextual factors (dominant clusters are highlighted in Figure~\ref{fig:multiscale_vis}.) As a result, these cluster identities can dominate the inner products used to estimate angles between atomic representations (see Appendix~\ref{sec:appendix_mean_centering}). Empirically, angles computed from raw representations are dominated by background cluster identities (Appendix Figure~\ref{fig:multiscale_discrete_ortho_bins}(a)).

To mitigate this effect, we apply mean-centering to the dominant clusters in raw representations before estimating salient feature encodings. Let \(x \in \mathcal{X}(C)\) be an example, let \(\gamma(x)\) denote its background cluster, and let \(\mu_{\gamma(x)}\) be the empirical mean residual representation of examples in that cluster. We define the cluster-centered residual representation as \(h_c(x) = h(x) - \mu_{\gamma(x)}\). We then use these centered residuals to estimate the salient feature encodings \(A_{\mathcal{S}(C)}\) used for angle computation. This centering places angle values from different background clusters on a more comparable scale, reducing artificial binning effects caused by dominant clusters (Appendix Figure~\ref{fig:multiscale_discrete_ortho_bins}(b)) and improving predictions (Figure~\ref{fig:appendix_multihop_other_results}(a,c)). \looseness-1

\paragraph{Estimating compositional interference.}
We define \textbf{compositional interference} (CI) for a composition with active concept set \(C\) as a normalized\footnote{Normalization ensures that differences in difficulty are not solely due to different salient feature counts.} variant of local cumulative coherence over its salient support:
\begin{equation}
\label{eq:ci}
\ci(C)
=
\max_{i\in \mathcal{S}(C)}
\frac{1}{|\mathcal{S}(C)|}
\sum_{\substack{j\in \mathcal{S}(C), j \neq i}}
|\cos(a_i,a_j)| .
\end{equation}
Here, \(a_i\) and \(a_j\) denote empirical encodings of the salient latent feature directions indexed by \(i\) and \(j\). CI lets us estimate interference for compositional inputs without evaluating the model on any specific composed input, since \(a_i\) and \(a_j\) are estimated from examples of the constituent atomic concepts.\footnote{Appendix~\ref{sec:alternative_ci_metrics} provides alternative metrics for quantifying compositional interference.}  According to Section~\ref{sec:theory_intuition}, higher interference among salient features predicts higher recovery error.

\subsection{Using and evaluating predictions for ACS}
\label{sec:metrics}
%Given the atomic representation estimates described above, we can compute CI as defined in Section~\ref{sec:theory_intuition}, \footnote{Additional ways of computing interference between local salient features are reported in Appendix Section~\ref{sec:appendix_other_metrics}.}. Importantly, this score is computed from atomic examples alone: it does not require running the model on any compositional inputs.
%add that : in each setting, We measure errors by full prediction exact-match accuracy.
We hypothesize that recovery error affects model accuracy in  compositional scenarios: if the model cannot reliably recover the atomic concepts in representation space, it is less likely to execute the composition correctly. This claim allows us to use CI, computed from examples of atomic concepts, for our proposed objective of Adversarial Concept Search. Generally, ACS aims to answer, ``if resource constraints only allow you to test inputs for a small subset of all possible concept combinations, which subset should you collect to maximize error coverage?'' To demonstrate the efficacy of CI in choosing these adversarial scenarios, we will use it to predict compositional errors. For each experiment, we measure errors by full-prediction exact-match accuracy. We select a representation layer to provide residual-stream feature representations using a 10\% validation set (see Appendix Section~\ref{sec:layer_sensitivity} for layer analysis).

In each setting, we illustrate that CI provides a useful ranking of difficulty for the purposes of selecting a challenge set of scenarios under specified resource constraints. Assuming higher CI is associated with more compositional failure, we construct subsets of concept combinations with high CI with varying minimum cutoffs. We measure accuracy on cutoffs as shown in Figure~\ref{fig:title}, where the left side of the curve contains only the examples predicted to be hardest. The $x$-axis reports decreasing CI cutoffs by percentile, and the $y$-axis reports accuracy on the cumulative challenge set. If we select the same size of test sets at random, we would expect a flat line at the model's overall accuracy. If $\ci$ is predictive, the curve should instead start low and monotonically increase toward the mean as lower-interference examples are included.

We also directly evaluate error predictions by their PR-AUC. We rank all examples by CI, assuming higher CI is associated with higher likelihood of errors. The PR-AUC baseline is the model's overall  failure rate, $1-\mathrm{accuracy}$, which is the score of a majority-label binary predictor baseline. 
% This metric is needed because failures can be class-imbalanced: cumulative accuracy measures whether $\ci$ gives a useful difficulty ordering, while PR-AUC measures whether errors are concentrated among the examples predicted to be hardest.

% To assess whether predictability is significantly above baseline, we use a permutation test for both metrics: we shuffle $\ci$ across examples, recompute each metric, and compare the observed value to the resulting null distribution.

\begin{figure}[t]
    \centering

    \begin{subfigure}[t]{0.48\textwidth}
        \centering
        \includegraphics[width=\linewidth]{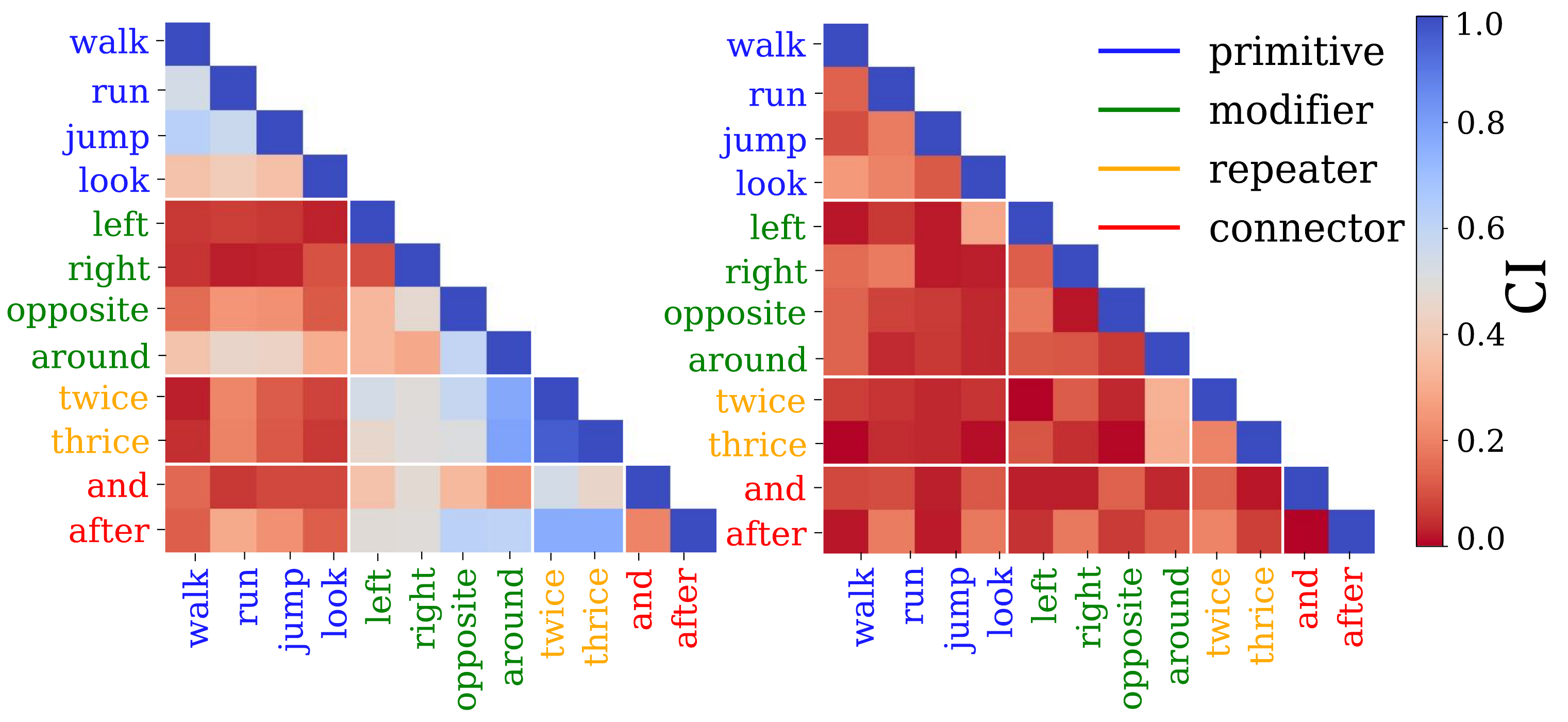}
        \caption{CI between atomic concepts.}
        
    \end{subfigure}
    \hspace{0.02\textwidth}
    \begin{subfigure}[t]{0.47\textwidth}
        \centering
        \includegraphics[width=\linewidth]{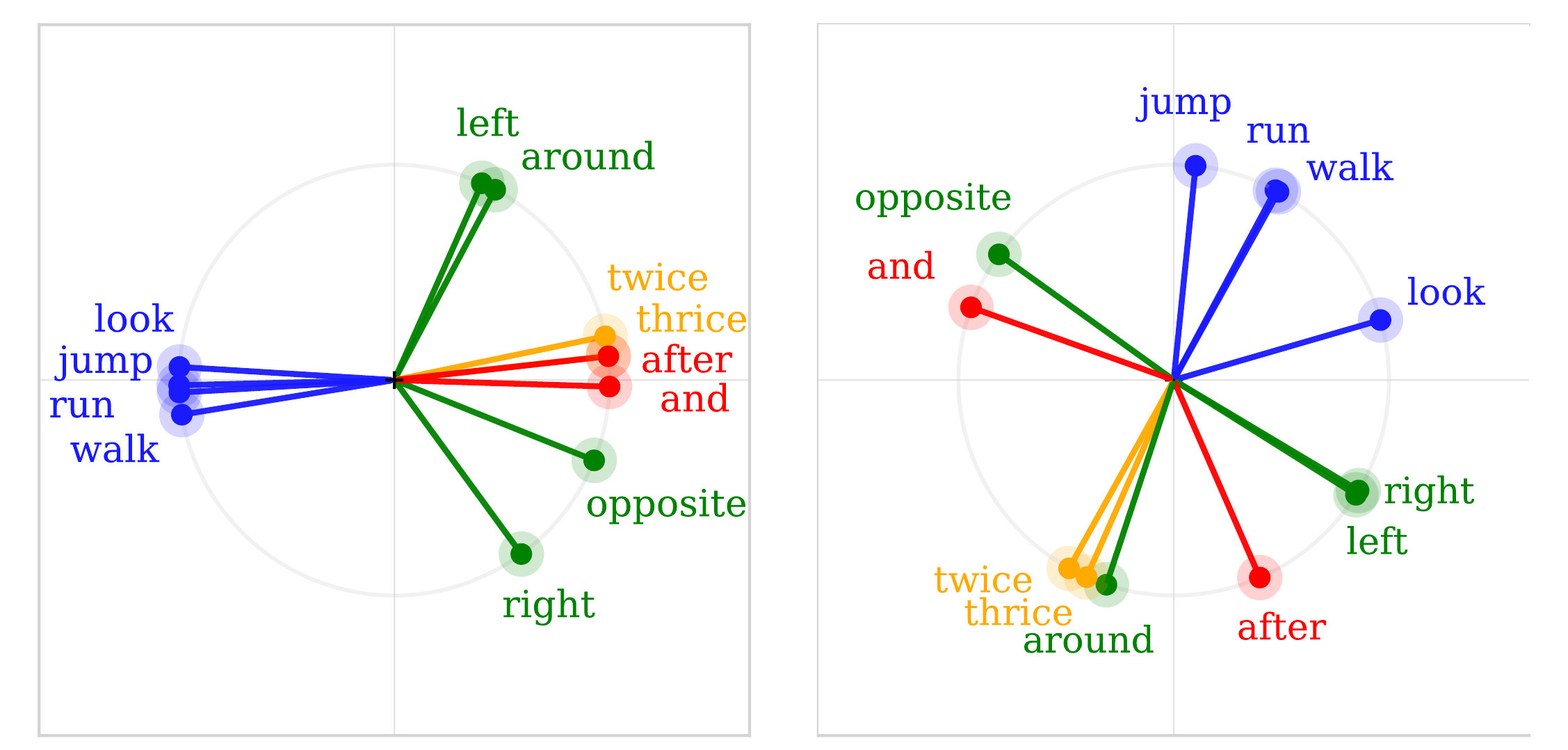}
        \caption{Atomic concept representations (PCA).}
        
    \end{subfigure}

    \caption{
    \textbf{Compositional models exhibit distinct representational geometries.} Representations of atomic concepts in 64-dimension SCAN models trained with 8\% coverage (left) and 80\% coverage (right). Models trained under different configurations exhibit distinct representational geometries, producing different patterns of alignment and potential interference among atomic concepts.
    }
    \vspace{-3mm}
    \label{fig:scan_heatmap}
\end{figure}

\section{Adversarial Concept Search for a synthetic task}
\label{sec:scan}

We first validate our hypothesis in a controlled synthetic setting by training toy models from scratch. This allows us to control the data distribution and model scale, isolate the role of feature interaction, and directly test whether representation geometry predicts compositional generalization.
\paragraph{Dataset} SCAN benchmark \citep{lake2018generalization} is a testbed for systematic compositional generalization.
% SCAN consists of synthetically generated commands paired with actions, defined by a deterministic interpretation function over a phrase-structure grammar. 
SCAN specifies a set of primitive concepts, including actions (e.g., \texttt{jump}, \texttt{walk}) and  operators (e.g., \texttt{left}, \texttt{twice}), which can be combined into instructions (see Figure~\ref{fig:scan_example}). The controlled setting helps examine whether the angles between atomic concepts predict compositional failures.

%\begin{figure}
%    \centering
%    \includegraphics[width=0.9\linewidth]{figs/scan_heatmap_pca.pdf}
%    \caption{For a toy model with dimension 64, comparing coverage ratios 0.08 and 0.8. Higher %training coverage leads to (a) greater pairwise orthogonality and (b) wider separation in PCA space.}
%    \label{fig:scan_heatmap}
%\end{figure}

\begin{figure}[t]
    \centering
    \includegraphics[width=1.0\columnwidth]{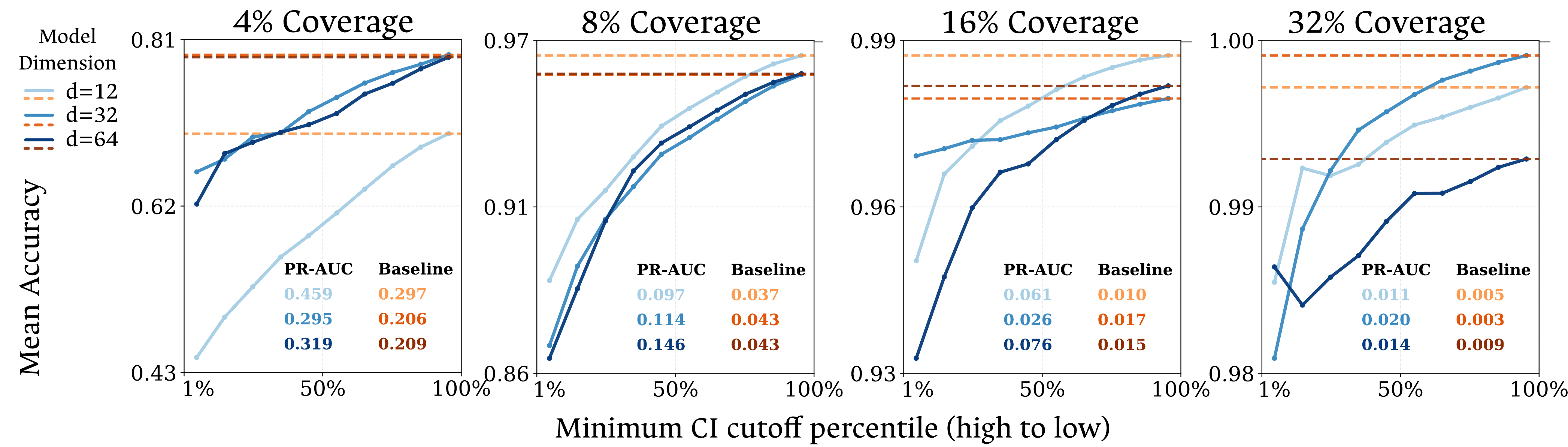}
    \caption{
    % Predictive power of representation geometry
    \textbf{SCAN models fail on examples with high interference.} Model accuracy on SCAN test sets with varying cutoffs for minimum $\ci$ (more models in Appendix Figure~\ref{fig:appendix_scan_cumulative}). Horizontal lines show each model's overall accuracy as a baseline; blue curves show test set accuracies when examples are sorted by \(\ci\). Subset accuracy responds near-monotonically to $\ci$, significantly outperforming a random ordering.  CI ranking's PR-AUC beats the error rate baseline for all models.}
    \label{fig:scan_results}
    \vspace{-3mm}
\end{figure}

\subsection{Experiments} 
%To predict model errors, we compute CI for each example using the representations of the concepts that appear in that example as the active concept set. We hypothesize that lower interference corresponds to near-orthogonal representations and reliable composition, while higher interference increases recovery error and leads to compositional failure. 
% This provides an interpretable mechanism linking representation geometry to compositional generalization.
%To predict model errors on SCAN, we compute CI for each example using the SCAN primitives and operators that appear in the command as the active concept set \(c\). In this setting, the salient support \(\mathcal{S}(\mathcal{Z}(c))\) is approximated by the extracted representations of these active concepts. We hypothesize that lower interference corresponds to near-orthogonal representations and reliable composition, while higher interference increases recovery error and leads to compositional failure. 
To predict model errors on SCAN, for each example, we use the primitive concepts that appear in the command as the active concept set \(C\), and estimate the salient feature encoding set \(A_{\mathcal{S}(C)}\) by averaging residual activations for each primitive across contexts containing that concept, e.g., averaging activations at the \texttt{jump} token across commands containing \texttt{jump}. We compute CI over these estimated salient feature encodings and hypothesize that lower interference corresponds to near-orthogonal representations and reliable composition, while higher interference increases recovery error and leads to compositional failure.

To study compositional generalization under varying difficulty, we follow \citet{lake2018generalization} in restricting the training set coverage to \(\{4, 8, 16, 32, 64, 80\}\%\) of distinct commands while keeping the total number of training examples fixed at 100K. We provide training details in Appendix~\ref{sec:appendix_scan}. Models trained with lower coverage underperform on novel compositions at test time. To study how model capacity affects  CI, we train autoregressive decoder-only Transformers with model sizes $d \in \{8, 12, 32, 64\}$ for each coverage ratio. Intuitively, smaller models enforce stronger compression and superposition, while larger models permit increasingly disentangled representations. 

To establish SCAN as a testbed for our hypothesis, we first show that it allows us to indirectly control CI. As shown in Figure~\ref{fig:scan_heatmap} and Appendix Figure~\ref{fig:appendix_scan_heatmaps}, representation structure varies with training coverage and model capacity. Figure~\ref{fig:scan_heatmap} shows that lower coverage causes representations to cluster by functional role (e.g., actions vs.\ compositional operators). As coverage increases, representations become more isotropic, allowing us to control interference and generalization through coverage ratio.

Figure~\ref{fig:scan_results} confirms that example-level \(\ci\) predicts compositional behavior within each SCAN model. The cumulative accuracy curves show that CI provides a useful ranking of example difficulty---stricter CI cutoffs provide harder test sets with lower model accuracy. Across all models with non-extreme accuracies,\footnote{Defined as $0.2 < \mathrm{acc} < 0.99$; outside of this range, compositional generalization either fails entirely or saturates. See discussions in Appendix~\ref{sec:scan_nonsaturated}.} this ranking is significantly better than random ordering ($p<0.01$).

The noncumulative curves in Appendix
Figure~\ref{fig:appendix_scan_noncumulative} further show that the effect is not
driven by a small set of extreme examples, as accuracy generally decreases
across CI-ranked bins. Moreover, in every evaluated non-saturated setting, CI's
failure-detection PR-AUC significantly exceeds the corresponding failure-rate
baseline.\footnote{Permutation tests yield \(p<0.001\); full results with confidence intervals are reported in Appendix
Table~\ref{tab:pr_auc_results}.} We rule out potential confounds such as command-length, co-occurrence and other training-data statistics using correlational tests and controlled random seeds experiments in Appendix~\ref{sec:confounds}. 

Overall, $\ci$ is a reliable signal for compositional difficulty across model sizes and training regimes. Using only a single geometric property---and without access to a specific input example---we can rank compositional scenarios by difficulty.

%: predict failures of compositional generalization,
\section{Adversarial Concept Search for LLMs}
\label{sec:real_LLM_section}

We now extend our toy proof-of-concept to predict compositional errors in real LLMs. We evaluate our approach on two different tasks: multihop QA and multilingual fact retrieval. This section will demonstrate the power to predict errors across diverse input distributions and surface forms.

\subsection{Experiments}

For both LLM tasks, we focus on examples where the LLM succeeds on the atomic components. This lets us isolate compositional errors---the model has the pieces but fails to combine them.
% We then test whether these compositional failures can be explained by compositional interference computed from atomic representations alone.

\paragraph{Multihop Reasoning.}
We study two-hop factual reasoning using the dataset and 10-shot prompts from \citet{apoorv_multihop} (Appendix~\ref{sec:appendix_multihop_dataset}). We evaluate Llama-3.2-3B~\citep{grattafiori2024llama}, OLMo-7B~\citep{olmo2025olmo3}, and Qwen2.5-14B~\citep{qwen2025qwen25technicalreport}, spanning model families and sizes. Each composed query \(g(f(x))\) combines first- and second-hop concepts \(f\) and \(g\). For example, a composed query \(g(f(x))\) may combine an atomic first-hop query \(f\), such as \texttt{author of 1984}, with an atomic second-hop query \(g\), such as \texttt{birthyear of George Orwell}, into the two-hop prompt \texttt{birthyear of the author of 1984}. We extract and cluster-center last-token residual activations from the atomic queries to obtain concept representations \(a_f\) and \(a_g\), which estimate \(A_{\mathcal{S}(C)}\). We then compute CI to predict failure on the composed query \(g(f(x))\), using only the atomic representations.

\paragraph{Multilingual Fact Recall.}
We investigate multilingual factual recall on Llama-3.2-3B~\citep{grattafiori2024llama}, Llama-2-7B~\citep{touvron2023llama2openfoundation}, and Qwen2.5-14B~\citep{qwen2025qwen25technicalreport}, using the KLAR dataset \citep{lostinmultilinguality}, which tests if models can answer facts across different languages consistently. When model capabilities differ by language, prior work has linked this inconsistency to a lack of alignment in internal representations~\citep{ai2025knowledgereferencemultilinguallanguage,lim2025languagespecificlatentprocesshinders,bu2026alignoncebenefitmultilingually,blum2025rosettastoneunificationforces}. Although this setting is not usually framed as a compositional task, prior work~\citep{lostinmultilinguality,lu_paths_2025,wendler-etal-2024-llamas, zhao2024large,shani-basirat-2025-language,blum_beyond_2025} suggests that models implement a multi-stage process: map a non-English query onto a language-agnostic representation, retrieve the answer, then generate it in the target language. This process induces a compositionality gap analogous to multihop reasoning: even when the model can perform the atomic steps, it still fails when they are combined.
We thus treat multilingual fact-recall as requiring the composition of two atomic parts: a factual representation and a language-specific representation. 
% The factual component captures \emph{what} the model knows, while the language component captures \emph{how} the model must express it. 

\begin{figure}[t]
    \centering
    \includegraphics[width=\columnwidth]{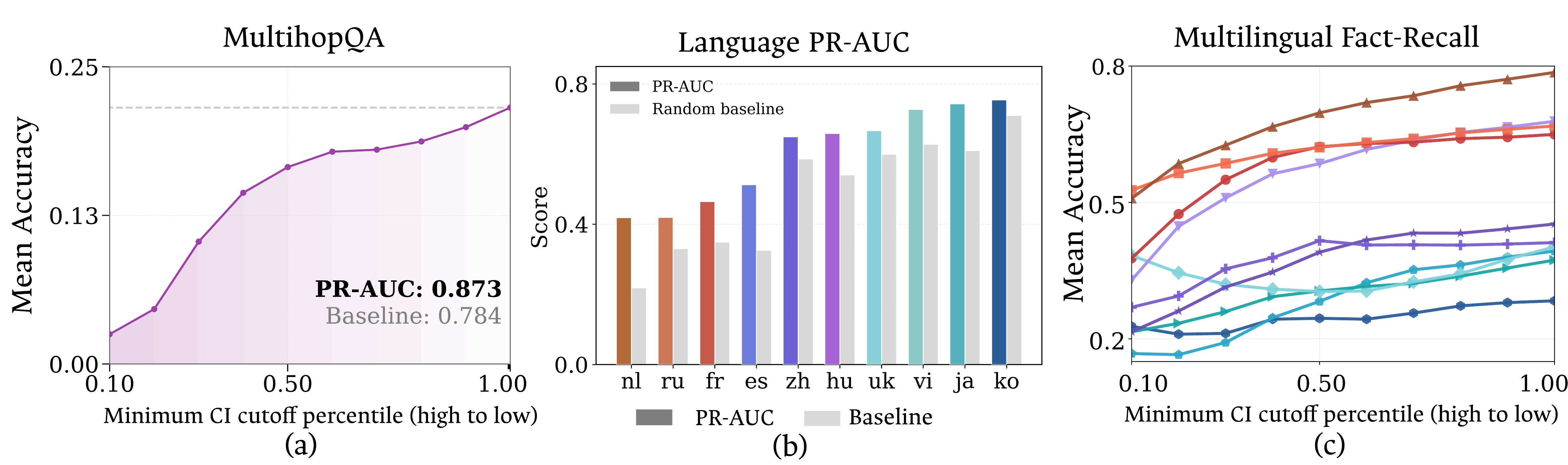}
    \caption{
    \textbf{Llama-3B fails on examples with high interference.}
    (a) For multihop QA, CI provides a strong ranking of example difficulty.
    (b) For multilingual fact recall, PR-AUC for the CI-based error prediction outperforms a majority baseline on every language.
    (c) For multilingual fact recall, CI ranking chooses difficult challenge sets for each language. (Each line is one language; color legend matches that in (b).)
    }
    \label{fig:real_LLM_results}
    \vspace{-6mm}
\end{figure}

For each query in target language \(\ell \in \mathcal{L}\), the active concept set \(c\) consists of a factual concept \(q\) and a language concept \(\ell\). We extract the fact representation from the corresponding English factual query \(q \in \mathcal{Q}\) and cluster-center it to obtain \(a_q\) (see Appendix~\ref{sec:appendix_mean_centering}). Drawing on prior work~\citep{tyler_multilingual}, we represent the language concept \(\ell\) as a low-rank subspace rather than a single feature vector. For each language \(\ell\), we collect residual representations from 1{,}200 sequences (each consists 200 tokens) in the multilingual OSCAR corpus~\citep{oscar_dataset} and apply uncentered SVD to obtain an orthonormal basis \(B_\ell\).\footnote{We retain the basis capturing \(0.99\) of the variance; we tune this variance threshold on the development set by sweeping \(0.85\), \(0.90\), \(0.95\), and \(0.99\).} In this setting, we estimate \(A_{\mathcal{S}(C)}\) = \(\{a_q\} \cup B_\ell\) and compute \(\mathrm{CI}\) to predict failure on the multilingual fact query. This setting makes the combinatorial challenge especially pronounced: each fact can be paired with any language, so the number of fact-language combinations is $|\mathcal{Q}| \times |\mathcal{L}|$. Evaluating---and, in a real-world setting, translating---all such combinations is expensive. We therefore ask whether $\ci$, computed from an English fact activation vector and a target-language subspace, can predict cross-lingual transfer failure without access to the translated input.

\subsection{Results}

\textbf{Ranking multihop reasoning examples.} \textbf{Ranking multihop reasoning examples.} Figure~\ref{fig:real_LLM_results}(a) presents results for Llama-3B, showing that CI is highly predictive of compositional failures in multihop reasoning. The curve shows a clear monotonic trend: the LLM has lower accuracy on ACS challenge sets with stricter CI cutoffs. Appendix Figure~\ref{fig:appendix_multihop_other_results}(b) shows the same relationship at the bin level, confirming a strong negative correlation between $\mathrm{CI}$ and accuracy ($r=-0.855$). Failure PR-AUC beats the model's failure-rate baseline, indicating that CI helps identify the specific multihop examples the model gets wrong (details in Appendix Table~\ref{tab:exp1_pr_auc_summary}). Crucially, this prediction is obtained without evaluating the composed query itself: we can anticipate which multihop questions will be difficult solely from interference between their constituent atomic queries. OLMo-7B and Qwen2.5-14B show similarly strong relationships between CI and failure in Appendix Figure~\ref{fig:appendix_generalizability_olmo7b} and~\ref{fig:appendix_generalizability_qwen14b}, demonstrating that CI generalizes across model families and scales.

\begin{figure}[t]
    \centering
    \includegraphics[width=0.9\columnwidth]{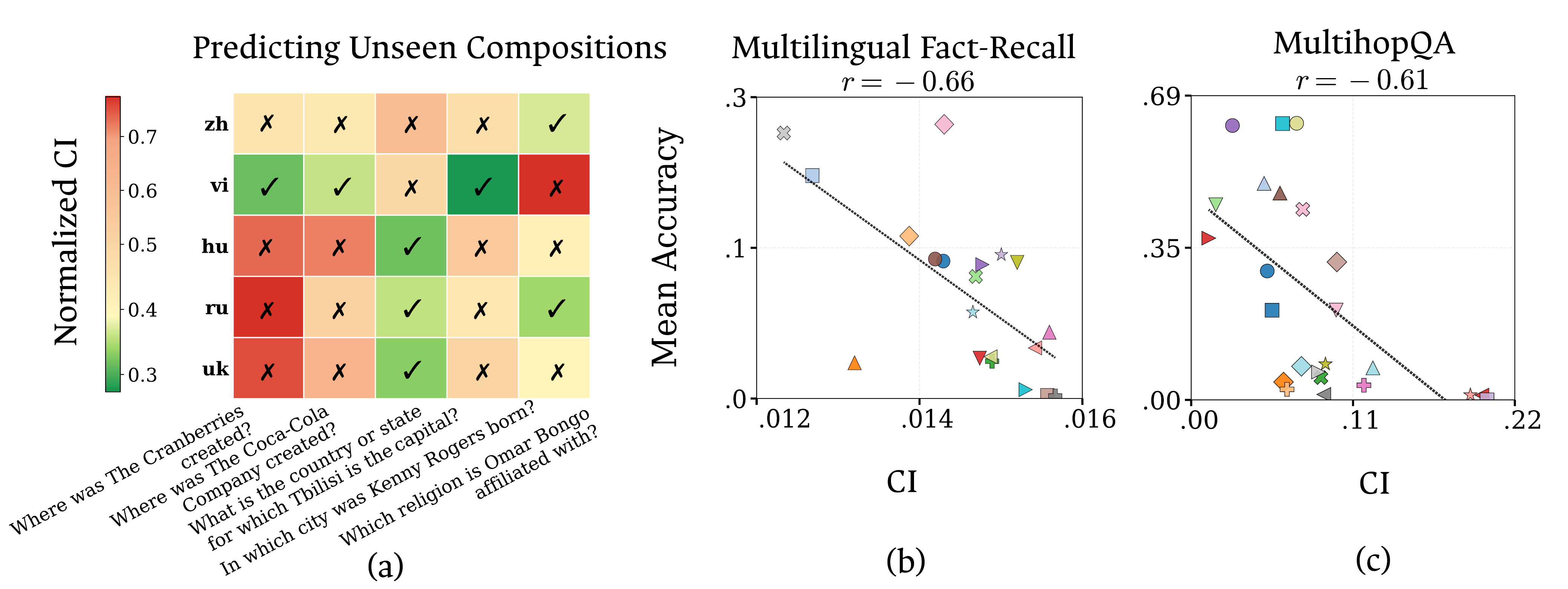}
    \caption{
    \textbf{LLMs fail on datasets with high interference.}
        Marks show the multilingual correctness for each fact-language pair; color indicates language-normalized CI. CI broadly predicts whether the model answers correctly.
        (b,c) Subspace-level CI strongly correlates (p<0.01) with accuracy in both tasks (Japanese shown; all languages in Appendix Figures~\ref{fig:appendix_subspace_subspace_multilingual1} and~\ref{fig:appendix_subspace_subspace_multilingual2}). Dataset and task legends are in Appendix Figure~\ref{fig:appendix_coarse_fig_dataset_labels}.
        }
    
    \label{fig:real_LLM_sub_sub}
        \vspace{-8mm}
\end{figure}

\textbf{Ranking multilingual fact examples.} We evaluate examples separately within each language, thereby controlling for language-level factors such as resource availability and orthographic similarity to English that may affect knowledge transfer.  For each target language, Figure~\ref{fig:real_LLM_results}(c) shows that Llama-3B performs worse on datasets with higher interference between the English fact representation and the target-language subspace. Appendix Figure~\ref{fig:appendix_multilingual_noncumulative} confirms that accuracy also correlates with individual CI bin ranges for each language.  PR-AUC metrics (Figure~\ref{fig:real_LLM_results}(b))  confirm the predictive accuracy of CI: for each language, CI beats the corresponding language-specific baseline. As an illustration, Figure~\ref{fig:real_LLM_sub_sub}(a) visualizes example fact-language combinations, displayed with CI scores and whether the LLM answers correctly. 
Qualitatively, we observe that correct model predictions have a lower CI, while model errors are associated with a higher CI. 

Results generalize to Llama-2-7B and Qwen2.5-14B (Appendix Figures~\ref{fig:appendix_generalization_multilingual_llama7b}--\ref{fig:appendix_generalization_multilingual_qwen14b}): CI beats the baseline except for Chinese on Qwen, possibly because its bilingual training enables facts to be represented directly in Chinese rather than through the compositional pathway (see Appendix Section~\ref{sec:appendix_generalizability}). Overall, these results suggest that CI could predict cross-lingual transfer failures using only the geometry between English fact representations and target-language subspaces, without access to the true translation.

\paragraph{Ranking datasets by coarse-grained concepts.}
Thus far, CI has predicted the difficulty of individual scenarios. We now extend it to entire datasets by representing each fact category as a subspace. For example, rather than predicting failure on \texttt{the capital of Spain} in Spanish, we predict overall accuracy on Spanish national-capital questions by measuring interference between the national-capitals and Spanish-language subspaces (Appendix Section~\ref{sec:appendix_coarse_grained}). In multilingual factual recall, CI shows a strong negative correlation with LLM accuracy for most languages (Figure~\ref{fig:real_LLM_sub_sub}(b), Appendix Figures~\ref{fig:appendix_subspace_subspace_multilingual1} and~\ref{fig:appendix_subspace_subspace_multilingual2}). In multihop reasoning, the correlation is also negative and significant (Figure~\ref{fig:real_LLM_sub_sub}(c)). These results suggest CI can predict the general difficulty of broad compositional tasks.

\section{Discussion and Future Work}
\label{sec:discussion}
\vspace{-2mm}
%\igl{maybe split this section -> (shorter) related work, discussion \& conclusion}
\paragraph{Compositionality and orthogonality.} 
Orthogonal structures are associated with adaptation to new contexts in brain scans \citep{flesch2022orthogonal,luettgau2024neural}. \citet{smolensky1990tensor} first proposes orthogonal representations as a path to variable binding in connectionist systems and \citet{plate1995holographic} later proposes a method for encoding compositional atomic properties with quasi-orthogonal vectors.  Recent work in toy models~\citep{uselis2025doesdatascalinglead} suggest that compositional models use a linearly factored representational structure. Relatedly, \citet{olah2023distributed} frames composition and superposition as competing strategies for allocating limited capacity among near-orthogonal features.
However, these previous works are speculative or limited to small toy settings and do not use their insights to predict whether a model will succeed at specific combinations from their geometry. Furthermore, superposition may not conflict with composition at all; compressed sensing avoids catastrophic interference between non-orthogonal feature encodings in realistic settings \citep{barin2026stop}, decoupling orthogonality from compositional fidelity in theory. By assuming superposition to be lossy, we explain the importance of orthogonality in practice and operationalize our hypothesis through testable predictions in realistic settings in LLMs. 

\paragraph{Adversarial examples.}
Neural networks make significant classification errors after small perturbations of their inputs, permitting adversarial generation  \citep{szegedy2014intriguingpropertiesneuralnetworks,goodfellow2014generativeadversarialnetworks}. This phenomenon has long been attributed to compression artifacts, as perturbations can coordinate many superposed features into a large downstream error~\citep{elhage2022toy,gorton2025adversarial}. Our approach exploits these same compression artifacts, but holds several key advantages. Because traditional adversarial generation continuously perturbs a ground-truth input, it is limited in discrete modalities such as language, where continuous perturbations rarely yield valid, coherent sequences. Furthermore, it requires a specific input nor a target output, while ACS needs neither.

% \igl{like here ish sounds like a good place for split}
% \paragraph{Why does compositional interference occur?} If two features never appear in similar contexts during training, there is no gradient to specifically separate them. \naomi{either expand this argument or cut it}

\paragraph{Predicting  model behavior}
In interpretability, hypotheses are often validated by predicting how a model will respond to targeted mechanistic interventions~\citep{saphra2024mechanistic}.
% , such as activation patching ~\citep{nanda2022mechanistic,meng_rome,wang_ioi,pearl}. 
However, researchers rarely aim to predict how an \textit{unaltered} model will naturally behave when processing novel, complex \textit{inputs}. If our goal is to understand holistic \textit{computation} rather than merely local \textit{implementation}, we must validate our theories by anticipating a model's edge-case failures from its internal representations without requiring manual perturbations. Recent work provides evidence that such prediction is possible: OOD behavior can be predicted from loss-landscape geometry~\citep{juneja2023linearconnectivityrevealsgeneralization}, hidden activations~\citep{li2025interpretationpredictbehaviorunseen}, circuit descriptions~\citep{gurnee2026modelsmanipulatemanifoldsgeometry}, causal mechanisms \citep{huang2025internal}, and in-distribution manifolds~\citep{sueyeon_geo_markers}. Specifically for compositional generalization, \citet{an2026representationalhomomorphismpredictsimproves} predicts OOD compositional generalization on SCAN, and \citet{blum_beyond_2025} show that alignment among training examples predicts cross-lingual generalization. However, both approaches require representations from compositional examples, whereas we only require the atomic concept representations. Overall, previous analyses are constrained to simple tasks or specific task domains, while our approach applies to any setting where models are expected to systematically compose atomic concepts.
%However, both approaches require representations from compositional examples, whereas we only require the atomic concept representations. 

% To study this prediction problem in practice, we focus on tasks requiring \emph{systematic compositionality}, where multiple atomic concepts must be combined to produce a correct output. Such tasks are particularly challenging to evaluate exhaustively, as the space of possible concept combinations grows combinatorially.

\paragraph{Limitations and Future work}
We deliberately restrict to a single geometric measurement calculated on simple, easily derived atomic concept representations. This minimal setup makes the analysis transparent and suggests that representation geometry contains useful information about compositional performance. Future work can explore richer ways to identify and analyze concepts, such as SAE features, causal features, or hierarchical manifold structures~\citep{elhage2022toy,cunningham2023sparse,ArchetypalSAE,chou2025featurelearninglazyrichdichotomy,melanipoincare}. Our analysis can also extend to reasoning models, which may accumulate interference across reasoning steps.

\section{Acknowledgements}

This work was enabled in part by a gift from the
Chan Zuckerberg Initiative Foundation to establish
the Kempner Institute for the Study of Natural and
Artificial Intelligence. IL and NS are supported by a grant from Coefficient Giving and the Berkeley Existential Research Institute (BERI).
We thank Melanie Weber, Thomas Fel, Joshua Batson, Annabelle Michael Carrell, Aaron Mueller, Yonatan Belinkov, Jing Huang, Victoria R. Li, 
Ekdeep Lubana, David Klindt, Sanchit Ahuja, SueYeon Chung, Ekaterina Shutova, Sebastian Ruder, Hadas Orgad, Sweta Karlekar, Apoorv Khandelwal, Michael Lepori, Tianze Hua, Zhuonan Yang and other members of the LUNAR lab for helpful discussion and feedback on this work.

%For our demonstration, we explore compositional scenarios which all have existing inputs. But future work could efficiently search larger combinatorial spaces and generate \textit{de novo} inputs from arbitrary combinations of concepts, enabling more systematic adversarial data synthesis for language models. The efficiency of the search itself could be improved; our method finds the highest interference pairs from $n$ concept combinations with $O(n)$ vector multiplications, but a carefully-designed search could test combinations more selectively. Another gap is our exclusive focus on destructive interference, where overlap between active representations impairs independent recovery. Correlated features can also exhibit \textit{constructive} interference, where overlap supports recovery~\citep{constructive_interference_prieto2026data}. Future work should also extend our framework to reasoning models, where failures along reasoning chains may additionally reflect the accumulation and propagation of interference across steps. 

%%%%%%%%%%%%%%%%%%%%%%%%%%%%%%%%%%%%%%%%%%%%%%%%%%%%%%%%%%%%
\bibliography{bib-noslop}
\bibliographystyle{iclr2026_conference}

\appendix

\clearpage

\section{Recovery bounds related to local cumulative coherence}
\label{sec:theory_apx}

This section expands on our claim that robust recovery bounds can be controlled by local cumulative coherence. Specifically, we want to recover an estimate $\hat{z}$ for an $m$-dimensional $k$-sparse feature vector $z$ from noisy conditions. Robust recovery guarantees provide high-probability bounds on the recovery error $\| \hat{z} - z \|_2$.

\subsection{Robust compressed sensing}

The specific robust compressed sensing bound provided by \citet{adcock2021oracle} limits the number of measurements required to guarantee a high probability of exact recovery. As is common in the superposition literature, we treat the hidden dimension $d$ as the number of sampled measurements. In robust compressed sensing, we assume that rather than access to the encoded $Az$, we instead observe the noisy representation $Az+\epsilon$, where the noise is bounded by some constant $\|\epsilon\| < \eta$. Therefore, the relevant bound can be paraphrased as claiming that we can guarantee that the recovery error on our estimated $\hat{z}$ from a noisy encoding  satisfies,
\begin{equation}
    \| z - \hat{z} \|_2 \leq (c_1 + c_2 \sqrt{k}) \eta,
\end{equation}
for constants $c_1, c_2$. 

 \citet{adcock2021oracle} guarantee this robust recovery  with probability $1-\varepsilon$ if the hidden dimension $d$ is at least,
\begin{equation}
    d \gtrsim  \Theta (\text{supp}(z),A)\log^2(m/\varepsilon),
    \label{eq:noisy_bound}
\end{equation}
where we define the interaction term as a positive real number,
\begin{equation}
    \Theta (\text{supp}(z),A) = \| A_{\mathcal{S}}^\top A_{\text{supp}(z)}  \|_{\infty \rightarrow \infty}
\end{equation}

In Equation \ref{eq:cum_coherence}, we refer to the bound in Equation \ref{eq:noisy_bound}  as effectively controlled by mutual coherence in the salient support,
\begin{align}
\alpha(\mathcal{S}) &=
\max_{i\in \mathcal{S}} \sum_{\substack{j\in \mathcal{S}}}
\left|\cos{(a_i, a_j)}\right|\\
&\geq \max_{i\in \mathcal{S}}
\sum_{\substack{j\in \textrm{supp}(z)}}
\left|\cos{(a_i, a_j)}\right|\\
&= \Theta (\text{supp}(z),A).
\end{align}
Therefore we can satisfy Equation \ref{eq:noisy_bound} if hidden dimension $d$ is at least,
\begin{align}
 d  \gtrsim  \alpha(\mathcal{S}) \log^2(m/\varepsilon).
\end{align}

\subsection{Robust linear compressed sensing}

We now assume the \textit{linear} compressed sensing setting of  \citet{LRH_garg}, which weakens the lossless recovery guarantee from compressed sensing to \textit{near-lossless} recovery by accepting the Linear Representation Hypothesis (LRH). Under the LRH, \citet{LRH_garg} showed that a linear decoder of dimension $d$ can still recover a number of features exponential in $d$ up to a small constant error $\epsilon$. 

In theory, the number of $k$-sparse features that can be linearly accessible is exponential in $d$ \citep{LRH_garg}---even if the encoded features are linearly correlated. However, the theory only guarantees that this near-lossless decoding \textit{exists}, not that it is applied by the neural network. We argue that if the neural network's next layer does linearly decode the representation, it may not reflect the theoretical ideal probe dictionary.  We will treat the LLM's linear encoding as explicit in its activations, but treat its linear decoding as implicit. We assume this implicit linear decoding to be a perturbation of the ideal decoding, leading to the following argument.

% We work in the linear representation setting of \citet{LRH_garg}. 
Let
${\enc} \in \mathbb{R}^{d\times m}$ be the representation matrix (columns $a_1,\dots,a_m$) and
${\dec} \in \mathbb{R}^{d\times m}$ be the probe matrix (columns $b_1,\dots,b_m$).
Given a feature vector $z \in [-1,1]^m$ with $k$ nonzero features, the decoded estimate is
$$\hat z = {\dec}^\top {\enc} z.$$
Due to the results of \citet{LRH_garg}, we can assume that for all $k$-sparse $z$,
$$\|\hat z - z\|_\infty \le \varepsilon.$$
Now assume that the implicit linear decoding during downstream processing, $\tilde{\dec}$, is a perturbation of the ideal probe ${\dec}$, specifically $\tilde{\dec} = {\dec} + \Delta {\dec}$ where 
$$\|\Delta {\dec}\|_{2\to\infty} \;\le\; \eta.$$
We will show that this perturbation has its largest impact on recovery error when the sparse active features have highly correlated encodings.

% ask: how does the new error $\|\tilde{\dec}^\top {\enc} z - z\|_\infty$ depend on correlations between features in ${\enc}$?

\begin{theorem}[Perturbation sensitivity bounded by correlations in ${\enc}_{\mathcal{S}}$]
\label{thm:sensitivity} Let $\tilde{\dec} = {\dec} + \Delta {\dec}$, and suppose each probe vector changes by at most $\eta$:
$$\|\Delta b_i\|_2 \le \eta \quad\text{for all } i.$$
Let $z$ be a $k$-sparse feature vector with support $\mathrm{supp}(z)$. 
Then
\begin{align}
\|\tilde{\dec}^\top {\enc} z - z\|_\infty
\;\le\; \varepsilon \;+\; \eta\, k\, \|{\enc}_{\mathrm{supp}(z)} \|_2.
\end{align}
In particular, if the product $\langle a_i,a_j\rangle$ for  $i,j \in \mathcal{S}$ is large and positive, that feature pair increases the perturbation sensitivity on this input.

\begin{proof}
We decompose the new error into the original decoding error plus the effect of changing the probe:
$$\tilde{\dec}^\top {\enc} z - z
= ({\dec}^\top {\enc} z - z) + \Delta {\dec}^\top {\enc} z.$$
By assumption, the first term has $\|{\dec}^\top {\enc} z - z\|_\infty \le \varepsilon$, so we only need to bound $\|\Delta {\dec}^\top {\enc} z\|_\infty$.
The $i$-th coordinate of this perturbation is bounded using Cauchy–Schwarz and the per-column bound $\|\Delta b_i\|_2 \le \eta$,
$$|(\Delta {\dec}^\top {\enc} z)_i|
\;\le\;
\|\Delta b_i\|_2 \,\|{\enc} z\|_2
\;\le\; \eta\,\|{\enc} z\|_2.$$

% Taking the maximum over all coordinates $i$, we use the triangle inequality to bound the recovery error,
% \begin{align}
%     \|\tilde{\dec}^\top {\enc} z - z\|_\infty
% &= \|({\dec}^\top {\enc} z - z) + \Delta {\dec}^\top {\enc} z\|_\infty \;\\ &\le\; \|{\dec}^\top {\enc} z - z\|_\infty + \|\Delta {\dec}^\top {\enc} z\|_\infty\\
% &\;\le\; \varepsilon \;+\; \eta\,\sqrt{\sum_{i \in [m]} \; \sum_{j \in [m]}  z_{i} z_j \langle a_i,a_j\rangle }\\
% \end{align}

Since $z$ is supported on $\mathrm{supp}(z)$, we can express the above using only the nonzero features and write ${\enc} z = {\enc}_{\mathrm{supp}(z)} z$, where $A_{\mathrm{supp}(z)}$ is the restriction of $A$ to indices in ${\mathrm{supp}(z)}$. Using Cauchy–Schwarz and the $k$-sparsity constraint on $z$,
$$|(\Delta {\dec}^\top {\enc} z)_i|
\;\le\; \eta\, \|{\enc}_{\mathrm{supp}(z)} z\|_2 \;\le\; \eta\, \|{\enc}_{\mathrm{supp}(z)}\|_2 \; \|z\|_2  \;\le\; \eta\, k\, \|{\enc}_{\mathrm{supp}(z)} \|_2 .$$

Taking the maximum over all coordinates $i$,
\begin{align}
\|\tilde{\dec}^\top {\enc} z - z\|_\infty
&\;\le\; \varepsilon \;+\; \eta\, k\, \|{\enc}_{\mathrm{supp}(z)} \|_2.
\end{align}
\end{proof}
\end{theorem}

Under Theorem \ref{thm:sensitivity},  deviations from an ideal decoder can damage feature recovery more on inputs when encodings of the active salient support are correlated. If an LLM implicitly processes feature encodings with a flawed decoder, it will therefore make more mistakes when composing non-orthogonal encodings. 

\textit{Note that the quantity of interest in our robust recovery bound differs depending on whether the model's intrinsic decoding is linear or nonlinear.} If we assume the model decodes nonlinearly, as in compressed sensing, then the quantity of interest is cumulative coherence in the salient support. If we assume the model uses linear decoding, then the quantity of interest is a simple norm on the active support---a subset of the salient support, and one that is not identifiable without analyzing specific input instantiating the scenario. Without studying specific inputs, we remain limited to measurements on the salient support, not the active subset. 

Either way, cosine similarity within the salient support causes destructive interference.

\section{Alternative Metrics to Capture Compositional Interference}
\label{sec:alternative_ci_metrics}

In the paper, we use cumulative coherence derived from~\citet{adcock2021oracle} and measure this quantity. We consider several alternative geometric metrics for estimating compositional interference among the salient features of a composition. Let \(\mathcal{S}(C)\) denote the salient feature support, and let \(a_i\) be the representation vector associated with feature \(i\). All metrics below are computed over pairwise similarities among active feature directions.

\paragraph{Minimum similarity.}
The minimum similarity captures the least aligned pair of salient features:
\begin{equation}
\label{eq:ci_min_sim}
\ci_{\min}(C)
=
\min_{\substack{i,j\in \mathcal{S}(C)\\ i\neq j}}
|\cos(a_i,a_j)| .
\end{equation}

\paragraph{Maximum similarity.}
The maximum similarity captures the most aligned pair of salient features:
\begin{equation}
\label{eq:ci_max_sim}
\ci_{\max}(C)
=
\max_{\substack{i,j\in \mathcal{S}(C)\\ i\neq j}}
|\cos(a_i,a_j)| .
\end{equation}

\paragraph{Mean similarity.}
The mean similarity averages pairwise alignment across all salient feature pairs:
\begin{equation}
\label{eq:ci_mean_sim}
\ci_{\mathrm{mean}}(C)
=
\frac{1}{|\mathcal{S}(C)|(|\mathcal{S}(C)|-1)}
\sum_{\substack{i,j\in \mathcal{S}(c)\\ i\neq j}}
|\cos(a_i,a_j)| .
\end{equation}
When \(|\mathcal{S}(C)|=2\), different CI aggregation metrics reduce to the same pairwise comparison and therefore yield the same correlational result. When \(|\mathcal{S}(C)|>2\), however, these metrics can diverge because they aggregate multiple pairwise interactions differently. Therefore, whenever examples contain more than two salient features, we additionally report results for alternative aggregation metrics to enable comparison across definitions of interference (Figures~\ref{fig:appendix_scan_other_metrics_max_cumulative},~\ref{fig:appendix_scan_other_metrics_max_noncumulative},~\ref{fig:appendix_scan_other_metrics_min_cumulative},~\ref{fig:appendix_scan_other_metrics_min_noncumulative},~\ref{fig:appendix_scan_other_metrics_mean_cumulative},~\ref{fig:appendix_scan_other_metrics_mean_noncumulative},~\ref{fig:appendix_multilingual_other_metrics},~\ref{fig:appendix_multihop_other_metrics_subspace_subspace}, and Table~\ref{tab:multilingual_subspace_subspace_language_corr_other_metrics}).

Across these comparisons, most similarity-based metrics remain predictive in most settings. Since the alternative metrics are partly intercorrelated, this suggests that the central signal comes from angular similarity among concept representations. Nevertheless, empirical results in the paper show that the cumulative coherence metric derived from the bounds in Section~\ref{sec:ci_theory} is the most stable metric to measure compositional interference across all scenarios.
\clearpage
\section{Cluster Mean Centering}
\label{sec:appendix_mean_centering}

We expand on Section~\ref{sec:ci_theory}, ``Accounting for multiscale structure,'' to provide more mathematical intuition and details.

For an input \(x \in \mathcal{X}(c)\), let \(C\) denote the active concept set, let \(\mathcal{S}(C)\) denote its salient support, and let \(A_{\mathcal{S}(C)} = \{a_i : i \in \mathcal{S}(C)\}\) denote the corresponding salient feature encodings. To provide a simple intuition, we consider the setting in which each salient feature is represented by a single linear direction, and we use the linear representation hypothesis to model residual stream activations. The empirical estimate \(\tilde a_i\) of a salient feature encoding may contain not only the direction of interest, but also background components induced by prompt format, task family, language, or other contextual structure. We write
\[
\tilde a_i
=
z_i a_i
+
\sum_{u \in \mathcal{B}_i} z_u^{(i)} a_u,
\]
where \(z_i a_i\) is the salient feature contribution we aim to isolate, \(z_i\) is its activation coefficient, and \(\mathcal{B}_i\) indexes other background directions mixed into the empirical estimate of feature \(i\). Then the raw inner product between two empirical estimates \(\tilde a_i\) and \(\tilde a_j\), for \(i,j \in \mathcal{S}(c)\), expands as
\begin{align}
\langle \tilde a_i, \tilde a_j \rangle
&=
z_i z_j \langle a_i, a_j \rangle
+
z_i \sum_{\ell \in \mathcal{B}_j} z_\ell^{(j)}
\langle a_i, a_\ell \rangle
\nonumber \\
&\quad+
z_j \sum_{u \in \mathcal{B}_i} z_u^{(i)}
\langle a_u, a_j \rangle
\nonumber \\
&\quad+
\sum_{u \in \mathcal{B}_i}
\sum_{\ell \in \mathcal{B}_j}
z_u^{(i)} z_\ell^{(j)}
\langle a_u, a_\ell \rangle .
\end{align}
The first term is the salient feature interaction we aim to measure, while the remaining terms arise from background structure. Thus, raw inner products between empirical feature estimates can be dominated by shared background structure rather than by the local interaction of interest.

To mitigate this effect, we apply cluster mean-centering at the level of residual representations before estimating salient feature encodings. Let \(\gamma(x)\) denote the background cluster associated with example \(x\), and let \(\mu_{\gamma(x)}\) be the empirical mean residual representation of examples in that cluster. We define the centered residual representation as
\[
h_c(x) = h(x) - \mu_{\gamma(x)}.
\]
We then use these centered residuals to estimate the salient feature encodings \(A_{\mathcal{S}(C)}\) used for angle computation. Mean-centering reduces dominant cluster-level structure and isolates residual variation that more directly reflects interactions between salient feature directions.

In this work, we identify background clusters by visualizing empirical representations with SVD and observing that they group by prompt type, task family, language, or other contextual factors. We then center each representation by the empirical mean of its corresponding cluster. Automatically discovering these background clusters is an important direction for future work.

Beyond the mathematical intuition, empirical observations in Appendix Figures~\ref{fig:multiscale_discrete_ortho_bins},~\ref{fig:appendix_multihop_other_results},~\ref{fig:appendix_multihop_mean_centering_simpson}, and~\ref{fig:appendix_multihop_mean_centering_subsets} further motivate the need for cluster mean-centering.
\begin{figure}[ht]
    \centering
    \includegraphics[width=1.0\columnwidth]{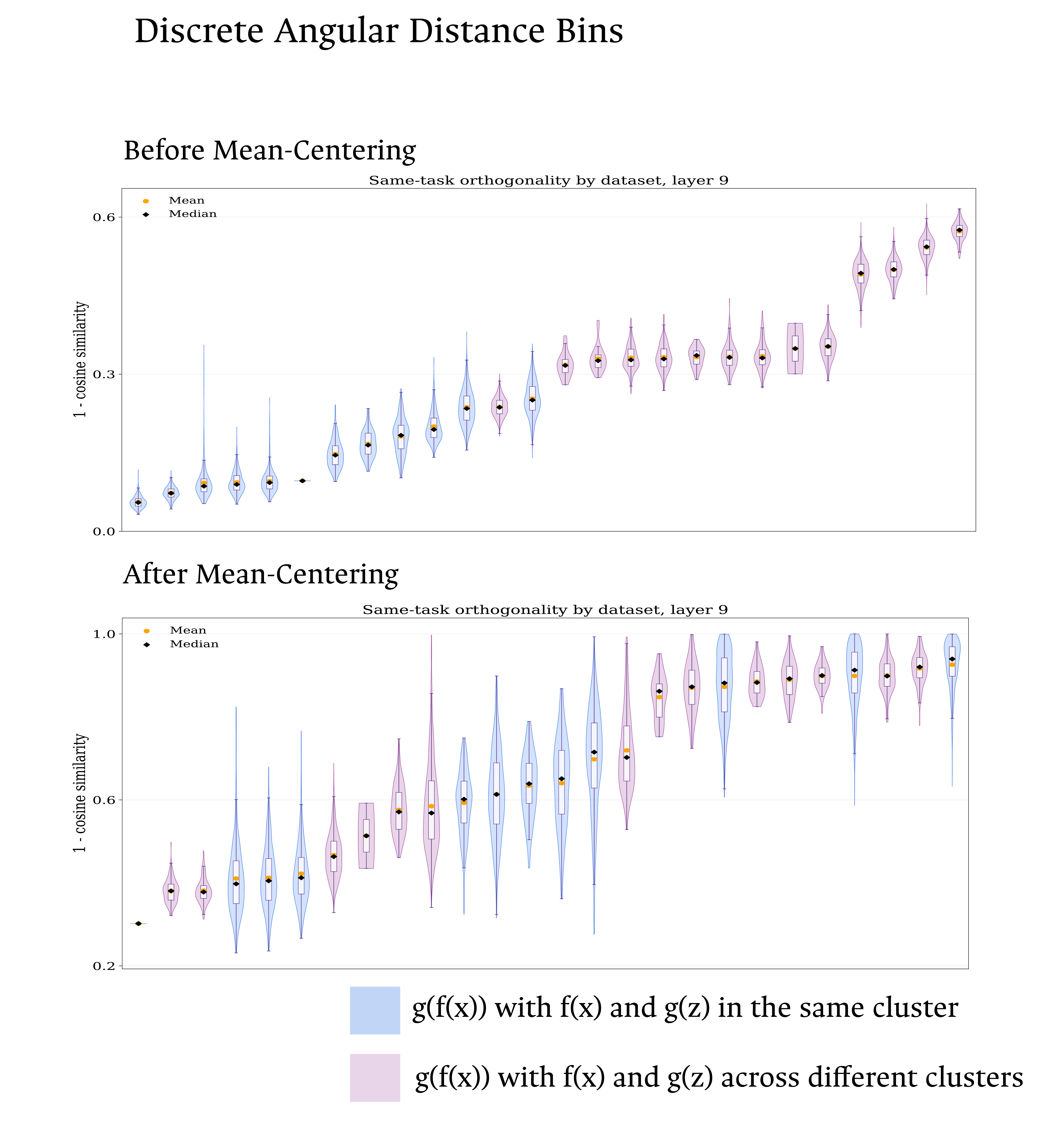}
    \caption{Angle distributions of concept interactions in the multi-hop dataset before and after mean-centering. The top panel shows that, without mean-centering, angles between interacting concepts from different clusters are dominated by global clustering structure. After mean-centering, shown in the bottom panel, this global effect is reduced, allowing us to better capture local interactions between concepts.}
    \label{fig:multiscale_discrete_ortho_bins}
\end{figure}

\clearpage

%%%%%%%%%%%%%%% SCAN %%%%%%%%%%%%%%%%%
\section{Additional SCAN Experiments and Analyses}
\label{sec:appendix_scan}

This section expands on the controlled SCAN experiments in
Section~\ref{sec:scan}. We first describe the training and evaluation protocol,
then examine how representation geometry changes across training regimes,
evaluate whether CI predicts item-level failures, and finally test whether the
observed signal can be explained by command length or training-data statistics.

\subsection{Experimental Setup}

\paragraph{Training data and model configurations.}
We train decoder-only Transformers on the SCAN benchmark while varying both
training coverage and model capacity. Specifically, we use
\(4\%, 8\%, 16\%, 32\%, 64\%,\) or \(80\%\) of the distinct commands in the
full SCAN training split, while keeping the total number of training examples
fixed. We train models with hidden dimension
\(d \in \{8,12,32,64\}\). Every model has 10 Transformer layers, 4 attention
heads, and a feed-forward dimension of \(4d\). We trained a range of models to test whether CI is a consistent predictor for models with different sizes and compositional performance.

We optimize with Adam using a learning rate of \(10^{-3}\) and a batch size of
256. Training runs for at most 3{,}000 epochs and stops when
development-set exact-match accuracy has not improved for 200 epochs. The
development set comprises 10\% of the held-out test split. The main experiments are done using random seed 42.

Input sequences take the form
\texttt{[BOS] input [SEP] output [EOS]}. We compute cross-entropy loss only on
the output tokens. At evaluation time, we use greedy autoregressive decoding
and report full-sequence exact-match accuracy.

\paragraph{Representations and model-specific evaluation.}
\label{sec:scan_within_model_control}
For each primitive concept, we estimate its representation by averaging its
residual-stream activations across training contexts containing that concept,
as described in Section~\ref{sec:scan}. When computing CI for a test example, we form the active concept set from its unique primitives, deduplicating repeated occurrences so that each primitive contributes only once. For each model, the layer used to
compute CI is selected on the development set. Because model capacity and
training coverage change the scale and distribution of CI, we do not compare
raw CI values across models. Instead, we ask whether CI ranks examples by
difficulty \emph{within each model}. This model-specific evaluation controls
for large differences in capacity, coverage, and overall accuracy.

\paragraph{Evaluation regime.}
\label{sec:scan_nonsaturated}
Our principal conclusions concern non-extreme settings with
\(0.2 < \mathrm{accuracy} < 0.99\). Below this range, the model fails on almost
all compositions; above it, fewer than 1\% of examples are failures. In either
case, cumulative curves become nearly flat and can change abruptly around the
few minority-class examples, while PR-AUC and its statistical significance
cannot be estimated reliably. We therefore report these settings for
completeness but restrict our main conclusions and significance tests to
non-saturated regimes. Within every evaluated non-saturated setting with
failure rate greater than 0.01, CI significantly outperforms the failure-rate
PR-AUC baseline.

\subsection{Representation Geometry Changes Across Training Regimes}

We first verify that the controlled SCAN setting induces distinct internal
geometries. Figure~\ref{fig:appendix_scan_heatmaps} shows the pairwise angular
distances between atomic concept representations. Both training coverage and
hidden dimension change the organization of the representation space, yielding
different patterns of alignment and potential interference across models.

\begin{figure}[htbp]
    \centering
    \includegraphics[width=1.0\columnwidth]{figs/appendix/appendix_scan_heatmaps.pdf}
    \caption{Heatmaps of pairwise angular distances between atomic SCAN concept
    representations across training coverage and model capacity. Columns show
    coverage ratios of \(4\%, 8\%, 16\%, 32\%,\) and \(64\%\); rows show hidden
    dimensions \(d\in\{8,12,32,64\}\). For each model, we visualize the layer
    selected on the development set.}
    \label{fig:appendix_scan_heatmaps}
\end{figure}

\begin{figure}[htbp]
    \centering
    \includegraphics[width=1.0\columnwidth]{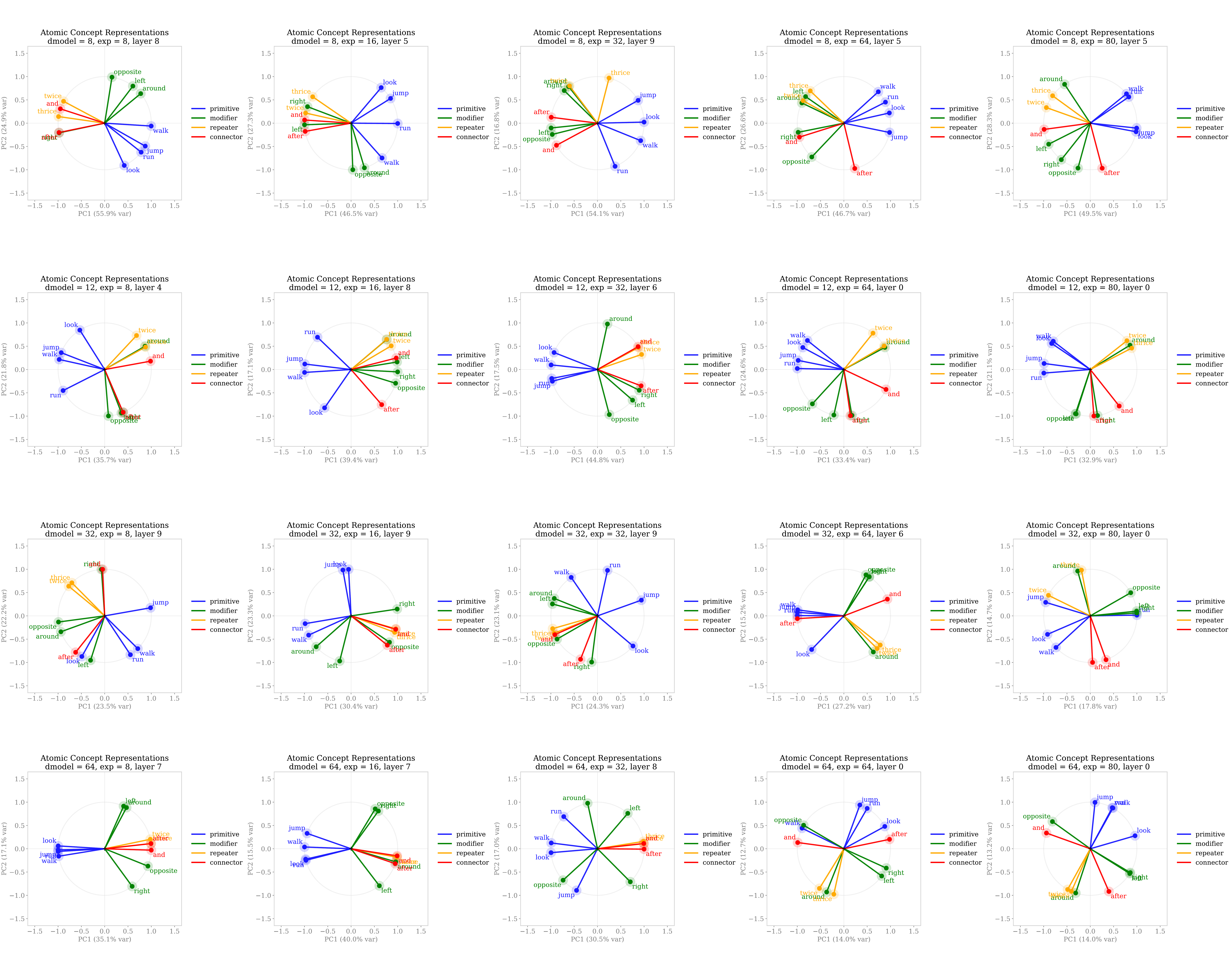}
    \caption{PCA projections of atomic SCAN concept representations across
    training coverage and model capacity. Columns show coverage ratios of
    \(4\%, 8\%, 16\%, 32\%,\) and \(64\%\); rows show hidden dimensions
    \(d\in\{8,12,32,64\}\). For each model, we visualize the layer selected on
    the development set. PCA is used only for visualization; CI is computed in
    the original representation space.}
    \label{fig:appendix_scan_features}
\end{figure}

Figure~\ref{fig:appendix_scan_features} provides a complementary PCA view. These visualizations motivate our
model-specific analysis: each model learns a distinct geometry, and we test
whether that geometry predicts the failures of that same model.

\clearpage

\subsection{CI Predicts Item-Level Compositional Failures}

We next test the central prediction that examples with greater interference
are more likely to fail. Table~\ref{tab:appendix_scan_acc}
first reports the overall accuracy of each model and Table~\ref{tab:appendix_scan_pbc} reports the point-biserial
correlation between CI and binary correctness. In nearly all non-saturated
settings, the correlation is significantly negative: examples with higher CI
are less likely to be answered correctly. The weaker and occasionally positive
correlations occur primarily in near-saturated regimes, where only a very small
number of failures remain.

\begin{table}[htbp]
\centering
\small
\begin{tabular}{lcccc}
\toprule
\textbf{Cov.} & \(d=8\) & \(d=12\) & \(d=32\) & \(d=64\) \\
\midrule
\(4\%\)  & 0.169 & 0.703 & 0.794 & 0.792 \\
\(8\%\)  & 0.468 & 0.963 & 0.957 & 0.957 \\
\(16\%\) & 0.767 & 0.990 & 0.983 & 0.985 \\
\(32\%\) & 0.950 & 0.995 & 0.997 & 0.992 \\
\(64\%\) & 0.947 & 0.995 & 0.999 & 0.998 \\
\(80\%\) & 0.960 & 0.996 & 0.998 & 0.999 \\
\bottomrule
\end{tabular}
\caption{SCAN exact-match accuracy across training coverage ratios
and model dimensions.}
\label{tab:appendix_scan_acc}
\end{table}

\begin{table}[htbp]
\centering
\small
\begin{tabular}{lcccc}
\toprule
\textbf{Cov.} & \(d=8\) & \(d=12\) & \(d=32\) & \(d=64\) \\
\midrule
\(4\%\)  & \(-0.067^{***}\) & \(-0.283^{***}\) & \(-0.183^{***}\) & \(-0.219^{***}\) \\
\(8\%\)  & \(-0.121^{***}\) & \(-0.160^{***}\) & \(-0.166^{***}\) & \(-0.176^{***}\) \\
\(16\%\) & \(-0.075^{***}\) & \(-0.112^{***}\) & \(-0.045^{***}\) & \(-0.139^{***}\) \\
\(32\%\) & \(-0.140^{***}\) & \(-0.049^{***}\) & \(-0.073^{***}\) & \(-0.038^{***}\) \\
\(64\%\) & \(-0.181^{***}\) & \(0.030^{*}\) & \(-0.015\) & \(-0.046^{***}\) \\
\(80\%\) & \(-0.078^{***}\) & \(0.051^{**}\) & \(0.033^{*}\) & \(-0.042^{*}\) \\
\bottomrule
\end{tabular}
\caption{Point-biserial correlation between CI and binary correctness
on SCAN across training coverage ratios and model dimensions. Because the
original analysis used \(1-\mathrm{CI}\), we reverse the signs when reporting
correlations with CI. Except in near-saturated regimes, the correlations are
consistently negative. Significance: \(^{*}p<0.05\),
\(^{**}p<0.01\), and \(^{***}p<10^{-3}\).}
\label{tab:appendix_scan_pbc}
\end{table}

\begin{figure}[htbp]
    \centering
    \includegraphics[width=1.0\columnwidth]{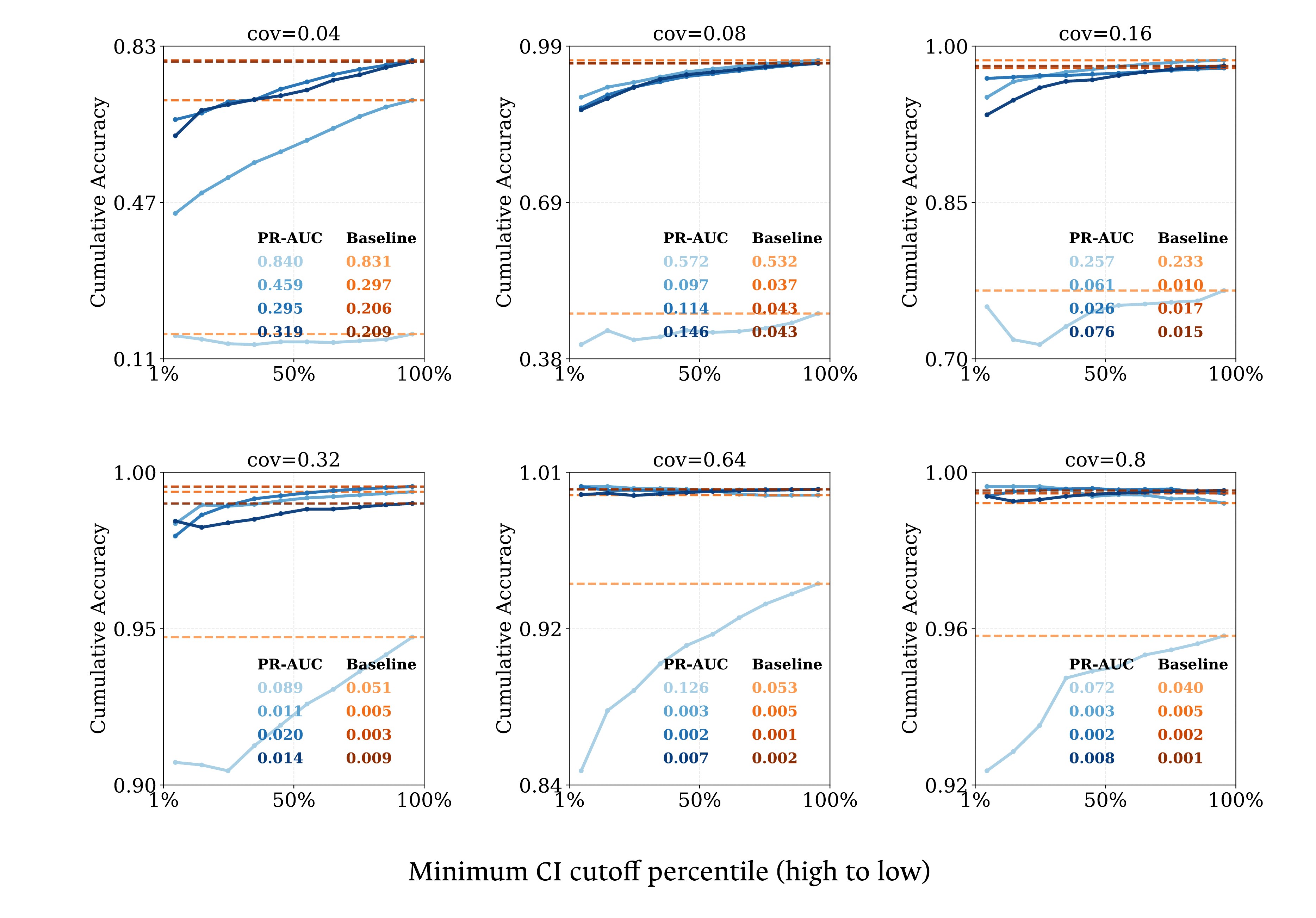}
    \caption{\textbf{Cumulative accuracy across the CI ranking on SCAN.}
    For each model dimension and training coverage ratio, test examples are ordered
    from highest CI (predicted most difficult) to lowest CI (predicted easiest).
    At each cutoff, we calculate the average accuracy over all examples from the
    start of the ranking through that cutoff. Across settings, the curves begin
    with low accuracy on the highest-interference examples and increase toward each
    model's overall mean accuracy as lower-interference examples are included,
    showing that CI successfully ranks concept combinations by difficulty.}
    \label{fig:appendix_scan_cumulative}
\end{figure}

The cumulative curves in Figure~\ref{fig:appendix_scan_cumulative} evaluate CI as a ranking signal. We rank concept combinations by predicted difficulty, from highest to lowest CI, and measure average accuracy over increasingly large top-ranked subsets. High-CI subsets are generally the most difficult, while their accuracy approaches the model-wide baseline as more examples are included. Thus, CI can be used to construct targeted challenge sets without first evaluating every composed input.

\begin{figure}[htbp]
    \centering
    \includegraphics[width=1.0\columnwidth]{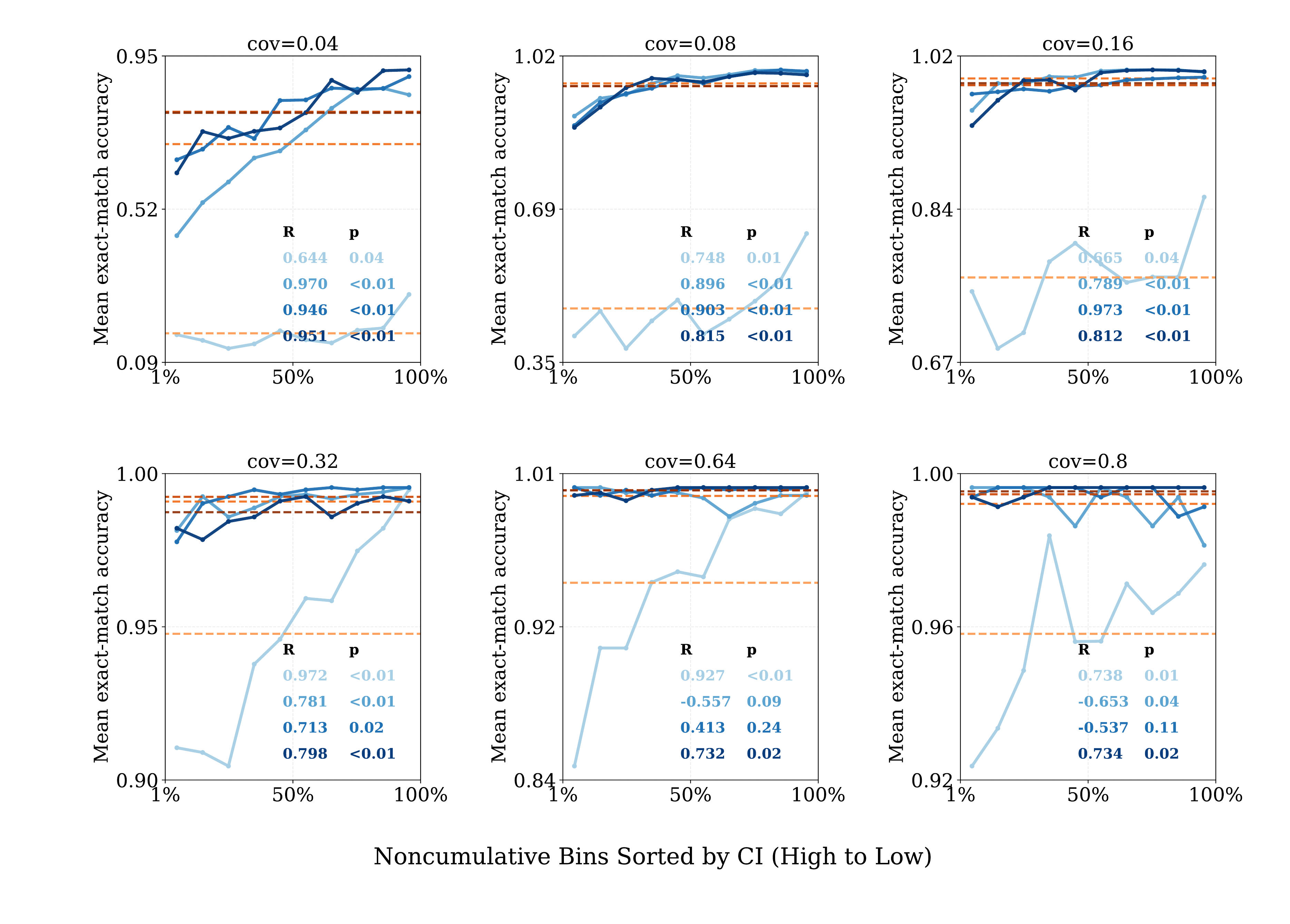}
    \caption{\textbf{Noncumulative accuracy across the CI ranking on SCAN.}
    For each model dimension and training coverage ratio, test examples are ordered
    from highest CI (predicted most difficult) to lowest CI (predicted easiest).
    For each pair of successive cutoffs, we calculate the average accuracy using
    only the examples that fall between those cutoffs; examples preceding the
    interval are not included. Across settings, accuracy is generally lower in
    the high-interference intervals and higher in the low-interference intervals,
    showing that CI successfully ranks concept combinations by difficulty.}
    \label{fig:appendix_scan_noncumulative}
\end{figure}

The noncumulative curves in Figure~\ref{fig:appendix_scan_noncumulative} divide
the ranking into disjoint bins. Their broadly decreasing relationship between
CI and accuracy shows that the cumulative trend is not produced by only a
small group of extreme examples: difficulty varies throughout the ranking.

Table~\ref{tab:pr_auc_results} evaluates CI as a failure-detection
ranking using permutation tests and bootstrap confidence intervals. For each
model, we compute the PR-AUC obtained by ranking examples from high to low CI;
the random-ranking baseline equals the model's overall failure rate. To test
statistical significance, we repeatedly permute CI scores across examples while
holding the failure labels fixed and recompute PR-AUC. The permutation
\(p\)-value is the proportion of shuffled rankings whose PR-AUC is at least as
large as the observed value. We additionally bootstrap examples and report the
95\% percentile confidence interval for the improvement in PR-AUC over the
failure-rate baseline. Across every reported non-saturated setting, the
permutation test yields \(p<0.001\), and the confidence interval for improvement
lies entirely above zero. Together with the point-biserial correlations and the
cumulative and noncumulative curves, these results provide complementary
evidence that CI predicts each model's item-level failures.

\begin{figure}[htbp]
    \centering
    \includegraphics[width=1.0\columnwidth]{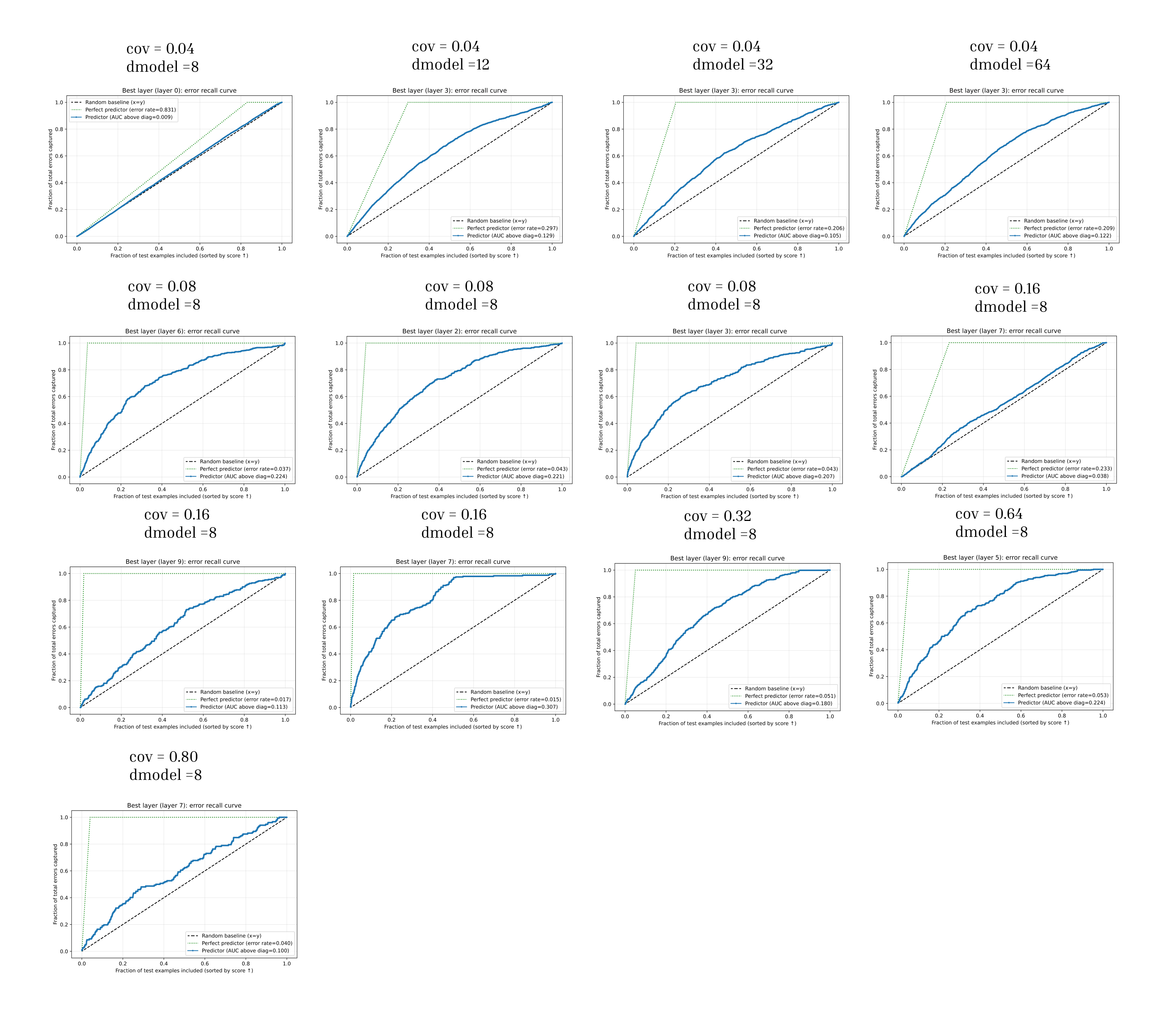}
    \caption{Error recovery under CI-based ranking across non-extreme SCAN settings. Examples are ranked from highest to lowest CI; the horizontal axis shows the fraction of examples inspected, and the vertical axis shows the fraction of all errors recovered. The diagonal \(y=x\) represents random ranking (AUC \(=0.5\)). All CI-based rankings achieve an AUC above this baseline, indicating that model errors are concentrated near the top of the ranking.}
    \label{fig:appendix_scan_error_recall}
\end{figure}

Figure~\ref{fig:appendix_scan_error_recall} reports a complementary top-\(k\)
error-recovery analysis. We rank examples from high to low CI and, for each
inspection budget \(k\), plot the fraction of all model errors recovered among
the top \(k\) examples against the fraction of examples inspected. Under a
random ranking, the expected recovery curve is \(y=x\), with an AUC of 0.5. The
CI-based rankings exceed this baseline for every evaluated model in the
non-extreme accuracy regime, showing that CI concentrates model errors near the
top of the ranking.

\begin{table}[htbp]
\centering
\small
\begin{tabular}{rrrrrrl}
\toprule
\(d\) & Cov. & Accuracy & Failure rate & PR-AUC & \(p_{\mathrm{perm}}\) & 95\% CI of improvement \\
\midrule
12 & 4\%  & 0.7032 & 0.2968 & 0.4585 & $<0.001$ & [0.1511, 0.1733] \\
32 & 4\%  & 0.7944 & 0.2056 & 0.2954 & $<0.001$ & [0.0795, 0.1006] \\
64 & 4\%  & 0.7915 & 0.2085 & 0.3193 & $<0.001$ & [0.0999, 0.1232] \\
8  & 8\%  & 0.4679 & 0.5321 & 0.5717 & $<0.001$ & [0.0318, 0.0479] \\
12 & 8\%  & 0.9629 & 0.0371 & 0.0970 & $<0.001$ & [0.0492, 0.0759] \\
32 & 8\%  & 0.9567 & 0.0433 & 0.1141 & $<0.001$ & [0.0582, 0.0888] \\
64 & 8\%  & 0.9570 & 0.0430 & 0.1456 & $<0.001$ & [0.0824, 0.1254] \\
8  & 16\% & 0.7665 & 0.2335 & 0.2567 & $<0.001$ & [0.0166, 0.0308] \\
32 & 16\% & 0.9829 & 0.0171 & 0.0260 & $<0.001$ & [0.0051, 0.0147] \\
64 & 16\% & 0.9851 & 0.0149 & 0.0763 & $<0.001$ & [0.0418, 0.0902] \\
8  & 32\% & 0.9495 & 0.0505 & 0.0888 & $<0.001$ & [0.0304, 0.0474] \\
\bottomrule
\end{tabular}
\caption{Failure-detection PR-AUC for CI at the development-selected layer.
Examples are ranked from high to low CI, with failure treated as the positive
class. The random-ranking PR-AUC baseline equals the model's overall failure
rate. The permutation \(p\)-value is computed by shuffling CI scores across
examples while holding failure labels fixed. The final column reports the 95\%
bootstrap confidence interval for the improvement in PR-AUC over the
failure-rate baseline. We omit extreme regimes in which the minority class is
too small for reliable estimation.}
\label{tab:pr_auc_results}
\end{table}

\clearpage

\subsection{Controlling for Command Complexity}
\label{sec:scan_length_control}

\begin{figure}[htbp]
    \centering
    \includegraphics[width=1.0\columnwidth]{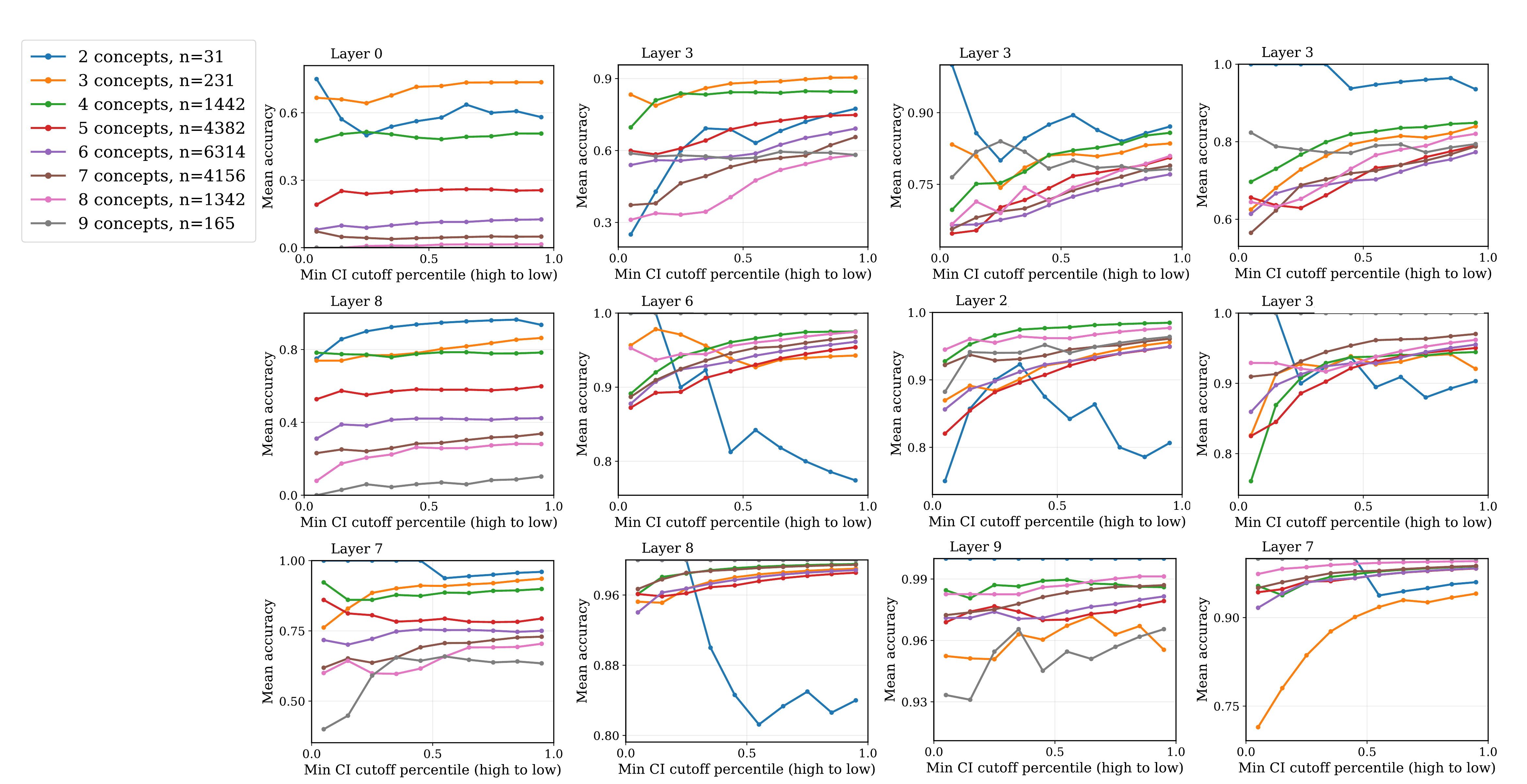}
    \caption{Cumulative accuracy curves within command-length groups for
    coverage ratios \(4\%, 8\%,\) and \(16\%\), with columns corresponding to
    \(d\in\{8,12,32,64\}\). The trend generally persists within length groups,
    but is weaker for the shortest, two-concept commands. Manual inspection
    suggests that many such failures reflect operator confusion---for example,
    mapping \texttt{jump twice} to three jumps---rather than interference among
    several simultaneously active components.}
    \label{fig:appendix_scan_bylen1}
\end{figure}

\begin{figure}[htbp]
    \centering
    \includegraphics[width=1.0\columnwidth]{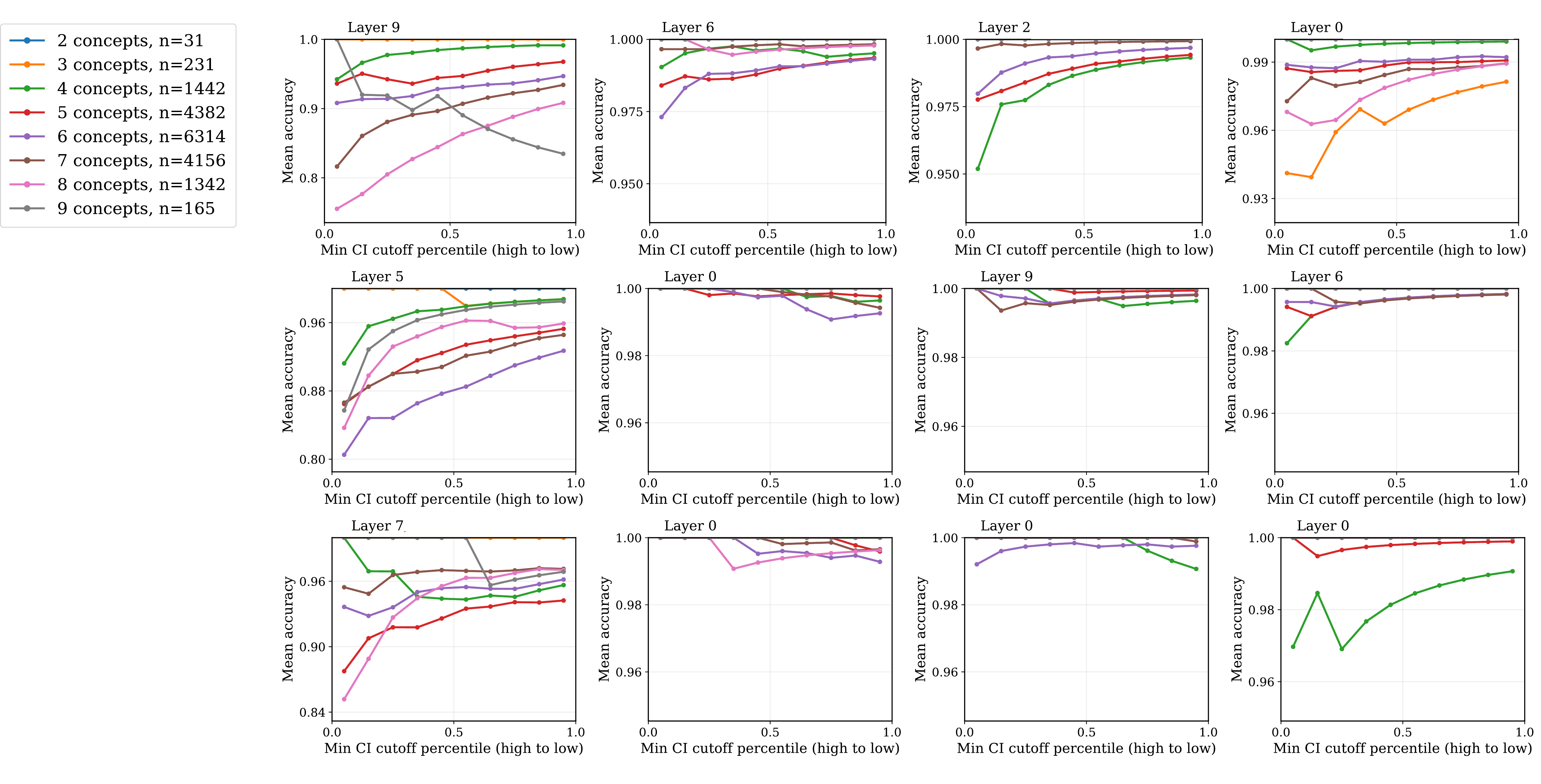}
    \caption{Cumulative accuracy curves within command-length groups for
    coverage ratios \(32\%, 64\%,\) and \(80\%\), with columns corresponding to
    \(d\in\{8,12,32,64\}\). Near-saturated settings are shown for completeness
    but are not used to support the main statistical conclusion.}
    \label{fig:appendix_scan_bylen2}
\end{figure}

One possible explanation for the preceding results is that CI merely tracks
command complexity: longer commands contain more primitives and are generally
more difficult. We therefore repeat the cumulative analysis separately within
groups containing the same number of primitive concepts. Figures
~\ref{fig:appendix_scan_bylen1} and~\ref{fig:appendix_scan_bylen2} show that the
relationship between CI and accuracy generally remains within these matched
groups. The main result therefore cannot be attributed solely to command
length, prompt type, or the overall number of primitives.

Figure~\ref{fig:appendix_scan_bylen2} reports the same control for the
higher-coverage regimes. Several of these models are near saturation, so their
curves are necessarily less stable; nevertheless, the non-saturated panels
continue to show that the CI-accuracy relationship is not simply a confound effect from command length.

\clearpage

\subsection{Robustness to Alternative Interference Metrics}

Our primary metric is cumulative coherence because it follows from the recovery
bound in Section~\ref{sec:theory}. To test whether the empirical signal depends
on this particular aggregation, we additionally compute the maximum, mean, and
minimum absolute pairwise cosine similarities defined in
Section~\ref{sec:alternative_ci_metrics}. For each alternative, we present the
cumulative ranking result followed immediately by the corresponding
noncumulative result.

\begin{figure}[htbp]
    \centering
    \includegraphics[width=1.0\columnwidth]{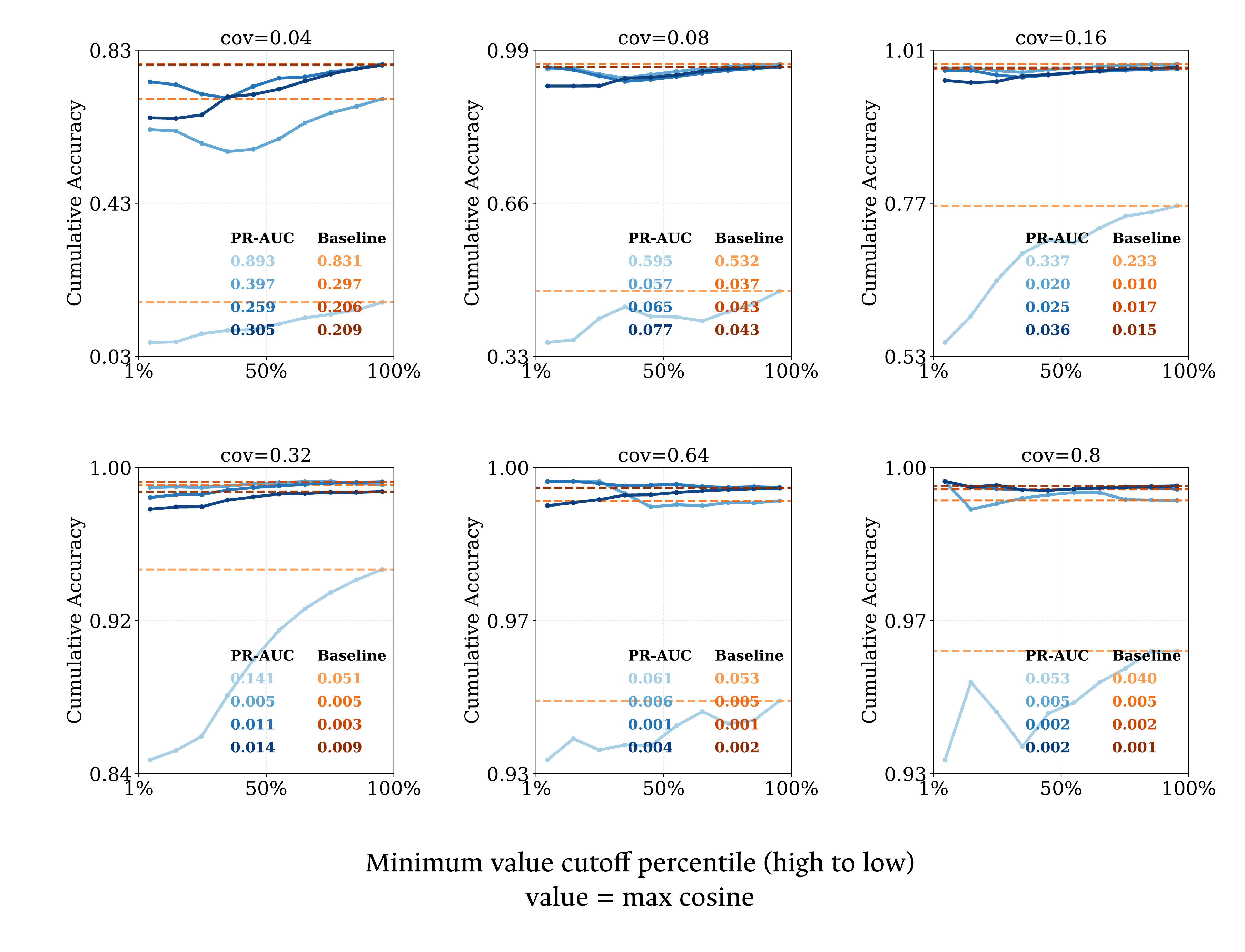}
    \caption{Cumulative SCAN accuracy curves when examples are ranked by maximum
    absolute pairwise cosine similarity.}
    \label{fig:appendix_scan_other_metrics_max_cumulative}
\end{figure}

\begin{figure}[htbp]
    \centering
    \includegraphics[width=1.0\columnwidth]{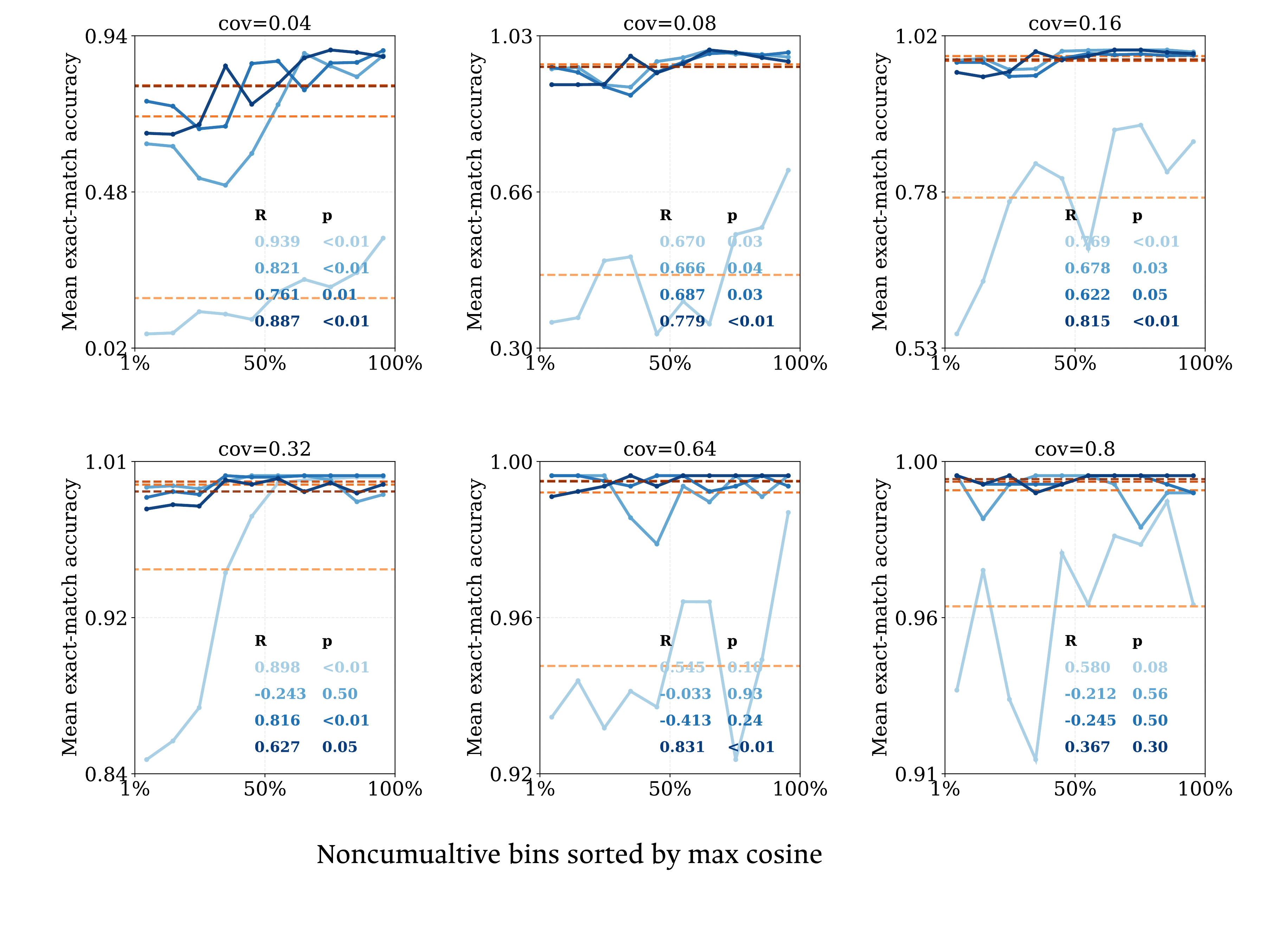}
    \caption{Noncumulative SCAN accuracy curves when examples are binned by
    maximum absolute pairwise cosine similarity.}
    \label{fig:appendix_scan_other_metrics_max_noncumulative}
\end{figure}

\paragraph{Maximum pairwise similarity.}
Figures~\ref{fig:appendix_scan_other_metrics_max_cumulative}
and~\ref{fig:appendix_scan_other_metrics_max_noncumulative} use the largest
pairwise cosine similarity among the concepts in each command. The metric asks
whether a single strongly aligned pair is sufficient to identify difficult
compositions.

\clearpage

\begin{figure}[htbp]
    \centering
    \includegraphics[width=1.0\columnwidth]{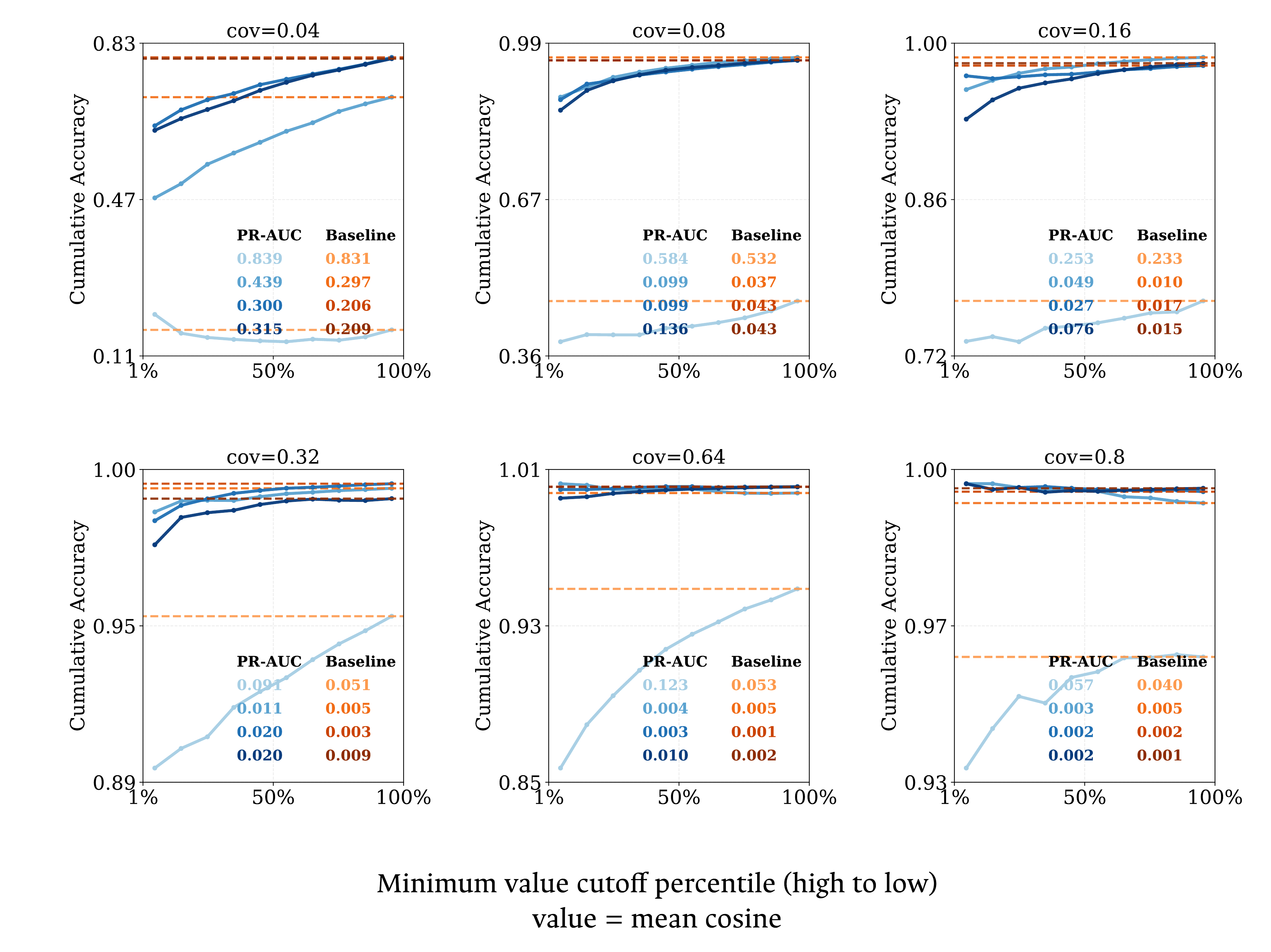}
    \caption{Cumulative SCAN accuracy curves when examples are ranked by mean
    absolute pairwise cosine similarity.}
    \label{fig:appendix_scan_other_metrics_mean_cumulative}
\end{figure}

\begin{figure}[htbp]
    \centering
    \includegraphics[width=1.0\columnwidth]{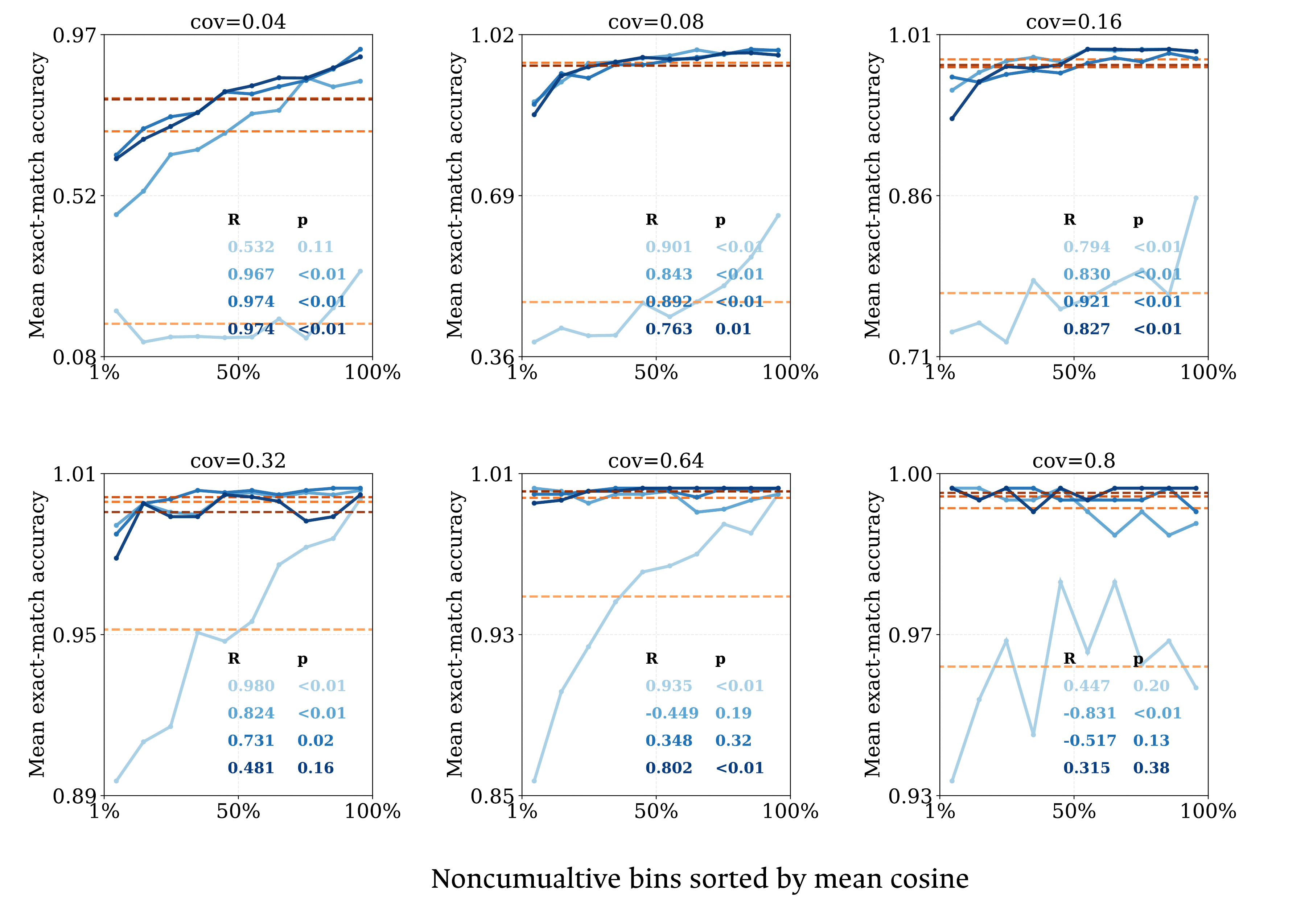}
    \caption{Noncumulative SCAN accuracy curves when examples are binned by mean
    absolute pairwise cosine similarity.}
    \label{fig:appendix_scan_other_metrics_mean_noncumulative}
\end{figure}

\paragraph{Mean pairwise similarity.}
Figures~\ref{fig:appendix_scan_other_metrics_mean_cumulative}
and~\ref{fig:appendix_scan_other_metrics_mean_noncumulative} average over all
pairwise interactions, testing whether the overall density of alignment is
predictive.

\clearpage

\begin{figure}[htbp]
    \centering
    \includegraphics[width=1.0\columnwidth]{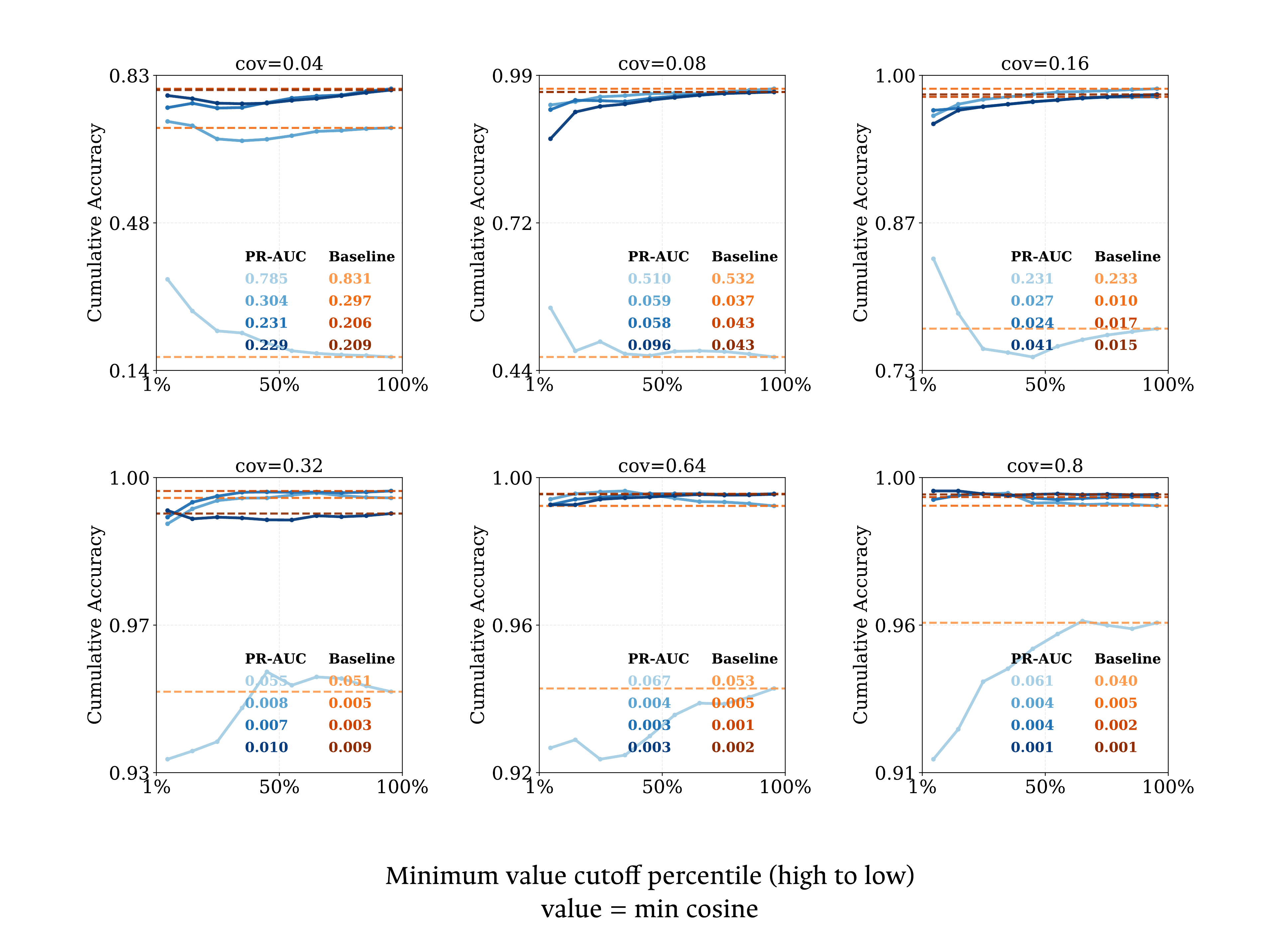}
    \caption{Cumulative SCAN accuracy curves when examples are ranked by minimum
    absolute pairwise cosine similarity.}
    \label{fig:appendix_scan_other_metrics_min_cumulative}
\end{figure}

\begin{figure}[htbp]
    \centering
    \includegraphics[width=1.0\columnwidth]{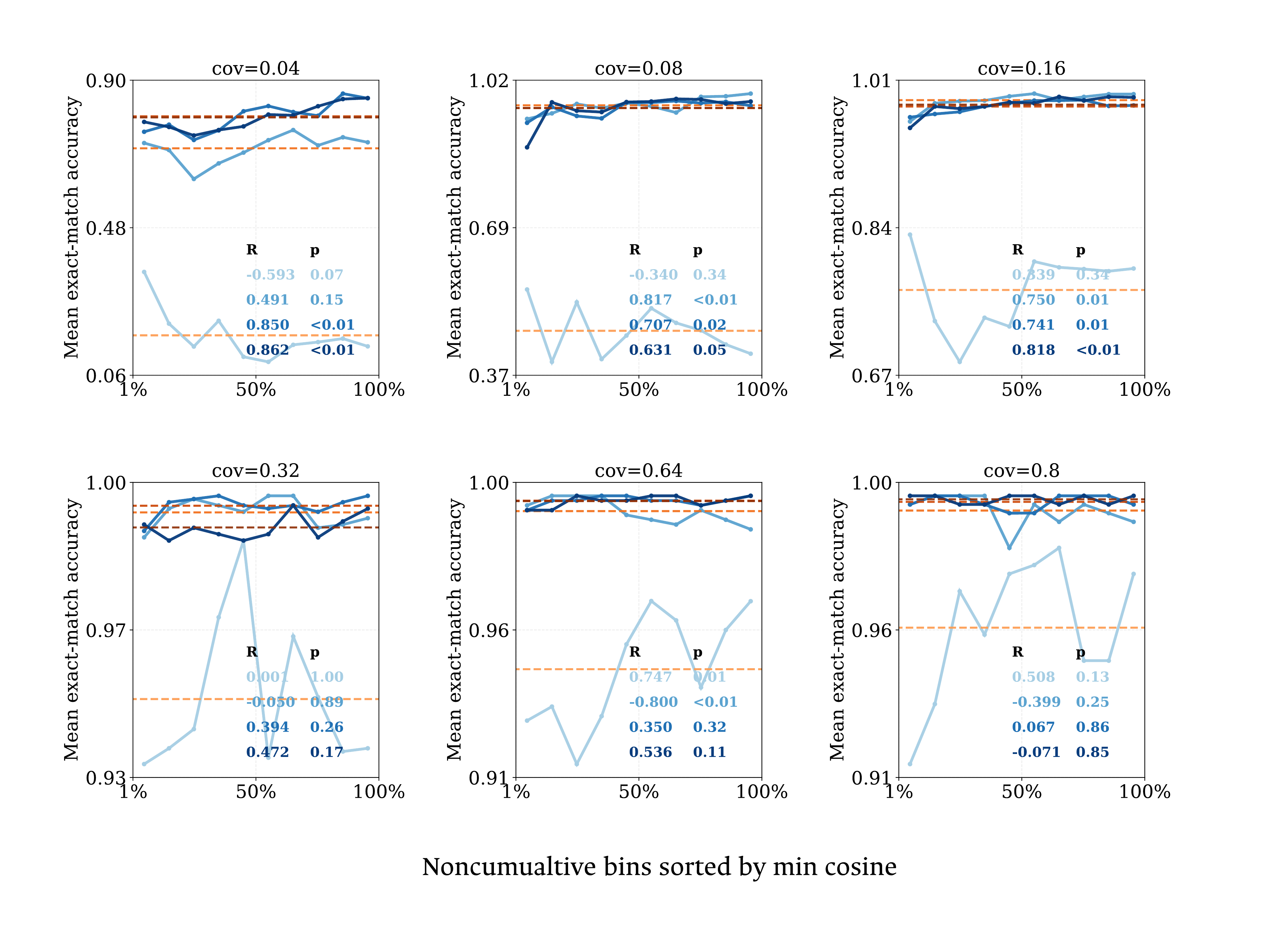}
    \caption{Noncumulative SCAN accuracy curves when examples are binned by
    minimum absolute pairwise cosine similarity.}
    \label{fig:appendix_scan_other_metrics_min_noncumulative}
\end{figure}

\paragraph{Minimum pairwise similarity.}
Figures~\ref{fig:appendix_scan_other_metrics_min_cumulative}
and~\ref{fig:appendix_scan_other_metrics_min_noncumulative} instead use the
least-aligned pair. Across the three alternatives, similarity-based metrics
remain predictive in many settings, suggesting that the common signal is the
angular relationship among concept representations. However, cumulative
coherence remains the theoretically motivated primary metric and is the most
stable across the scenarios studied in the paper.

\clearpage

\subsection{Controls for Potential Confounds and Training-Data Alternatives}
\label{sec:confounds}

We summarize how the preceding analyses address potential confounds before
considering training-data statistics as an additional alternative explanation.

First, all primary SCAN analyses rank examples \emph{within the same model},
rather than comparing raw CI values or accuracy across models with different
dimensions, training coverage, or overall performance
(Section~\ref{sec:scan_within_model_control}). CI must therefore explain which
examples a particular model fails on, not merely distinguish weaker models from
stronger ones. 

Second, we control for command complexity by comparing examples
with the same number of constituent primitives. The CI--accuracy relationship
generally persists within these matched groups
(Section~\ref{sec:scan_length_control}), indicating that it is not explained
solely by command length or the number of composed concepts. 

These controls address variation across models and coarse command complexity. A remaining question is whether CI simply recovers
information already available from the \textit{training distribution}. For modern LLMs,
entity popularity, co-occurrence, and related pretraining-data statistics are
often unavailable or impractical to obtain: pretraining corpora are typically
not public, exhaustive counting at pretraining scale is costly, and it is not
clear which statistics should reliably predict compositional failures. SCAN
provides a controlled setting in which the complete training distribution is
available. We therefore use it for two additional analyses: (1) a direct
comparison of CI with constituent frequency and pairwise co-occurrence, and
(2) a fixed-data experiment testing whether seed-specific learned geometry
contains predictive information beyond the shared training set.

\paragraph{Direct comparison with frequency and co-occurrence.}
We compare CI with two natural data-based predictors: the average training
frequency of an example's constituent concepts and their average pairwise
co-occurrence. Table~\ref{tab:data_statistics} reports the point-biserial
correlation of each metric with binary correctness across non-degenerate
settings.

\begin{table}[htbp]
\centering
\small
\begin{tabular}{ccrrr}
\toprule
\textbf{Model dimension} & \textbf{Training coverage} &
\textbf{Frequency} & \textbf{Co-occurrence} & \textbf{CI} \\
\midrule
12 & 4\%  & $-0.1122$ & $-0.1369$ & $\mathbf{-0.2828}$ \\
32 & 4\%  & $-0.0683$ & $-0.0914$ & $\mathbf{-0.1830}$ \\
64 & 4\%  & $-0.0706$ & $-0.0968$ & $\mathbf{-0.2185}$ \\
8  & 8\%  & $-0.0027$ & $-0.0238$ & $\mathbf{-0.1206}$ \\
12 & 8\%  & $-0.0791$ & $-0.0950$ & $\mathbf{-0.1599}$ \\
32 & 8\%  & $-0.1073$ & $-0.1232$ & $\mathbf{-0.1662}$ \\
64 & 8\%  & $-0.0950$ & $-0.1113$ & $\mathbf{-0.1761}$ \\
8  & 16\% & $-0.0354$ & $-0.0561$ & $\mathbf{-0.0746}$ \\
32 & 16\% & $-0.0504$ & $\mathbf{-0.0587}$ & $-0.0449$ \\
64 & 16\% & $-0.0887$ & $-0.0982$ & $\mathbf{-0.1387}$ \\
8  & 32\% & $-0.0546$ & $-0.0605$ & $\mathbf{-0.1400}$ \\
8  & 64\% & $-0.1321$ & $-0.1420$ & $\mathbf{-0.1809}$ \\
8  & 80\% & $-0.1048$ & $\mathbf{-0.1179}$ & $-0.0778$ \\
\bottomrule
\end{tabular}
\caption{Point-biserial correlations between binary correctness and CI or
training-data statistics (Frequency, Co-occurrence) across non-degenerate SCAN settings. Bold indicates
the largest absolute correlation in each row. All reported CI correlations are
significant at \(p<0.001\).}
\label{tab:data_statistics}
\end{table}

Frequency and co-occurrence have only weak associations with correctness. CI
has the largest absolute correlation in 11 of the 13 settings and a
substantially larger effect in most cases. In the corresponding failure-ranking
analysis, CI also outperforms both frequency- and co-occurrence-based rankings
across all included settings. Thus, these simple training-data statistics do
not explain CI's predictive performance.

\paragraph{Fixed training data with seed-specific geometry.}
\label{sec:scan_fixed_data_control}
The preceding comparison rules out two particular data statistics, but it does
not establish whether the fixed training set fully determines the geometry
captured by CI. We therefore train models with fixed dimension (d = 32) on exactly the same data (coverage ratio of \(32\%\)) while varying
\textit{only} the random seed, retaining the 75 models with non-extreme accuracy
\((\mathrm{accuracy}<0.99)\). If training data completely determined the
predictive ranking, a CI ranking imported from another seed should perform as
well as the target model's own ranking.

For each target model, we evaluate its self-ranking and the 74 rankings imported
from the other seeds using failure-detection PR-AUC. We rank these 75 candidates
from worst to best for each target. The self-ranking achieves a mean rank of
\(42.07/75\), significantly above the null expectation of \(38.0\) (one-sided
permutation test with 100{,}000 permutations, \(p=0.008\)). The null distribution
is constructed by randomly permuting the correspondence between rankings and
target models.

This difference is especially informative because the models have nearly
identical aggregate performance (mean accuracy \(=0.9855\), standard deviation
\(=0.004\)). Nevertheless, each model's own CI ranking predicts its failures
better than rankings imported from models trained on the same examples.

These results do not imply that representation geometry is independent of
training data. Rather, geometry is a model-internal summary shaped by training
data together with optimization and initialization. The direct baselines and
fixed-data control jointly show that this geometry contains model-specific
predictive information not captured by the simple training-data statistics
considered here.

\section{Multihop Dataset Details}
\label{sec:appendix_multihop_dataset}

We expand on the dataset details provided in Section~\ref{sec:real_LLM_section} for multihop QA task. We study two-hop factual reasoning tasks using the datasets and 10-shot prompting setup from \citet{apoorv_multihop}. 

To test whether CI computation with the mean-centering substep generalizes beyond common compositions, we additionally construct several datasets targeting less common multihop compositions that span different representation clusters. These datasets include \texttt{int-plus8-parity}, \texttt{int-plus2-parity}, \texttt{int-plus5-parity}, \texttt{int-plus2-str}, \texttt{int-plus5-str}, \texttt{int-plus8-str}, and \texttt{artist-birthyear-times-two}. Since these are structured tasks, we construct the integer-based datasets by sampling 1000 valid input--output combinations from the appropriate sets. The exception is \texttt{artist-birthyear-times-two}, which we construct by adapting the original artist birth-year data from \citet{apoorv_multihop}.

We filter to examples for which the model answers both constituent single-hop queries correctly, ensuring that errors reflect compositional failures rather than missing single-hop knowledge. In total, our multihop evaluation contains 26 datasets.

%\clearpage
\section{Multihop Additional Results}
\label{sec:appendix_multihop}

We expand on the multihop QA experiments from
Section~\ref{sec:real_LLM_section}.

\begin{figure}[ht]
    \centering
    \includegraphics[width=1.0\columnwidth]{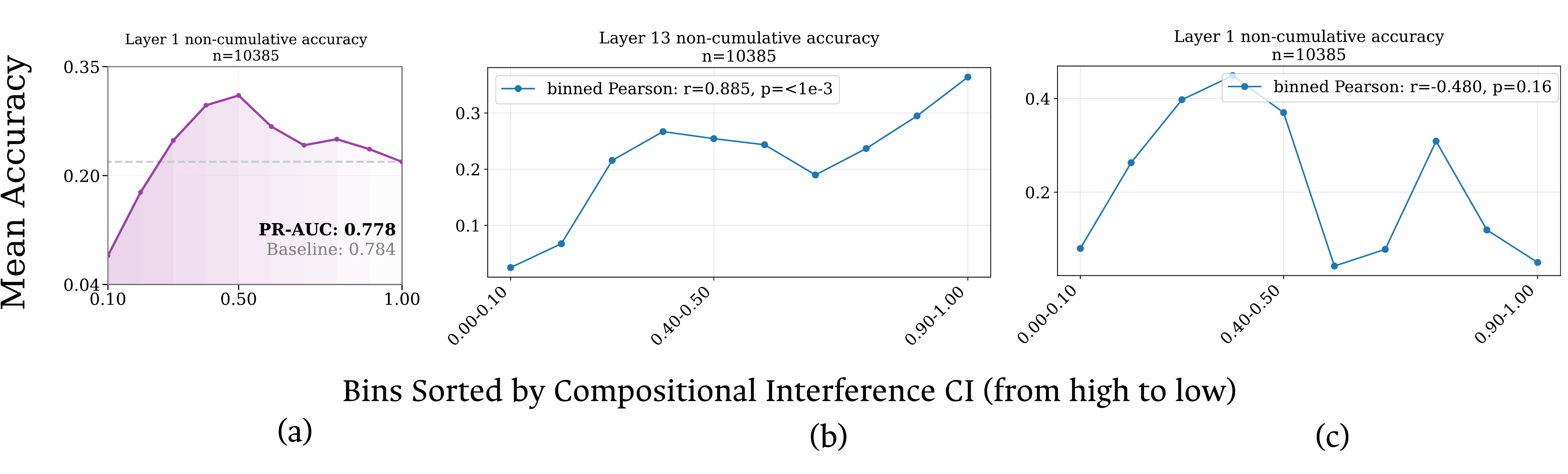}
    \caption{\textbf{Cumulative and noncumulative accuracy across the CI ranking
    for multihop QA.} Examples are ordered from highest CI (predicted most
    difficult) to lowest CI (predicted easiest). In the cumulative plot, each
    point reports accuracy over all examples from the start of the ranking
    through the corresponding cutoff. In the noncumulative plots, each point
    reports accuracy only for examples falling within the corresponding
    interval of the ranking. \textbf{(a)} Cumulative accuracy using atomic
    representations without mean-centering. \textbf{(b)} Noncumulative accuracy
    using mean-centered atomic representations. \textbf{(c)} Noncumulative
    accuracy using atomic representations without mean-centering. The lower
    accuracy among high-CI examples across all three settings shows that CI
    consistently ranks multihop examples by difficulty.}
    \label{fig:appendix_multihop_other_results}
\end{figure}

Calculating CI as defined in Section~\ref{sec:ci_theory},
panel (a) of Figure~\ref{fig:real_LLM_results} and panel (b) of
Figure~\ref{fig:appendix_multihop_other_results} show the expected trend:
accuracy increases as the ranking moves from high-CI examples (predicted to be
more difficult) to low-CI examples (predicted to be easier). This relationship
holds in both the cumulative and noncumulative analyses. CI achieves a
failure-detection PR-AUC of $0.873$, compared with the failure-rate baseline of
$0.784$. In the noncumulative analysis, accuracy also increases strongly from
high- to low-CI intervals (binned Pearson $r=0.885$, $p<10^{-3}$).
Table~\ref{tab:appendix_multihop_rpb} further confirms the relationship at the
example level: CI is significantly negatively correlated with correctness
($r_{\mathrm{pb}}=-0.210$). Together, these results show that CI reliably ranks
multihop examples by failure likelihood.

\begin{table}[ht]
\centering
\small
\begin{tabular}{lc}
\toprule
\textbf{Setting} & \(r_{\mathrm{pb}}\) \\
\midrule
With mean-centering & \(-0.210\) {\scriptsize \((p<0.01)\)} \\
Without mean-centering & \(0.178\) {\scriptsize \((p<0.01)\)} \\
\bottomrule
\end{tabular}
\caption{
Point-biserial correlations between CI and binary correctness (before and after mean-centering). With mean-centering, CI is negatively correlated with correctness, while without mean-centering, the sign reverses.
}
\label{tab:appendix_multihop_rpb}
\end{table}

\begin{figure}[ht]
    \centering
    \includegraphics[width=0.7\columnwidth]{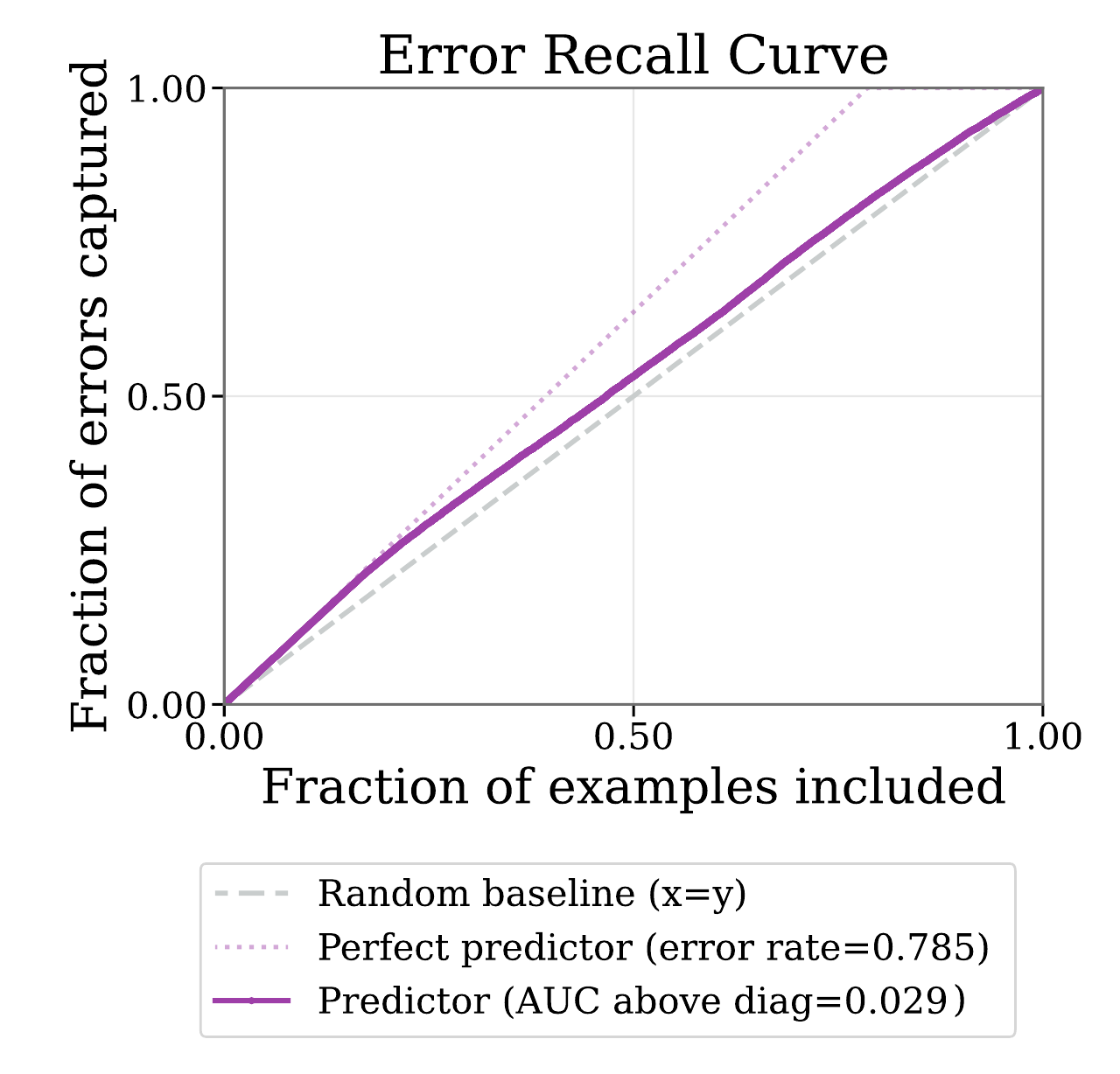}
    \caption{\textbf{Top-$k$ error recovery using CI for multihop QA.}
    Examples are ranked from highest to lowest CI and inspected in that order.
    The $x$-axis gives the fraction of all examples inspected, and the $y$-axis
    gives the fraction of all model errors recovered among those examples.
    Random inspection follows the diagonal $y=x$ and has an expected AUC of
    $0.5$. The CI-ranked curve lies above this baseline, showing that CI
    concentrates multihop failures near the beginning of the ranking and
    recovers them more efficiently than random inspection.}
    \label{fig:appendix_multihop_error_recall}
\end{figure}

Another way to see CI’s practical value is to ask a simple search question: how much of the dataset do we need to inspect to find a specific portion of the model’s errors? For example, if we inspect examples in a random order, we would need to examine 25\% of the dataset on average to find 25\% of the errors – random search does not concentrate errors toward the beginning. But when we inspect examples from highest to lowest CI, Figure~\ref{fig:appendix_multihop_error_recall} shows that we consistently needs less data to find the specific porition of errors. At each cutoff, the $x$-axis reports
the fraction of all examples inspected, while the $y$-axis reports the fraction
of the model's total errors recovered within the inspected subset. Thus, a
point $(x,y)$ means that inspecting the highest-CI fraction $x$ of the examples
identifies a fraction $y$ of all multihop errors.

Under random inspection, the expected fraction of recovered errors equals the
fraction of inspected examples, producing the diagonal $y=x$ and an expected
AUC of $0.5$. A curve above this diagonal indicates that errors are concentrated
near the top of the CI ranking: the farther the curve lies above the diagonal,
the fewer examples must be inspected to recover a given fraction of failures.
The observed curve shows that ranking examples by CI recovers multihop failures
more efficiently than random inspection, supporting the use of CI to construct
targeted multihop challenge sets.

We also ablate the cluster-based mean-centering step from
Section~\ref{sec:ci_theory} to evaluate its importance in calculating CI.
Panels (a) and (c) of Figure~\ref{fig:appendix_multihop_other_results} show that
the predictive relationship does not hold without mean-centering. The
failure-detection PR-AUC decreases to $0.778$, below the failure-rate baseline
of $0.784$, and the noncumulative relationship is not significant (binned
Pearson $r=-0.480$, $p=0.16$). Table~\ref{tab:appendix_multihop_rpb} similarly
shows that the negative example-level correlation disappears without
mean-centering. This ablation demonstrates that the cluster-based mean-centering
procedure introduced under \emph{Accounting for multiscale structure} in
Section~\ref{sec:ci_theory} is necessary to obtain CI scores that are comparable
across clusters and predictive at the global level.

\begin{figure}[ht]
    \centering
    \includegraphics[width=0.75\columnwidth]{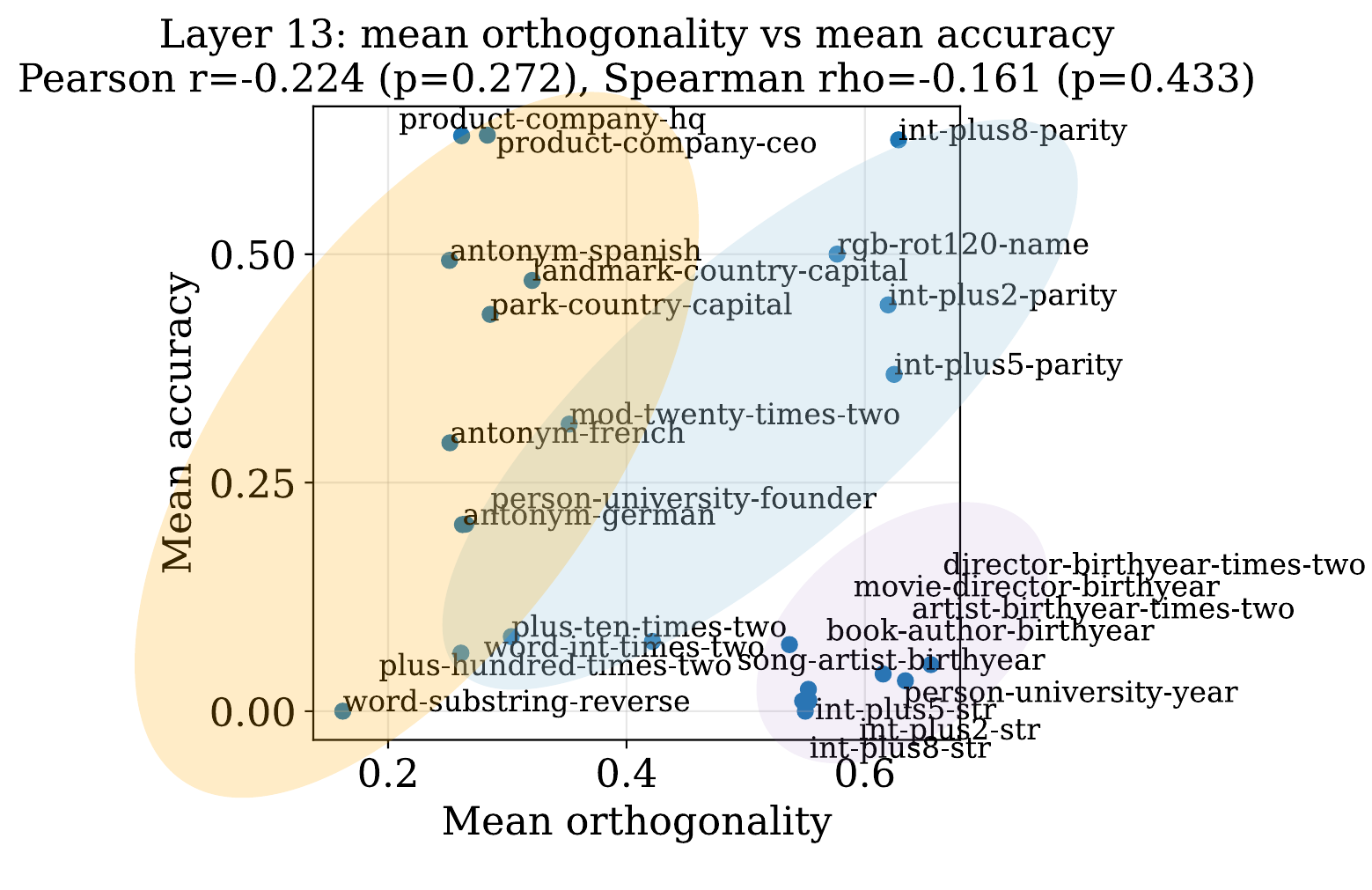}
    \caption{\textbf{Group-level structure obscures the relationship between CI
    and multihop accuracy before mean-centering.} Examples are grouped according
    to their cluster-pair assignments. Within individual groups, higher CI is
    associated with lower accuracy; however, differences in the CI ranges and
    baseline accuracies of these groups mask this relationship when all examples
    are pooled. This Simpson's paradox effect motivates cluster-based
    mean-centering, which removes shared group-level components and makes CI
    comparable across cluster pairs.}
    \label{fig:appendix_multihop_mean_centering_simpson}
\end{figure}

\begin{figure}[ht]
    \centering
    \includegraphics[width=1.0\columnwidth]{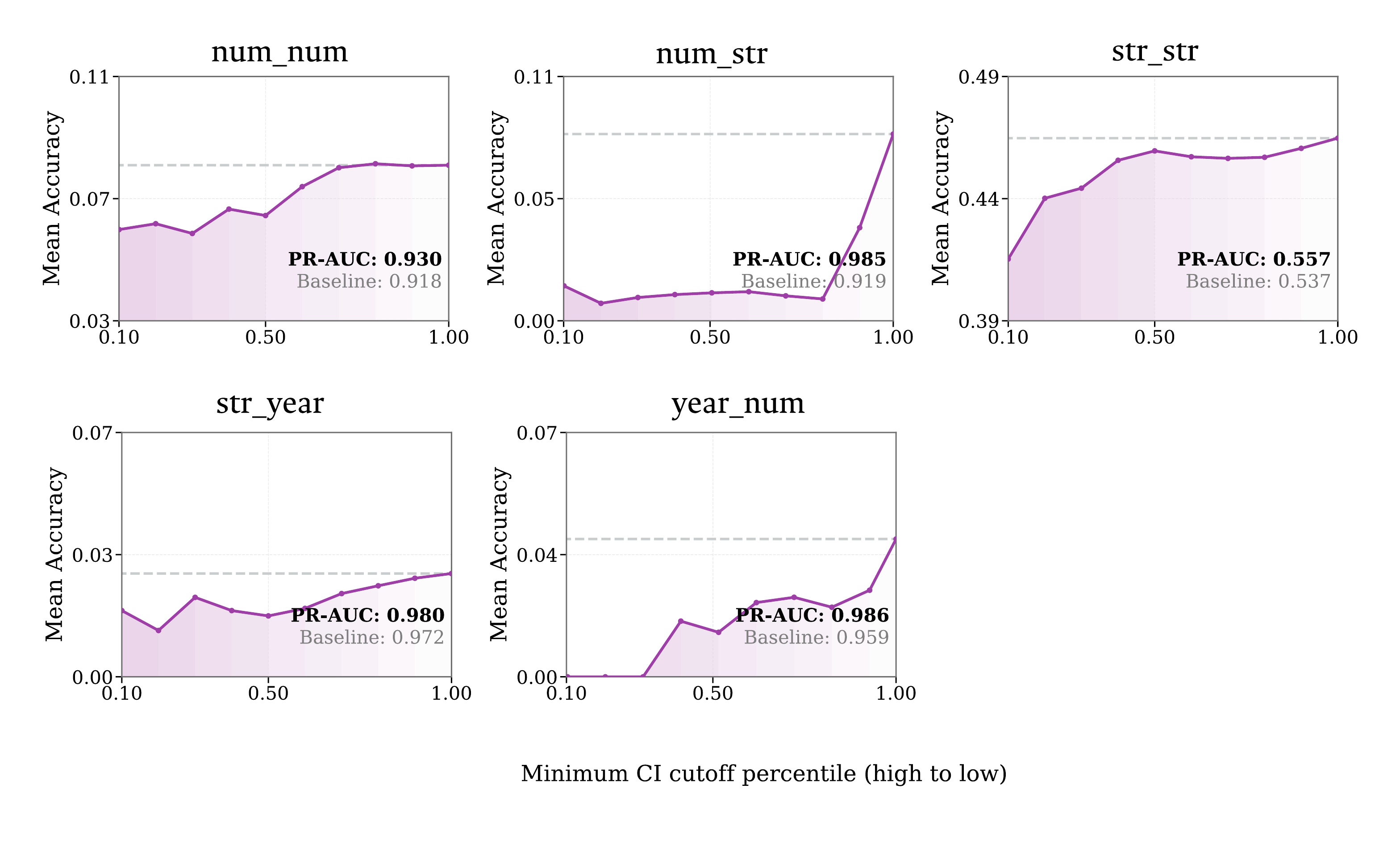}
    \caption{As discussed in Section~\ref{sec:ci_theory}, model representations can cluster by domain-level structure that is not directly relevant to compositional interference. When we break examples into within-cluster-pair settings, the trends remain strong within each setting, indicating that CI is still predictive of interference and model failure. However, the absolute CI value ranges are shifted across cluster pairs, making raw CI values difficult to compare globally. This motivates our mean-centering procedure: by adjusting for these cluster-level offsets, we can recover a comparable CI signal across different representation clusters.}
    \label{fig:appendix_multihop_mean_centering_subsets}
\end{figure}

Figure~\ref{fig:appendix_multihop_mean_centering_simpson} explains why
mean-centering is necessary. Before mean-centering, cluster-pair groups occupy
different ranges of CI and have different baseline accuracies. Consequently,
pooling examples across groups obscures the negative relationship between CI
and correctness that is present within individual groups -- a Simpson's-paradox
effect. Figure~\ref{fig:appendix_multihop_mean_centering_subsets} confirms that,
within each cluster-pair subset, higher CI predicts a greater likelihood of
failure. Cluster-based mean-centering removes shared group-level components
from the atomic representations, making CI comparable across cluster pairs.

\clearpage
\section{Multilingual Fact-Recall Experiment Setup and Additional Results}
\label{sec:appendix_multilingual}

We provide additional example-level results for the multilingual fact-recall
experiments from Section~\ref{sec:real_LLM_section}. Because the figures use
language abbreviations, Table~\ref{tab:language_abbrev_mapping} first maps each
abbreviation to its full language name.

\begin{table}[ht]
\centering
\begin{tabular}{ll}
\toprule
\textbf{Abbreviation} & \textbf{Language} \\
\midrule
\texttt{nl} & Dutch \\
\texttt{ru} & Russian \\
\texttt{fr} & French \\
\texttt{es} & Spanish \\
\texttt{zh} & Chinese \\
\texttt{hu} & Hungarian \\
\texttt{uk} & Ukrainian \\
\texttt{vi} & Vietnamese \\
\texttt{ja} & Japanese \\
\texttt{ko} & Korean \\
\bottomrule
\end{tabular}
\caption{\textbf{Language abbreviations used in the multilingual fact-recall
results.}}
\label{tab:language_abbrev_mapping}
\end{table}

Table~\ref{tab:multilingual_fact_rpb} evaluates the direct example-level
relationship between CI and binary correctness. The point-biserial correlation
is negative for every language, ranging from $-0.288$ for Dutch to $-0.081$
for Chinese, and all correlations are significant at $p<0.01$. Thus, across
all ten languages, examples with greater interference are more likely to be
answered incorrectly.

\begin{table}[t]
\centering
\small
\begin{tabular}{lc}
\toprule
\textbf{Lang.} & \(r_{\mathrm{pb}}\) \\
\midrule
es & \(-0.261\) {\scriptsize \((p<0.01)\)} \\
fr & \(-0.126\) {\scriptsize \((p<0.01)\)} \\
hu & \(-0.186\) {\scriptsize \((p<0.01)\)} \\
ja & \(-0.267\) {\scriptsize \((p<0.01)\)} \\
ko & \(-0.088\) {\scriptsize \((p<0.01)\)} \\
nl & \(-0.288\) {\scriptsize \((p<0.01)\)} \\
ru & \(-0.125\) {\scriptsize \((p<0.01)\)} \\
uk & \(-0.181\) {\scriptsize \((p<0.01)\)} \\
vi & \(-0.191\) {\scriptsize \((p<0.01)\)} \\
zh & \(-0.081\) {\scriptsize \((p<0.01)\)} \\
\bottomrule
\end{tabular}
\caption{\textbf{Point-biserial correlations between CI and correctness in
multilingual fact recall.} Negative values indicate that examples with higher
CI are less likely to be answered correctly. The relationship is negative and
statistically significant in all ten languages.}
\label{tab:multilingual_fact_rpb}
\end{table}

Table~\ref{tab:exp1_pr_auc_summary} evaluates CI as a ranking signal for
failure detection. CI achieves a higher PR-AUC than the failure-rate baseline
in every language, with improvements ranging from $0.059$ for Korean to
$0.200$ for Dutch. All improvements are significant under permutation testing
($p_{\mathrm{perm}}<0.001$), and every bootstrap confidence interval excludes
zero. These results show that CI identifies fact-recall failures more
effectively than an uninformative ranking across all ten languages.

\begin{table}[t]
\centering
\small
\begin{tabular}{lcccc}
\toprule
\textbf{Language} & \textbf{PR-AUC} & \textbf{Baseline}
& \(p_{\mathrm{perm}}\) & \textbf{95\% CI} \\
\midrule
Spanish    & 0.509 & 0.326 & $<0.001$ & [0.152, 0.218] \\
French     & 0.494 & 0.349 & $<0.001$ & [0.116, 0.180] \\
Hungarian  & 0.659 & 0.541 & $<0.001$ & [0.096, 0.145] \\
Japanese   & 0.697 & 0.610 & $<0.001$ & [0.061, 0.114] \\
Korean     & 0.769 & 0.710 & $<0.001$ & [0.041, 0.082] \\
Dutch      & 0.419 & 0.219 & $<0.001$ & [0.154, 0.235] \\
Russian    & 0.495 & 0.331 & $<0.001$ & [0.132, 0.196] \\
Ukrainian  & 0.667 & 0.600 & $<0.001$ & [0.046, 0.096] \\
Vietnamese & 0.715 & 0.628 & $<0.001$ & [0.068, 0.112] \\
Chinese    & 0.650 & 0.587 & $<0.001$ & [0.040, 0.089] \\
\bottomrule
\end{tabular}
\caption{\textbf{Failure-detection performance of CI across languages.}
The baseline is the failure rate, corresponding to the expected PR-AUC of an
uninformative ranking. Significance is assessed by permuting CI scores relative
to the fixed correctness labels. The 95\% bootstrap intervals are computed for
the improvement in PR-AUC over the baseline, with both quantities recomputed
in each resample.}
\label{tab:exp1_pr_auc_summary}
\end{table}

\begin{figure}[ht]
    \centering
    \includegraphics[width=0.6\columnwidth]{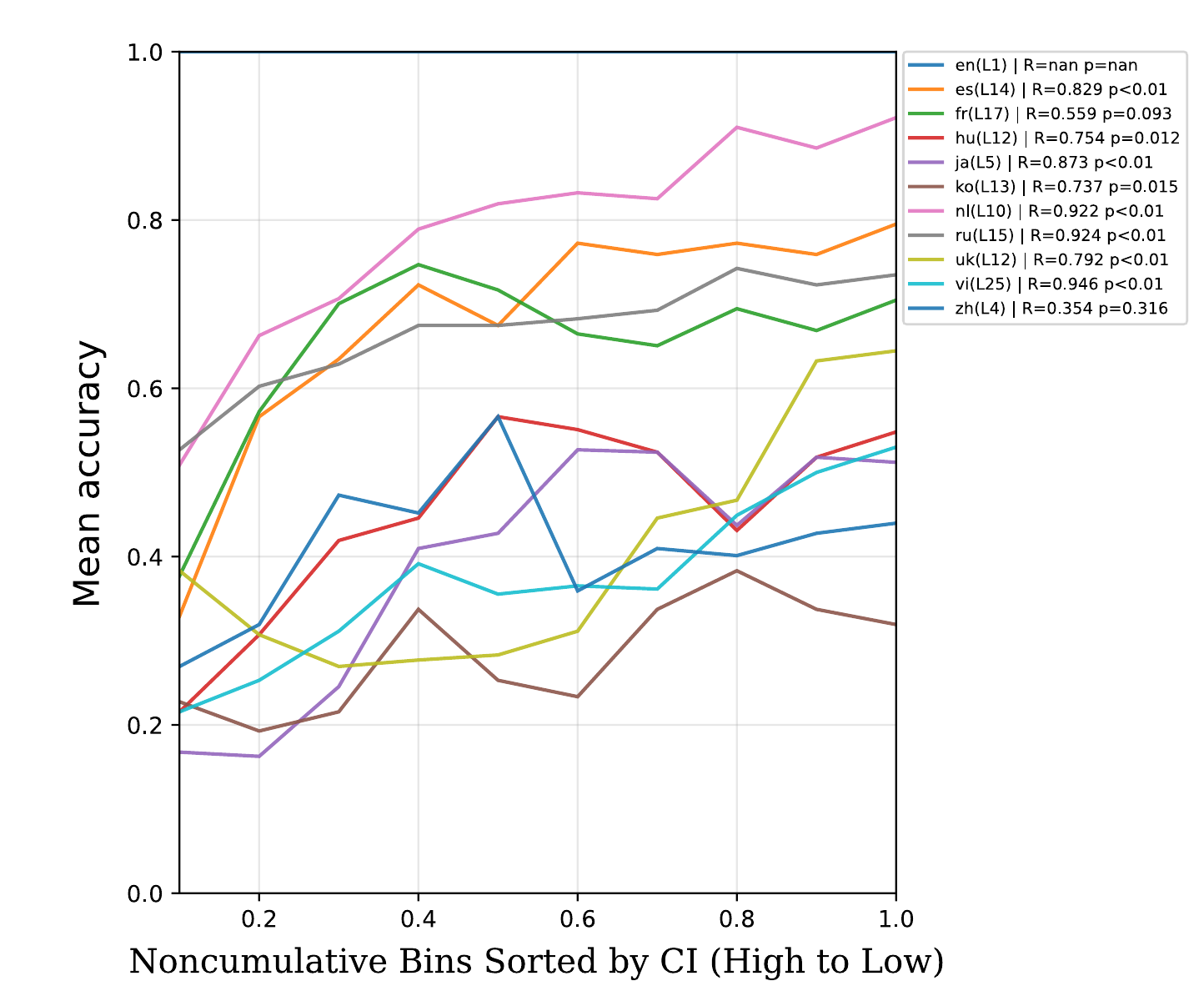}
    \caption{\textbf{Noncumulative accuracy across the CI ranking in
    multilingual fact recall.} Within each language, examples are ordered from
    highest CI (predicted most difficult) to lowest CI (predicted easiest).
    Each point reports accuracy only for examples in the corresponding,
    non-overlapping interval of the ranking. Accuracy generally increases from
    the high-CI intervals to the low-CI intervals, showing that CI ranks
    examples by difficulty across languages.}
    \label{fig:appendix_multilingual_noncumulative}
\end{figure}

Figure~\ref{fig:appendix_multilingual_noncumulative} further examines whether the
predictive relationship holds. Within each language,
examples are ordered from highest CI to lowest CI, and accuracy is calculated
non-cumulatively over non-overlapping intervals of that ordering.
Accuracy generally increases as CI decreases, showing that the relationship
between interference and failure persists throughout the ranking.

\begin{figure}[ht]
    \centering
    \includegraphics[width=0.6\columnwidth]{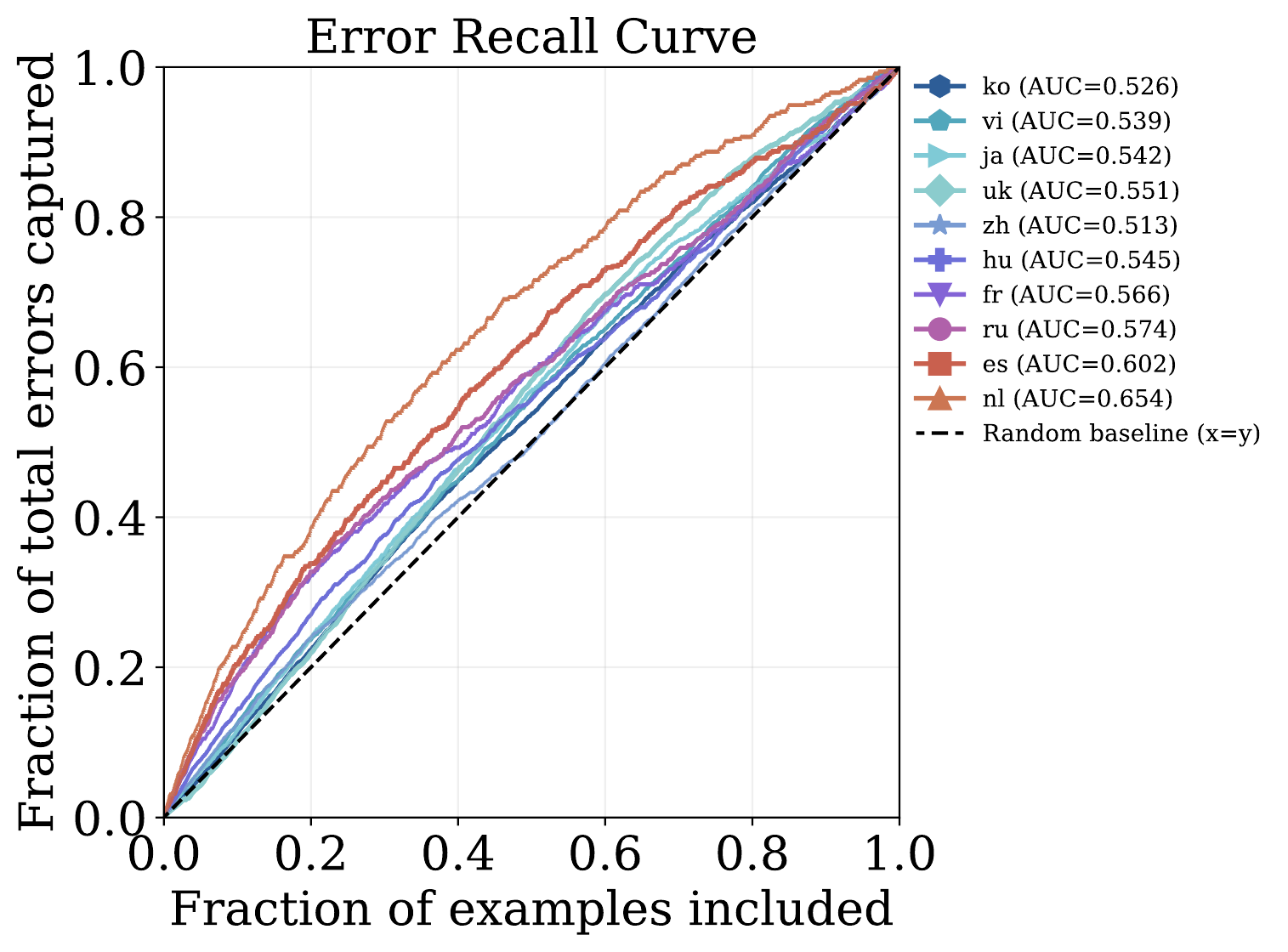}
    \caption{\textbf{Top-$k$ error recovery using CI in multilingual fact
    recall.} Within each language, examples are ranked from highest to lowest
    CI. The $x$-axis gives the fraction of examples inspected from the top of
    this ranking, and the $y$-axis gives the fraction of all errors recovered
    among those examples. Random inspection follows the diagonal $y=x$ and has
    an expected AUC of $0.5$; curves above the diagonal indicate that CI
    concentrates errors near the beginning of the ranking. CI outperforms
    randomly permuted rankings for every language
    ($p_{\mathrm{perm}}<0.01$), showing that it recovers failures faster than
    random inspection.}
    \label{fig:appendix_multilingual_error_recall}
\end{figure}

Figure~\ref{fig:appendix_multilingual_error_recall} provides a complementary
evaluation of how effectively CI can guide targeted error inspection. For each
language, we rank examples from highest CI (predicted most likely to fail) to
lowest CI. At each inspection cutoff, the $x$-axis reports the fraction of all
examples inspected, while the $y$-axis reports the fraction of the model's
total errors contained in that inspected subset. For example, a point
$(x,y)$ indicates that inspecting the highest-CI fraction $x$ of the dataset
recovers a fraction $y$ of all errors. Under a random ranking, the expected fraction of errors recovered equals the
fraction of examples inspected, producing the diagonal $y=x$ and an area under
the error-recall curve of $0.5$. A curve above this diagonal indicates that
errors are concentrated near the top of the CI ranking and can therefore be
recovered with less inspection. A perfect ranking, in which all errors precede
all correct examples, would approach the upper-left corner and attain the
maximum possible AUC. Across all languages, CI achieves significantly better
error recovery than randomly permuted rankings
($p_{\mathrm{perm}}<0.01$). Thus, inspecting examples in descending CI order
recovers multilingual fact-recall failures more efficiently than random
inspection, supporting the use of CI to construct targeted challenge sets.

Finally, Figure~\ref{fig:appendix_multilingual_other_metrics} compares CI with
the alternative interference metrics defined in
Appendix~\ref{sec:alternative_ci_metrics}. This analysis tests whether the
predictive signal depends on the particular aggregation used by CI or also
appears under related geometric measures.

\begin{figure}[ht]
    \centering
    \includegraphics[width=1.0\columnwidth]{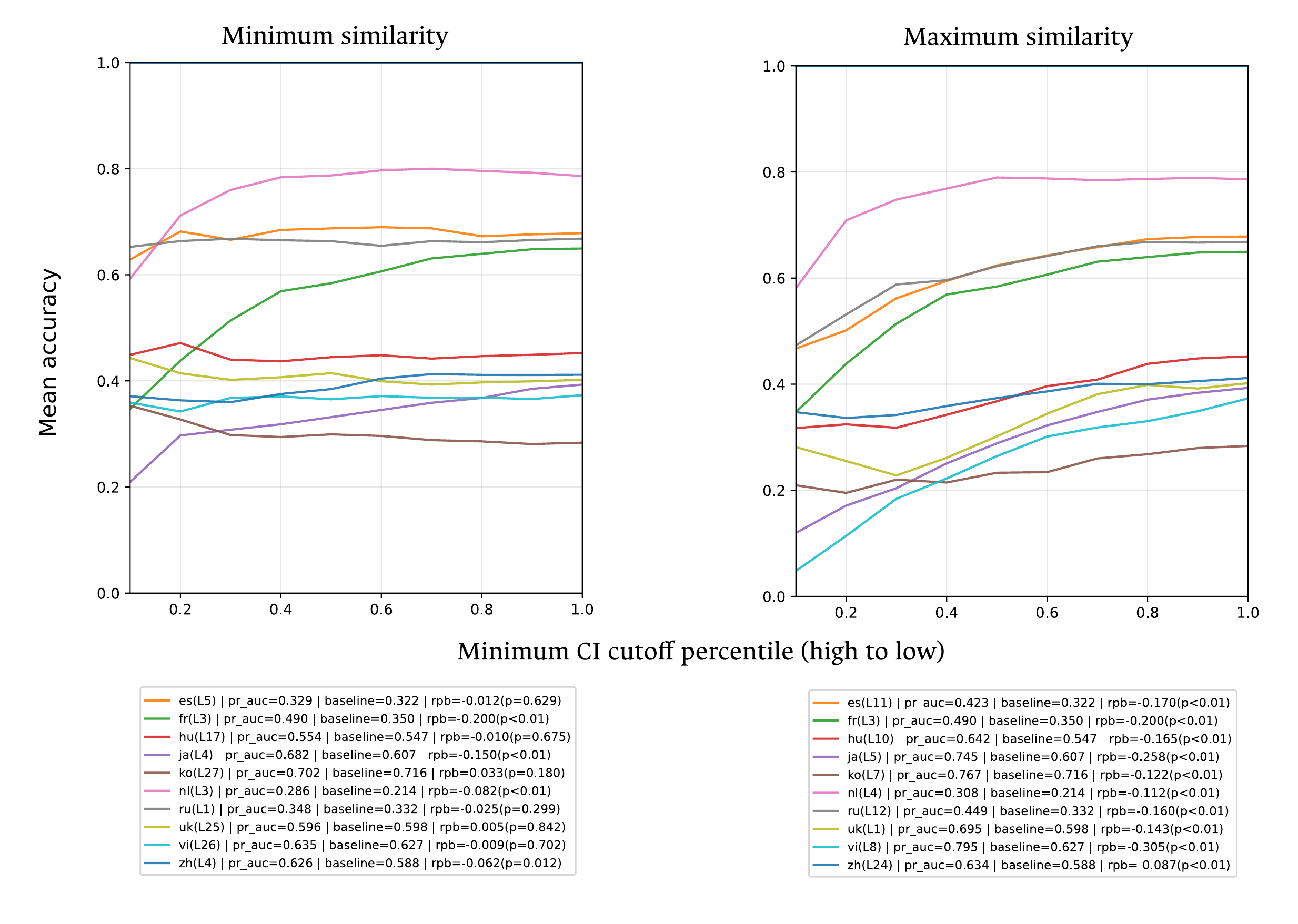}
    \caption{\textbf{Alternative interference metrics for multilingual
    fact recall.} We compare CI with the alternative geometric measures
    described in Appendix~\ref{sec:alternative_ci_metrics}. We omit the
    mean-similarity metric because it is not distinct from CI in this setting
    when the subspace bases are orthonormal.}
    \label{fig:appendix_multilingual_other_metrics}
\end{figure}

\clearpage
\section{Coarse-Grained Concept Analysis}
\label{sec:appendix_coarse_grained}

The main experiments evaluate whether CI predicts failures for individual
compositions. We now ask whether the same geometric signal predicts difficulty
at a coarser level: can the geometry of broad concept categories predict a
model's average accuracy on the corresponding task group? For example, rather
than predicting whether the model will recall one particular capital in
Japanese, we predict its average accuracy on the broader category of national
capitals in Japanese.

For each coarse category, we construct a task-level subspace from the residual
representations of examples belonging to that category. 
We extract subspaces by applying uncentered SVD to the cluster-centered residual representations and retaining the basis vectors that explain a fixed proportion of variance. We sweep variance thresholds of 0.85, 0.90, 0.95, and 0.99 across layers on the development set, and report the best-performing configuration. When constructing category subspaces, we discard category groups with fewer than five correct examples. We then compute subspace-level
CI between the categories being composed and compare it with mean accuracy on
the corresponding group. Category groups with fewer than five valid examples
are excluded. Thus, each point in the following analyses represents an
aggregate task group rather than an individual test example.

\begin{figure}[htbp]
    \centering
    \includegraphics[width=0.5\columnwidth]{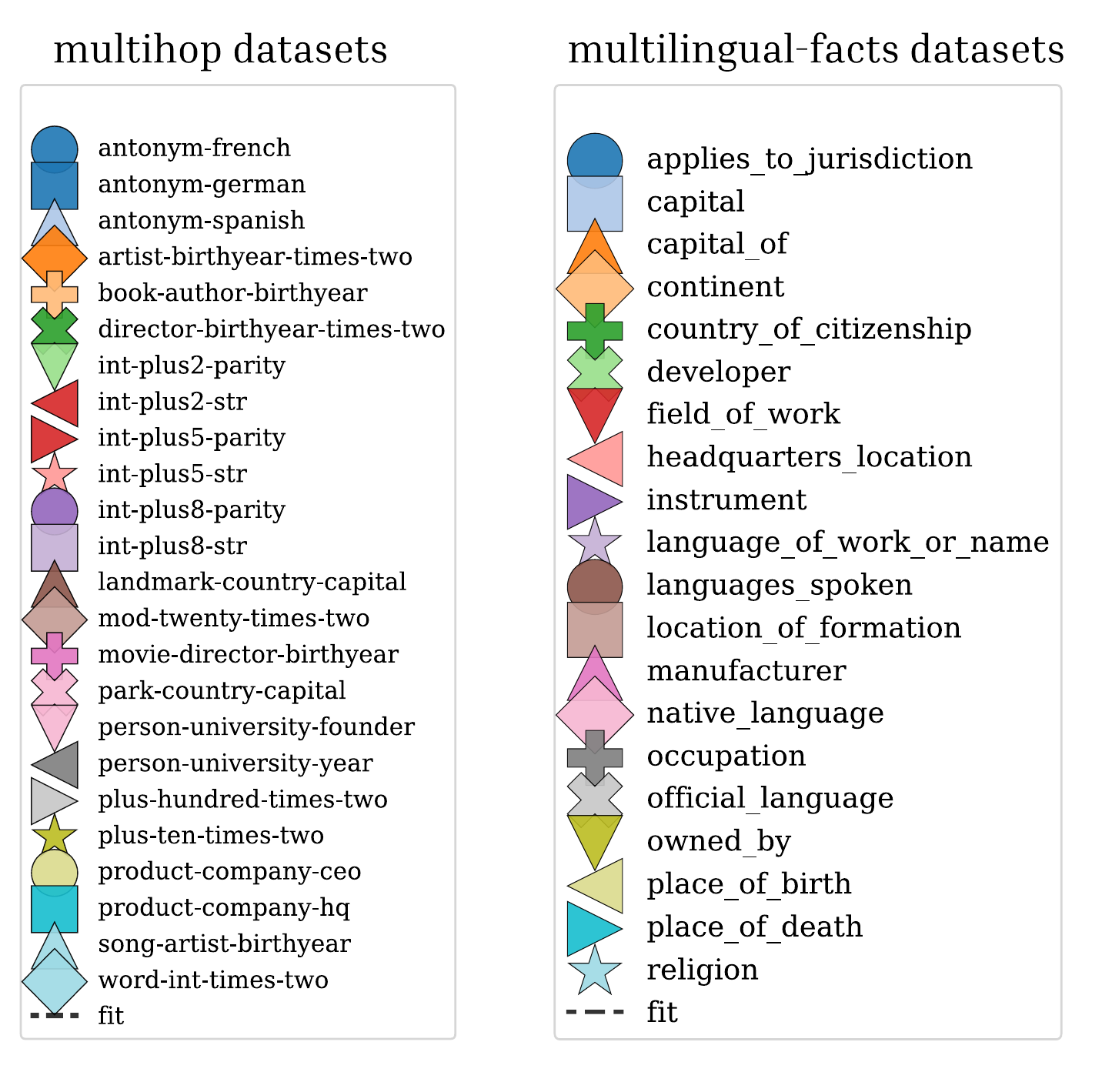}
    \caption{Legends for the coarse-grained analyses in
    Figure~\ref{fig:real_LLM_sub_sub}(b,c). The legends identify the
    multilingual factual-recall task groups and multihop datasets represented
    by the individual points.}
    \label{fig:appendix_coarse_fig_dataset_labels}
\end{figure}

The legends identifying the task groups and datasets shown in the
coarse-grained panels of Figure~\ref{fig:real_LLM_sub_sub} are provided in
Figure~\ref{fig:appendix_coarse_fig_dataset_labels}.

\subsection{Coarse-Grained Multihop QA}

\begin{figure}[htbp]
    \centering
    \includegraphics[width=1.0\columnwidth]{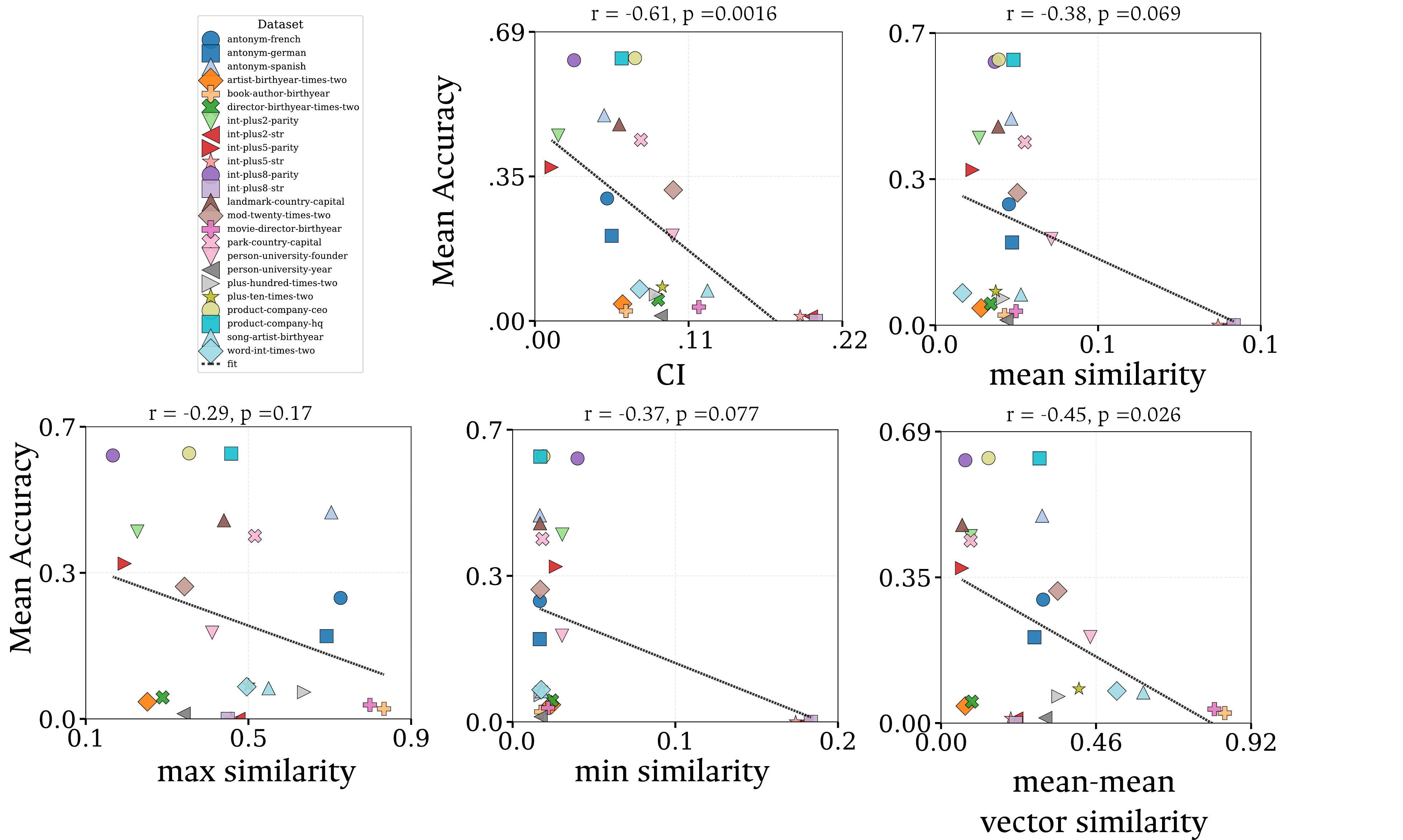}
    \caption{Coarse-grained multihop QA. Each point represents a multihop task
    category, with mean task accuracy plotted against interference between its
    constituent task-level subspaces. We report the primary subspace-level CI,
    alternative subspace-similarity metrics, and a mean task-vector similarity
    baseline. The primary CI has the strongest negative correlation with task
    accuracy (\(r=-0.61\), \(p=0.0016\)).}
    \label{fig:appendix_multihop_other_metrics_subspace_subspace}
\end{figure}

Figure~\ref{fig:appendix_multihop_other_metrics_subspace_subspace} evaluates
whether interference between task-level subspaces predicts mean accuracy across
multihop datasets. The primary subspace-level CI exhibits a significant
negative correlation with accuracy (\(r=-0.61\), \(p=0.0016\)): task categories
whose subspaces interfere more strongly are, on average, more difficult for the
model.

The figure also evaluates the minimum, maximum, and mean subspace-similarity
metrics defined in Section~\ref{sec:alternative_ci_metrics}. All exhibit
negative correlations, although their magnitudes are weaker and not all are
statistically significant. This consistency suggests that angular alignment
between category representations carries the central predictive signal, while
the theoretically motivated CI provides the strongest predictor in this
analysis. Finally, we evaluate a task-vector variant in which examples from a
category are averaged into a single vector, following
\citet{hendel2023context}. Similarity between these mean task vectors is also a
moderate negative predictor of accuracy, but is weaker than subspace-level CI.

\clearpage

\subsection{Coarse-Grained Multilingual Factual Recall}

For multilingual factual recall, each point corresponds to a coarse fact-recall datapoint in KLAR. We compute the interference between that
topic's English factual-representation subspace and the target-language
subspace, then compare it with the model's mean accuracy for that specific multilingual fact-recall dataset. We perform the correlation analysis separately
within each language so that the result is not driven by large differences in
overall accuracy or resource level across languages.

\begin{figure}[htbp]
    \centering
    \includegraphics[width=1.0\columnwidth]{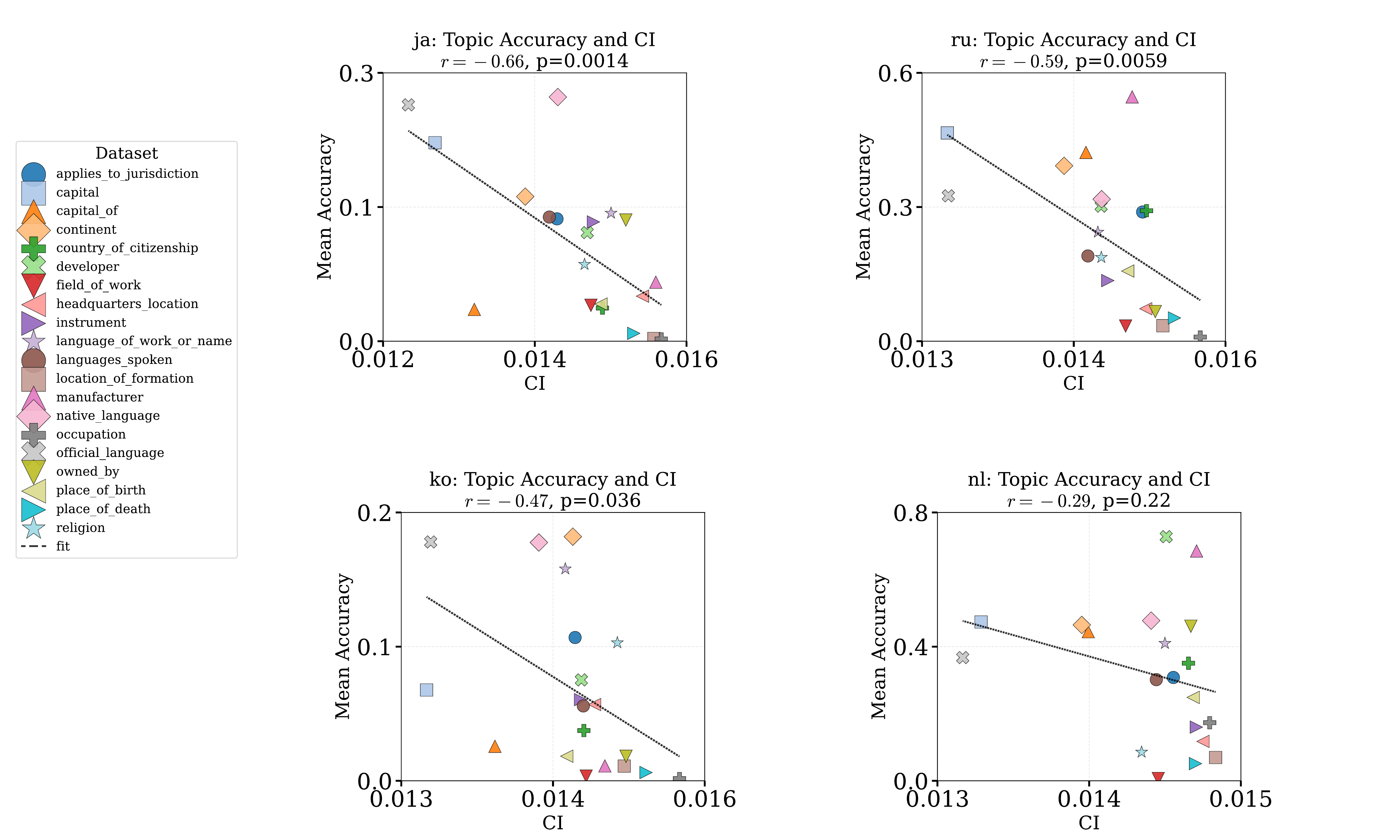}
    \caption{Within-language coarse-grained results for Japanese, Russian,
    Korean, and Dutch. Each point represents one semantic topic group. The
    horizontal axis shows subspace-level CI between the English factual
    category and the target-language subspace, and the vertical axis shows mean
    factual-recall accuracy for that language--topic group.}
    \label{fig:appendix_subspace_subspace_multilingual1}
\end{figure}

Figure~\ref{fig:appendix_subspace_subspace_multilingual1} reports the
results for Japanese, Russian, Korean, and Dutch. The
correlations are negative in all four languages and are strongest for Japanese,
Russian, and Korean, showing that subspace-level interference can distinguish
the relative difficulty of semantic topics within a language.
\begin{figure}[htbp]
    \centering
    \includegraphics[width=1.0\columnwidth]{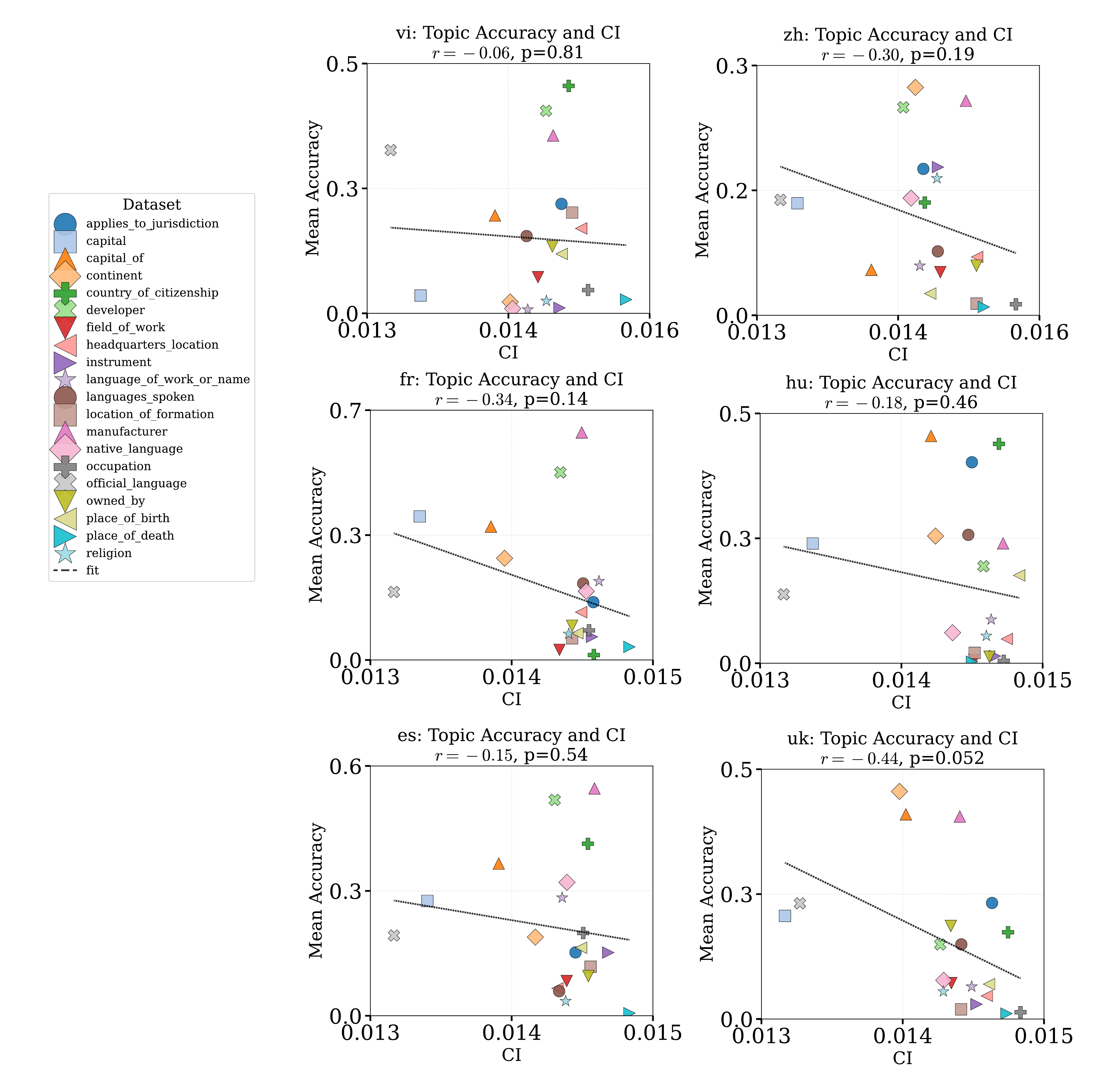}
    \caption{Within-language coarse-grained results for Vietnamese, Chinese,
    French, Hungarian, Spanish, and Ukrainian. Each point represents one
    semantic topic group; higher subspace-level CI is generally associated with
    lower mean factual-recall accuracy, although the strength of the
    relationship differs across languages.}
    \label{fig:appendix_subspace_subspace_multilingual2}
\end{figure}

Figure~\ref{fig:appendix_subspace_subspace_multilingual2} shows the remaining
target languages: Vietnamese, Chinese, French, Hungarian, Spanish, and
Ukrainian. The correlations again have the predicted negative sign, although
their strength varies across languages. Accordingly, we view the coarse-grained
analysis as complementary to the stronger example-level results: category-level
aggregation retains a geometric difficulty signal, but also introduces
language- and topic-specific variation.

\subsection{Robustness Across Subspace-Level Metrics}

Table~\ref{tab:multilingual_subspace_subspace_language_corr_other_metrics}
compares CI with minimum, mean, and maximum subspace similarity. Across
languages, most correlations are negative, supporting the broader hypothesis
that greater angular alignment is associated with lower compositional
accuracy. 
\begin{table}[htbp]
  \centering
  \small
  \begin{tabular}{lcccc}
  \toprule
  \textbf{Lang.}
  & \(\mathrm{CI}\)
  & \textbf{Min. Sim.}
  & \textbf{Mean Sim.}
  & \textbf{Max Sim.} \\
  \midrule
  nl & -0.48 {\scriptsize \((p=0.032)\)} & -0.46 {\scriptsize
  \((p=0.039)\)} & -0.62 {\scriptsize \((p=0.004)\)} & -0.34
  {\scriptsize \((p=0.139)\)} \\
  ru & -0.07 {\scriptsize \((p=0.762)\)} & -0.38 {\scriptsize
  \((p=0.103)\)} & -0.56 {\scriptsize \((p=0.011)\)} & 0.19
  {\scriptsize \((p=0.415)\)} \\
  fr & -0.34 {\scriptsize \((p=0.138)\)} & -0.12 {\scriptsize
  \((p=0.622)\)} & -0.54 {\scriptsize \((p=0.014)\)} & -0.07
  {\scriptsize \((p=0.757)\)} \\
  es & -0.45 {\scriptsize \((p=0.048)\)} & 0.09 {\scriptsize
  \((p=0.715)\)} & -0.43 {\scriptsize \((p=0.060)\)} & -0.16
  {\scriptsize \((p=0.510)\)} \\
  zh & -0.06 {\scriptsize \((p=0.811)\)} & 0.15 {\scriptsize
  \((p=0.529)\)} & -0.21 {\scriptsize \((p=0.375)\)} & -0.35
  {\scriptsize \((p=0.131)\)} \\
  hu & -0.50 {\scriptsize \((p=0.025)\)} & -0.20 {\scriptsize
  \((p=0.391)\)} & -0.29 {\scriptsize \((p=0.212)\)} & -0.10
  {\scriptsize \((p=0.661)\)} \\
  uk & -0.68 {\scriptsize \((p<10^{-3})\)} & -0.44 {\scriptsize
  \((p=0.052)\)} & -0.27 {\scriptsize \((p=0.253)\)} & 0.12
  {\scriptsize \((p=0.602)\)} \\
  vi & 0.11 {\scriptsize \((p=0.651)\)} & 0.07 {\scriptsize
  \((p=0.763)\)} & 0.38 {\scriptsize \((p=0.109)\)} & 0.29
  {\scriptsize \((p=0.229)\)} \\
  ja & -0.26 {\scriptsize \((p=0.264)\)} & -0.20 {\scriptsize
  \((p=0.408)\)} & -0.36 {\scriptsize \((p=0.124)\)} & 0.32
  {\scriptsize \((p=0.168)\)} \\
  ko & -0.51 {\scriptsize \((p=0.021)\)} & -0.40 {\scriptsize
  \((p=0.079)\)} & 0.60 {\scriptsize \((p=0.005)\)} & -0.18
  {\scriptsize \((p=0.446)\)} \\
  \bottomrule
  \end{tabular}
  \caption{Per-language Spearman correlations between mean
  topic-group accuracy and
  subspace-level interference metrics, computed on held-out test
  facts at the layer
  selected on the development facts. Each cell reports \(\rho\),
  with its \(p\)-value
  in parentheses; correlations are computed over \(n=20\) semantic
  topic groups per
  language. Negative values indicate that
  greater subspace
  interference is associated with lower mean accuracy.}
  \label{tab:multilingual_subspace_subspace_language_corr_other_metrics}
  \end{table}

\clearpage

\subsection{Within-Language Versus Pooled Analysis}

\begin{figure}[htbp]
    \centering
    \includegraphics[width=1.0\columnwidth]{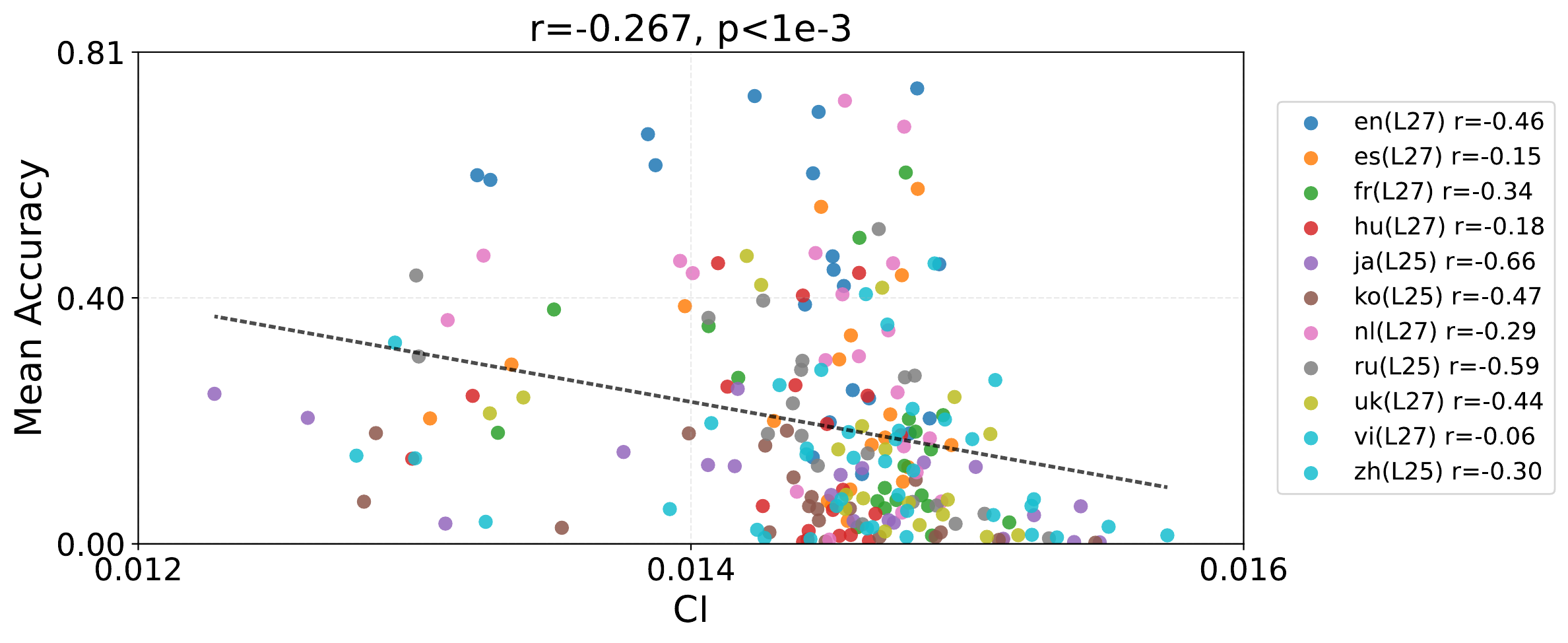}
    \caption{Pooled coarse-grained multilingual analysis. Each point represents
    one language--topic pair, and all languages are included in a single
    CI--accuracy correlation. The pooled relationship is only weakly negative,
    as between-language differences in baseline difficulty partially obscure
    the stronger within-language relationships.}
    \label{fig:appendix_subspace_subspace_merged_lans}
\end{figure}

The preceding analyses deliberately compute correlations within each language.
Figure~\ref{fig:appendix_subspace_subspace_merged_lans} instead pools all
language--topic pairs into a single correlation. The pooled relationship is
only weakly negative, despite the consistently negative and often moderate or
strong within-language correlations above. This result is expected when
languages differ in baseline difficulty: resource level, orthographic
similarity to English, and the degree of transfer from English representations
can shift both accuracy and representation geometry across languages. Pooling
therefore mixes the within-language CI signal with between-language variation.
This result motivates our primary within-language evaluation, which asks the
operationally relevant question of which semantic topics a model is most likely
to fail on in a given target language.

\clearpage

\section{Generalizability to Other LLMs}
\label{sec:appendix_generalizability}

We evaluate whether CI generalizes beyond the base
\texttt{meta-llama/Llama-3.2-3B} experiments to models from different families
and at different scales. For multihop QA, we additionally evaluate
\texttt{allenai/OLMo-7B} and \texttt{Qwen/Qwen2.5-14B}. For multilingual
factual recall, we evaluate \texttt{meta-llama/Llama-2-7b-hf} and
\texttt{Qwen/Qwen2.5-14B}. Across both tasks, the results show that the
example-level CI signal is not specific to a single architecture or model
family.

Internal representations can exhibit different large-scale cluster structures
across model families. We identify representation clusters
automatically using \(k\)-means and mean-center each representation relative to
its assigned cluster before computing CI. This applies the same processing rule
to every model without requiring manually specified concept clusters. As in the
main experiments, the representation layer and processing configuration are
selected using the development set and then fixed for held-out evaluation.

\subsection{Multihop QA}

For both additional models, ranking examples from high to low CI produces
generally increasing cumulative-accuracy curves: the subsets assigned the
highest interference are the most difficult, and accuracy approaches the
model-wide average as lower-CI examples are included. Both curves show a small
non-monotonicity within approximately the first 10\% of the ranking. One
possible explanation is that concepts with highest-similarity may lie in the constructive interference regime~\citet{constructive_interference_prieto2026data}. Thus, future work should examine the difference and interaction between constructive and destructive interference.

\begin{figure}[htbp]
    \centering
    \includegraphics[width=1.0\columnwidth]{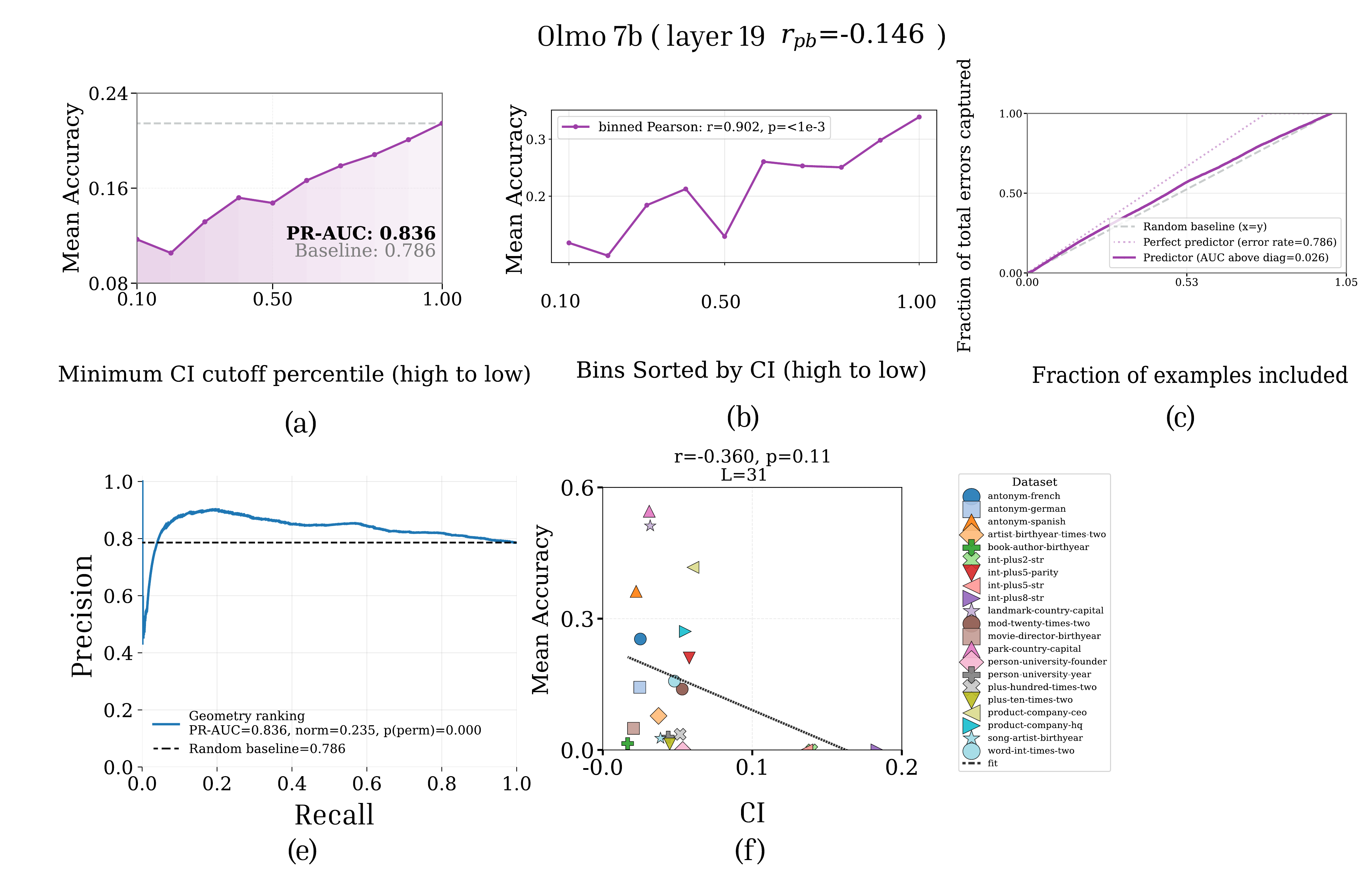}
    \caption{Generalizability to \texttt{allenai/OLMo-7B} at the
    development-selected layer 19 (\(r_{\mathrm{pb}}=-0.146\)).
    (a) Cumulative accuracy after ranking examples from high to low CI.
    (b) Accuracy in noncumulative CI-ranked bins
    (\(r=0.902\), \(p<10^{-3}\)).
    (c) Top-\(k\) error recall relative to random and perfect rankings.
    (e) Failure-detection precision--recall curve; CI achieves PR-AUC
    \(0.836\), compared with the failure-rate baseline \(0.786\).
    (f) Coarse-grained dataset-level analysis
    (\(r=-0.360\), \(p=0.11\)). The example-level evaluations consistently
    support CI, whereas the aggregated relationship is weaker.}
    \label{fig:appendix_generalizability_olmo7b}
\end{figure}

\paragraph{OLMo-7B.}
Figure~\ref{fig:appendix_generalizability_olmo7b} reports the complete results
for \texttt{allenai/OLMo-7B}. Development-set selection chooses layer 19. At
this layer, CI has a significant point-biserial correlation of
\(r_{\mathrm{pb}}=-0.146\) with correctness. The cumulative and noncumulative
analyses agree: higher-CI examples are less accurate, with the noncumulative
bins exhibiting a strong correlation of \(r=0.902\) (\(p<10^{-3}\)). CI also
improves failure detection, achieving PR-AUC \(=0.836\) compared with the
failure-rate baseline of \(0.786\), and the error-recall curve exceeds random
ranking. By contrast, the coarse-grained dataset-level relationship is
moderately negative but nonsignificant (\(r=-0.360\), \(p=0.11\)), suggesting
that aggregation obscures some of the example-level signal.

\begin{figure}[htbp]
    \centering
    \includegraphics[width=1.0\columnwidth]{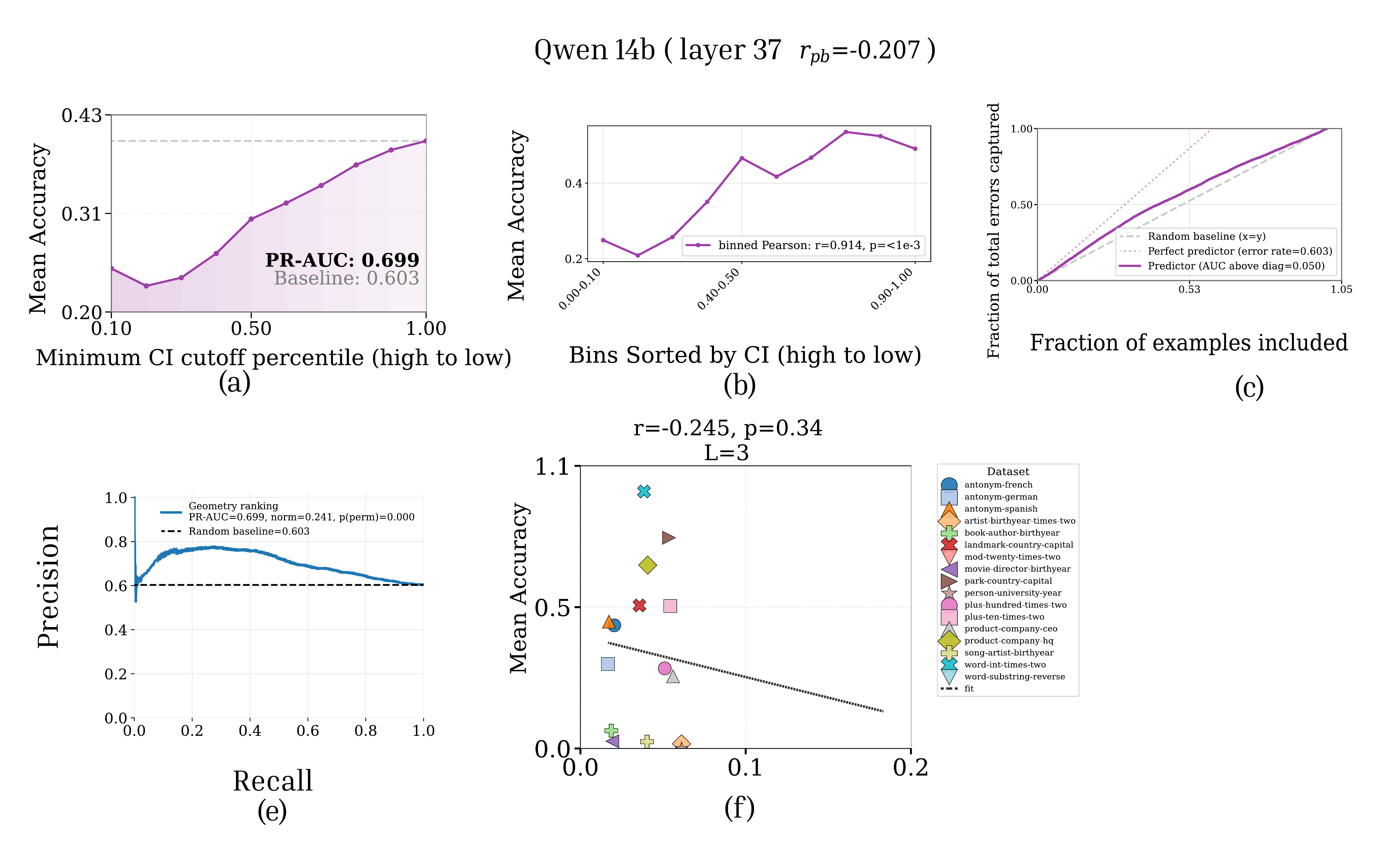}
    \caption{Generalizability to \texttt{Qwen/Qwen2.5-14B} on multihop QA at
    the development-selected layer 37 (\(r_{\mathrm{pb}}=-0.207\)).
    (a) Cumulative accuracy after ranking examples from high to low CI.
    (b) Accuracy in noncumulative CI-ranked bins
    (\(r=0.914\), \(p<10^{-3}\)).
    (c) Top-\(k\) error recall relative to random and perfect rankings.
    (e) Failure-detection precision--recall curve; CI achieves PR-AUC
    \(0.699\), compared with the failure-rate baseline \(0.603\).
    (f) Coarse-grained dataset-level analysis
    (\(r=-0.245\), \(p=0.34\)).}
    \label{fig:appendix_generalizability_qwen14b}
\end{figure}

\paragraph{Qwen2.5-14B.}
Figure~\ref{fig:appendix_generalizability_qwen14b} shows the corresponding
analysis for \texttt{Qwen/Qwen2.5-14B}, with layer 37 selected on the
development set. CI is again negatively correlated with correctness
(\(r_{\mathrm{pb}}=-0.207\)). The noncumulative bins show a strong relationship
(\(r=0.914\), \(p<10^{-3}\)), and CI achieves PR-AUC \(=0.699\), exceeding the
failure-rate baseline of \(0.603\). The error-recall curve likewise outperforms
random inspection. As with OLMo, the coarse-grained correlation is negative but
weaker and nonsignificant (\(r=-0.245\), \(p=0.34\)). Thus, across both model
families, the primary example-level analysis provides substantially clearer
evidence than the supplementary dataset-level aggregation.

\clearpage

\subsection{Multilingual Factual Recall}

For each model and target language, we collect token-level residual-stream
representations from 100 OSCAR sentences, using 30 tokens per sentence. CI measures the
alignment between each English fact representation and that subspace. All
rankings and evaluations are performed separately within each language, so the
results predict which facts a model will fail to recall in a given language
rather than merely ranking languages by their overall difficulty.

\begin{figure}[htbp]
    \centering
    \includegraphics[width=1.0\columnwidth]{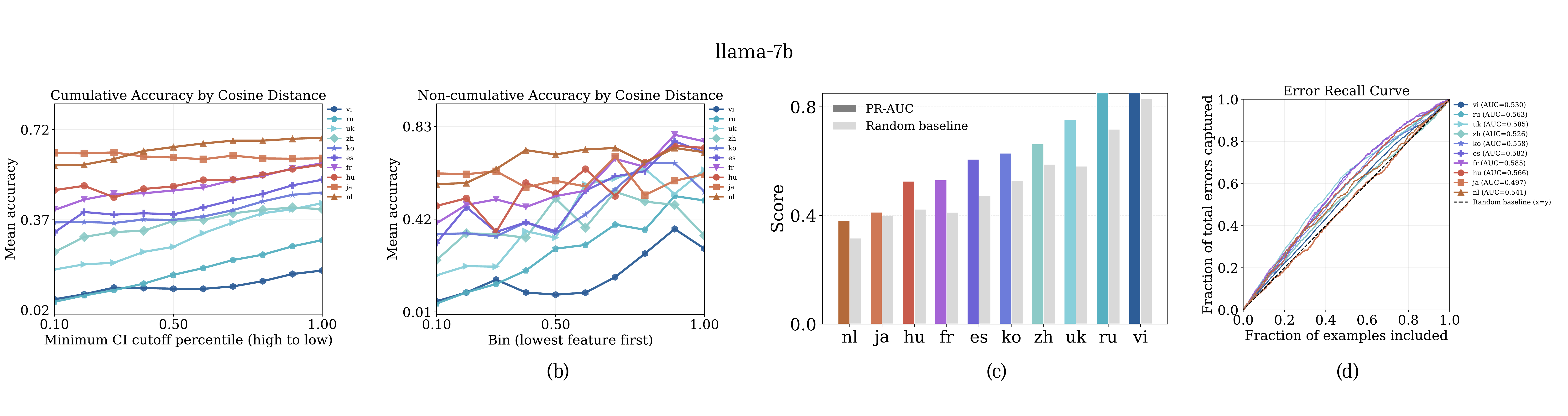}
    \caption{Generalizability to \texttt{meta-llama/Llama-2-7b-hf} on
    multilingual factual recall. (a) Cumulative accuracy after ranking facts
    from high to low CI within each language. (b) Accuracy in noncumulative
    CI-ranked bins. (c) Failure-detection PR-AUC compared with each language's
    failure-rate baseline. CI exceeds the baseline in all ten languages.
    (d) Top-\(k\) error-recall curves; error-recall AUC exceeds \(0.5\) for nine
    languages, with Japanese at \(0.497\).}
    \label{fig:appendix_generalization_multilingual_llama7b}
\end{figure}

\paragraph{Llama-2-7B.}
Figure~\ref{fig:appendix_generalization_multilingual_llama7b} shows that the
cumulative curves generally increase across languages, while the noncumulative
curves provide a noisier but broadly consistent view of the ranking. CI's
PR-AUC exceeds the failure-rate baseline in all ten languages. The top-\(k\)
error-recall AUC also exceeds the random-ranking value of \(0.5\) in nine of ten
languages; Japanese is the sole exception (AUC \(=0.497\)).

\begin{figure}[htbp]
    \centering
    \includegraphics[width=1.0\columnwidth]{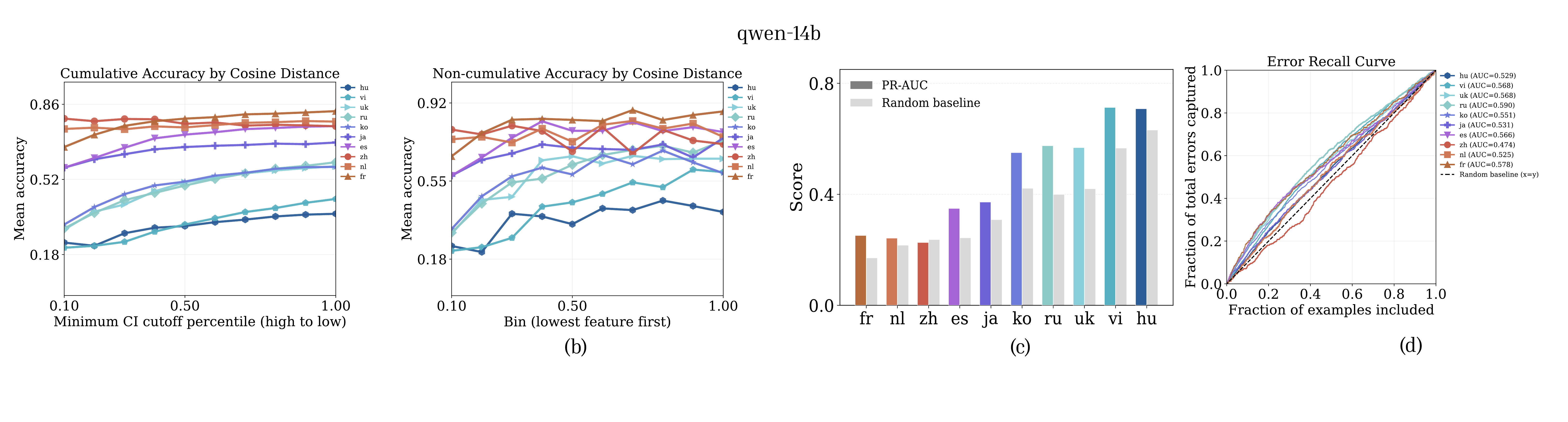}
    \caption{Generalizability to \texttt{Qwen/Qwen2.5-14B} on multilingual
    factual recall. (a) Cumulative accuracy after ranking facts from high to low
    CI within each language. (b) Accuracy in noncumulative CI-ranked bins.
    (c) Failure-detection PR-AUC compared with each language's failure-rate
    baseline; CI exceeds the baseline in nine of ten languages.
    (d) Top-\(k\) error-recall curves; error-recall AUC exceeds \(0.5\) for the
    same nine languages, with Chinese at \(0.474\).}
    \label{fig:appendix_generalization_multilingual_qwen14b}
\end{figure}

\paragraph{Qwen2.5-14B.}
Figure~\ref{fig:appendix_generalization_multilingual_qwen14b} shows a similar
pattern for \texttt{Qwen/Qwen2.5-14B}. The cumulative curves generally increase,
and CI's PR-AUC exceeds the failure-rate baseline in nine of ten languages. The
top-\(k\) error-recall AUC also exceeds \(0.5\) for the same nine languages.
Chinese is the consistent exception: its PR-AUC falls below the failure-rate
baseline, and its error-recall AUC is \(0.474\).

\subsection{Statistical Evaluation Across Languages}

Table~\ref{tab:generalizability_multilingual_pr_auc} evaluates each
model--language ranking with a permutation test. CI outperforms the
failure-rate baseline in 19 of the 20 model--language pairs and does so
significantly in 18. For Llama-2-7B, all ten PR-AUC values are numerically above
baseline, with nine significant at \(p<0.001\); Japanese is above baseline but
not significant (\(p=0.250\)). For Qwen2.5-14B, all nine improvements are
significant (\(p\leq0.020\)); Chinese is the only pair in which CI falls below
baseline.

\begin{table}[htbp]
\centering
\small
\begin{tabular}{llrrl}
\toprule
\textbf{Model} & \textbf{Lang.} & \textbf{PR-AUC} & \textbf{Baseline} & \(p_{\mathrm{perm}}\) \\
\midrule
Llama-2-7B & es & \textbf{0.607} & 0.472 & $<0.001$ \\
            & fr & \textbf{0.531} & 0.411 & $<0.001$ \\
            & hu & \textbf{0.526} & 0.422 & $<0.001$ \\
            & ja & \textbf{0.412} & 0.398 & 0.250 \\
            & ko & \textbf{0.635} & 0.528 & $<0.001$ \\
            & nl & \textbf{0.380} & 0.316 & $<0.001$ \\
            & ru & \textbf{0.870} & 0.717 & $<0.001$ \\
            & uk & \textbf{0.751} & 0.581 & $<0.001$ \\
            & vi & \textbf{0.903} & 0.829 & $<0.001$ \\
            & zh & \textbf{0.663} & 0.588 & $<0.001$ \\
\midrule
Qwen2.5-14B & es & \textbf{0.349} & 0.243 & $<0.001$ \\
              & fr & \textbf{0.252} & 0.171 & $<0.001$ \\
              & hu & \textbf{0.708} & 0.632 & $<0.001$ \\
              & ja & \textbf{0.372} & 0.309 & $<0.001$ \\
              & ko & \textbf{0.550} & 0.422 & $<0.001$ \\
              & nl & \textbf{0.242} & 0.217 & 0.020 \\
              & ru & \textbf{0.575} & 0.400 & $<0.001$ \\
              & uk & \textbf{0.568} & 0.420 & $<0.001$ \\
              & vi & \textbf{0.713} & 0.567 & $<0.001$ \\
              & zh & 0.226 & 0.237 & 0.920 \\
\bottomrule
\end{tabular}
\caption{Failure-detection PR-AUC across models and target languages. The
random-ranking baseline equals the failure rate within each model--language
pair. Bold indicates PR-AUC above that baseline; \(p_{\mathrm{perm}}\) is the
permutation-test significance of the observed ranking.}
\label{tab:generalizability_multilingual_pr_auc}
\end{table}

\paragraph{Interpreting the exceptions.}
Our multilingual metric models factual recall as composition between an English
fact representation and a target-language subspace. Qwen2.5-14B is trained on
substantial Chinese data and may store or retrieve some facts directly in
Chinese rather than composing an English-centered factual representation with a
Chinese transformation. In that case, the English-fact--Chinese-subspace angle
need not predict failure, providing one possible explanation for the Chinese
exception. For Japanese with Llama-2-7B, PR-AUC remains above baseline but the
improvement is not significant. This may similarly indicate greater reliance on
language-specific memorization, although the present experiment does not
directly identify the underlying retrieval pathway. These localized exceptions
qualify the mechanism assumed by the metric but do not alter the broader
cross-model result: CI provides an above-baseline failure ranking in 19 of 20
model--language evaluations across the two additional models.

\clearpage

\section{Discussion: Middle layers are best for computing interference}
\label{sec:layer_sensitivity}

CI's predictive performance varies across representation layers, with intermediate
layers generally selected as most predictive. Using the development-set procedure
described in Section~\ref{sec:metrics}, the median selected layer is 5/10 for SCAN
and 12/27 for multilingual factual recall, while multihop QA selects layer 13/27.

This pattern is consistent with our interference-based hypothesis. Prior work
suggests that intermediate layers integrate concepts needed for the second hop in
multihop reasoning~\citep{apoorv_multihop} and support English-mediated factual
recall in multilingual settings~\citep{lu_paths_2025}. Interference may therefore be most
predictive at layers where the relevant concepts must be jointly represented yet
remain recoverable. This interpretation is suggestive rather than conclusive; a
complete account of how interference emerges and changes across layers remains an
important direction for future work.

\clearpage
\section{Code and Compute}
\label{sec:code_gpu}

We will release the code upon the paper's acceptance. For the SCAN experiments in Section~\ref{sec:scan}, each toy model was trained on a single NVIDIA GeForce RTX 3090 GPU. For the real-LLM experiments in Section~\ref{sec:real_LLM_section}, all runs on Llama are conducted on a single 3090 GPU as well.

%%%%%%%%%%%%%%%%%%%%%%%%%%%%%%%%%%%%%%%%%%%%%%%%%%%%%%%%%%%%

\clearpage

\end{document}

%% file: math_commands.tex
\usepackage{amsmath,amsfonts,bm}

\def\eqref#1{equation~\ref{#1}}
\def\1{\bm{1}}

\DeclareMathAlphabet{\mathsfit}{\encodingdefault}{\sfdefault}{m}{sl}
\SetMathAlphabet{\mathsfit}{bold}{\encodingdefault}{\sfdefault}{bx}{n}